\documentclass[10pt,journal,compsoc]{IEEEtran}

\usepackage{times}
\usepackage{epsfig}
\usepackage{graphicx}
\usepackage{float}
\usepackage{wrapfig}
\usepackage{amsmath,amssymb,amsthm}
\usepackage{bm,xspace}
\usepackage{comment}
\usepackage{verbatim}
\usepackage{multirow}
\usepackage{balance}
\usepackage{url}
\usepackage{booktabs}
\usepackage{etoolbox,siunitx}
\usepackage{calc}
\usepackage{pifont,hologo}
\usepackage[usenames, dvipsnames]{color}
\usepackage{nicefrac}

\newcommand{\ra}[1]{\renewcommand{\arraystretch}{#1}}

\usepackage[super]{nth}
\usepackage{nicefrac}
\robustify\bfseries

\def\cc{\mathbf{c}}
\def\dd{\mathbf{d}}
\def\ee{\mathbf{e}}

\def\hh{\mathbf{h}}

\def\rr{\mathbf{r}}
\def\sss{\mathbf{s}}
\def\tt{\mathbf{t}}
\def\uu{\mathbf{u}}

\def\xx{\mathbf{x}}

\def\EE{\mathbf{E}}

\def\dD{\mathcal{D}}
\def\eE{\mathcal{E}}
\def\fF{\mathcal{F}}

\def\hH{\mathcal{H}}
\def\iI{\mathcal{I}}

\def\lL{\mathcal{L}}

\def\pP{\mathcal{P}}

\def\rR{\mathcal{R}}

\def\wW{\mathcal{W}}

\def\eps{\varepsilon}

\DeclareMathSymbol{@}{\mathord}{letters}{"3B}

\newcommand\norm[1]{\left\lVert#1\right\rVert}

\newcommand\timess{\mathbin{\!\times\!}}

\newcommand\mypara[1]{\vspace{1mm}\noindent\textbf{#1}}

\def\latex/{\LaTeX}
\def\bibtex/{\hologo{BibTeX}}

\makeatletter
\DeclareRobustCommand\onedot{\futurelet\@let@token\@onedot}
\def\@onedot{\ifx\@let@token.\else.\null\fi\xspace}

\def\eg{\emph{e.g}\onedot} 
\def\ie{\emph{i.e}\onedot}

\def\etal{\emph{et al}\onedot}
\makeatother

\def\Event{\ee}
\def\EventSequence{\eps}

\def\VoxelGrid{\EE}
\def\NumEvents{N}
\def\NumBins{B}
\def\Loss{\lL}
\def\OpticFlowMap{\fF}
\def\ReconstructionLoss{\lL^R}
\def\TemporalLoss{\lL^{TC}}

\def\WeightTemporalLoss{\lambda_{TC}}
\def\Image{\iI}
\def\EventTensor{\mathbf{\EE}}
\def\Reconstructed{\hat{\iI}}

\def\Brightness{L}
\def\Pixel{\uu}

\def\NetworkState{\sss}
\def\EncoderState{\cc}
\def\NumEncoders{N_E}
\def\NumResBlocks{N_R}
\def\NumDecoders{\NumEncoders}
\def\NumBaseFeatureChannels{N_b}

\def\Warp{\wW}

\def\ConvLSTM{\mbox{ConvLSTM}}
\def\Head{\hH}
\def\Encoder{\eE}
\def\ResBlock{\rR}
\def\Decoder{\dD}
\def\Prediction{\pP}

\ifCLASSOPTIONcompsoc
  \usepackage[nocompress]{cite}
\else
  \usepackage{cite}
\fi

\newcommand\MYhyperrefoptions{bookmarks=true,bookmarksnumbered=true,
pdfpagemode={UseOutlines},plainpages=false,pdfpagelabels=true,
colorlinks=true,citecolor={black},
pdftitle={High Speed and High Dynamic Range Video with an Event Camera},%
pdfsubject={Computer Vision, Event-based Vision, Video Reconstruction, Neuromorphic Engineering, High Speed, High Dynamic Range},%
pdfauthor={Henri Rebecq, Rene Ranftl, Vladlen Koltun, Davide Scaramuzza}}
\usepackage[\MYhyperrefoptions,pdftex]{hyperref}

\ifCLASSINFOpdf
  \usepackage{graphicx}
\else
\fi

\usepackage{url}

\usepackage{hyperref}

\begin{document}
\title{High Speed and High Dynamic Range Video\\with an Event Camera}

\author{Henri~Rebecq,
        Ren\'e~Ranftl,
        Vladlen~Koltun,
        and~Davide~Scaramuzza%
\IEEEcompsocitemizethanks{\IEEEcompsocthanksitem H. Rebecq and D. Scaramuzza are with the Robotics and Perception Group, affiliated with both the Dept.~of~Informatics of the University~of~Zurich and the Dept.~of~Neuroinformatics of the University~of~Zurich and ETH Zurich, Switzerland.\protect\\
\IEEEcompsocthanksitem R. Ranftl and V. Koltun are with the Intelligent Systems Lab, Intel Labs.}%
\thanks{This work was supported by the the Swiss National Center of Competence Research Robotics (NCCR), the SNSF-ERC Starting Grant and Qualcomm (through the Qualcomm Innovation Fellowship 2018).
}}

\markboth{}%
{Rebecq \MakeLowercase{\textit{et al.}}: High Speed and High Dynamic Range Video\\with an Event Camera}

\IEEEtitleabstractindextext{%
\begin{abstract}
Event cameras are novel sensors that report brightness changes in the form of a stream of asynchronous ``events'' instead of intensity frames.
They offer significant advantages with respect to conventional cameras: high temporal resolution, high dynamic range, and no motion blur.
While the stream of events encodes in principle the complete visual signal, the reconstruction of an intensity image from a stream of events is an ill-posed problem in practice.
Existing reconstruction approaches are based on hand-crafted priors and strong assumptions about the imaging process as well as the statistics of natural images.
In this work we propose to learn to reconstruct intensity images from event streams directly from data instead of relying on any hand-crafted priors.
We propose a novel recurrent network to reconstruct videos from a stream of events, and train it on a large amount of simulated event data.
During training we propose to use a perceptual loss to encourage reconstructions to follow natural image statistics.
We further extend our approach to synthesize color images from color event streams.
Our quantitative experiments show that our network surpasses state-of-the-art reconstruction methods by a large margin in terms of image quality ($>\!20\%$), while comfortably running in real-time.
We show that the network is able to synthesize high framerate videos ($>5@000$ frames per second) of high-speed phenomena (\eg~a bullet hitting an object) and is able to provide high dynamic range reconstructions in challenging lighting conditions.
As an additional contribution, we demonstrate the effectiveness of our reconstructions as an intermediate representation for event data.
We show that off-the-shelf computer vision algorithms can be applied to our reconstructions for tasks such as object classification and visual-inertial odometry and that this strategy consistently outperforms algorithms that were specifically designed for event data.
We release the reconstruction code and a pre-trained model to enable further research.
\end{abstract}

}

\maketitle

\IEEEdisplaynontitleabstractindextext

\IEEEpeerreviewmaketitle

\section*{Multimedia Material}

A video of the experiments, as well as the reconstruction code and a pretrained model are available at: \url{http://rpg.ifi.uzh.ch/E2VID}.

\vspace{1.2cm}

\IEEEraisesectionheading{\section{Introduction}\label{sec:introduction}}

\IEEEPARstart{E}{vent} cameras are bio-inspired vision sensors that work radically differently from conventional cameras.
Instead of capturing intensity images at a fixed rate, event cameras measure \emph{changes} of intensity asynchronously at the time they occur.
This results in a stream of \emph{events}, which encode the time, location, and polarity (sign) of brightness changes
(Fig.~\ref{fig:principle_operation} - top).
Event cameras such as the Dynamic Vision Sensor (DVS)~\cite{Lichtsteiner08ssc} possess outstanding properties when compared to conventional cameras.
They have a very high dynamic range (\SI{140}{\decibel} versus \SI{60}{\decibel}), do not suffer from motion blur, and provide measurements with a latency as low as one microsecond.
Event cameras thus provide a viable alternative (or complementary) sensor in conditions that are challenging for conventional cameras.

\setlength{\fboxsep}{0pt}%
\begin{figure}[t]
    \centering
    \setlength{\tabcolsep}{-0.1cm}
    \begin{tabular}{ccc}
      \includegraphics[trim={1.5cm 1.2cm 1.5cm 1.5cm},clip,height=2.65cm]{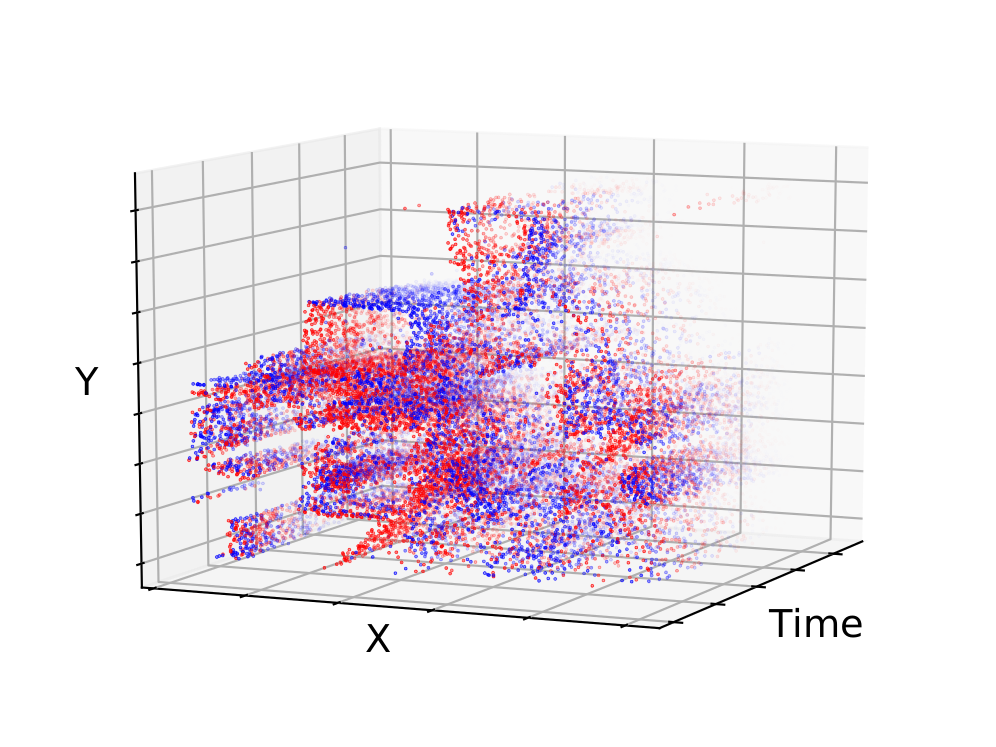}\hspace{0.25cm} &
      \includegraphics[width=1.6cm,trim={0 -2.1cm 0 0cm},clip]{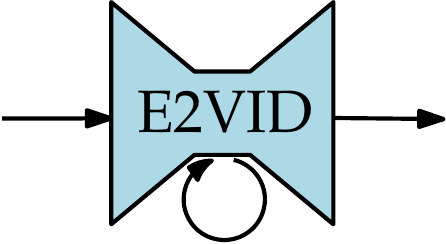}\hspace{0.4cm} &
      \includegraphics[height=3.0cm]{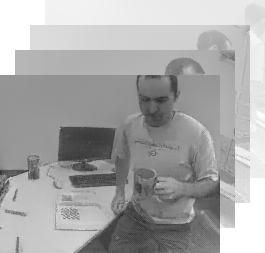}\\[0.2cm]
    \end{tabular}

    \setlength{\tabcolsep}{0.1cm}
    \begin{tabular}{ccc}
      \includegraphics[height=2.5cm]{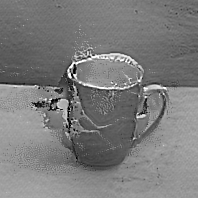} &
      \scalebox{-1}[1]{\includegraphics[height=2.5cm]{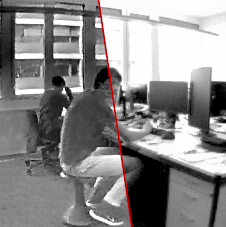}} &  %
      \includegraphics[trim={0cm 25.28cm 10.85cm 0cm},clip,height=2.5cm]{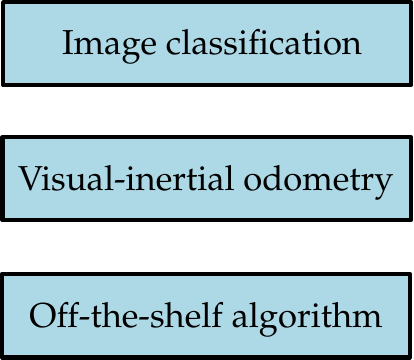}\\
      \footnotesize (a) High framerate video & \footnotesize (b) HDR video & \footnotesize (c) Downstream applications
    \end{tabular}

    \caption{Our network converts a spatio-temporal stream of events with microsecond temporal resolution (top left) into a high-quality video (top right).
    This enables synthesis of videos of high-speed phenomena such as a bullet piercing a mug (a), or scenes with high dynamic range (b).
    The reconstructions can also be used as input to off-the-shelf computer vision algorithms, thereby serving as an intermediate representation between event data and mainstream computer vision (c).
    The images in the figure were produced by the presented technique.%
    \label{fig:eye_catcher}
    }
\end{figure}

In theory, the stream of events contains the entire visual signal -- in a highly compressed form -- and could thus be decompressed to recover a video with arbitrarily high framerate and high dynamic range.
However, real event cameras are noisy and differ significantly from the ideal camera model, which renders the reconstruction problem ill-posed.
Naive integration of the event stream leads to very fast degradation of image quality due to accumulating noise.
As a remedy, earlier works have proposed hand-crafted image priors to constrain the problem \cite{Bardow16cvpr, Barua16wacv, Munda18ijcv, Scheerlinck18accv}.
However, these priors make strong assumptions about the statistics of natural images, leading to unrealistic reconstructions and artifacts.
As a result, high-quality video reconstruction from event data has so far not been convincingly demonstrated. %

In this work, we propose to bridge this gap by learning high-quality video reconstruction from sparse event data using a recurrent neural network.
In contrast to previous image reconstruction approaches \cite{Bardow16cvpr,Munda18ijcv,Scheerlinck18accv}, we do not embed handcrafted smoothness priors into our reconstruction framework.
Instead, we learn video reconstruction from events using a large amount of simulated event data, and encourage the reconstructed images to have natural image statistics through a perceptual loss that operates on mid-level image features.
Our network outperforms prior methods in terms of image quality by a large margin ($>$20\% improvement), demonstrating for the first time event-camera-based synthesized video sequences that are qualitatively on par with conventional cameras in terms of visual appearance.
Our approach opens the door to a variety of applications, some of which we examine in this paper.

We explore the possibility of using an event camera to capture videos in scenarios that are challenging for conventional cameras.
First, we show that our network can leverage the high temporal resolution of event data to synthesize high framerate ($>5@000$ frames per second) videos of high-speed physical phenomena (Section~\ref{sec:highspeed_reconstructions}).
The resulting videos reveal details that are beyond the grasp of the naked eye or conventional cameras, which operate at a few hundred frames per second at best.
Second, we show that our reconstructions preserve the high dynamic range of event cameras (Section~\ref{sec:hdr_reconstructions}), thus offering a viable alternative to conventional sensors in high dynamic range settings.
We additionally present a simple strategy to synthesize color videos using a recent color event camera \cite{Taverni18tcsii}, without the need to retrain the network (Section~\ref{sec:color_reconstruction}).

Beyond pure imaging, we also consider using our approach for downstream applications.
Since the output of an event camera is an asynchronous stream of events (a representation that is fundamentally different from natural images), existing computer vision techniques cannot be directly applied to this data.
As a consequence, a number of algorithms have been specifically tailored to leverage event data, either processing the event stream in an event-by-event fashion \cite{Conradt09iscas,Cook11ijcnn,Benosman14tnnls,Mueggler14iros,Kim16eccv,Gallego17pami}, or
by building intermediate, ``image-like'' representations from event data \cite{Lagorce17pami,Sironi18cvpr,Zhu18rss,Zhou18eccv,Zhu18rss}.
In the spirit of the second category of methods, we explore the use of our image reconstructions as a novel representation for event data in Section~\ref{sec:downstream_applications}.
Specifically, we apply existing computer vision algorithms to images reconstructed from event data.
We focus on object classification and visual-inertial odometry with event data, and show that this strategy consistently yields state of the art results in terms of accuracy.
This suggests that high-quality reconstructions can be used as a bridge that brings the main stream of computer vision research to event cameras: mature algorithms, modern deep network architectures, and weights pretrained from large natural image datasets.

In summary, the contributions of this work are:
\begin{itemize}
  \item A novel recurrent network that reconstructs video from a stream of events and outperforms the state of the art in terms of image quality by a large margin.
  \item We establish that networks trained from simulated event data generalize remarkably well to real events.
  \item Qualitative results showing that our method can be used in a variety of settings, such as high framerate video synthesis of high-speed phenomena (Section~\ref{sec:highspeed_reconstructions}), reconstruction of high dynamic range video (Section~\ref{sec:hdr_reconstructions}), and reconstruction of color video (Section~\ref{sec:color_reconstruction}).
  \item Application of our method to two downstream problems: object classification and visual-inertial odometry from event data. Our method outperforms state-of-the-art algorithms designed specifically for event data in both applications.
\end{itemize}

\section{Related Work}
\label{sec:related_work}

Because of its far reaching applications, events-to-video reconstruction is a popular topic in the event camera literature.
The first evidence that it is possible to recover intensity information from event data was provided by \cite{Cook11ijcnn} and \cite{Kim14bmvc}, in the context of rotation estimation with an event camera.
They showed how to reconstruct a single image from a large set of events collected by an event camera moving through a static scene and exploited the fact that every event provides one equation relating the intensity gradient and optic flow through brightness constancy~\cite{Gehrig18eccv}.
Specifically, Cook \etal~\cite{Cook11ijcnn} used bio-inspired, interconnected networks to simultaneously recover intensity images, optic flow, and angular velocity from an event camera undergoing small rotations.
Kim \etal~\cite{Kim14bmvc} developed an Extended Kalman Filter to reconstruct a 2D panoramic gradient image (later upgraded to a full intensity frame by 2D Poisson integration) from a rotating event camera.
They later extended their approach to static 3D scenes and 6 degrees-of-freedom (6DOF) camera motion \cite{Kim16eccv}.
Bardow \etal~\cite{Bardow16cvpr} proposed to estimate optic flow and intensity \emph{simultaneously} from sliding windows of events through a variational energy minimization framework.  %
They showed the first video reconstruction framework from events that is applicable to dynamic scenes. %
However, their energy minimization framework employs multiple hand-crafted regularizers, which can result in severe loss of detail in the reconstructions. %

Recently, methods based on direct event integration have emerged.
These approaches do not rely on any assumption about the scene structure or motion dynamics, and can naturally reconstruct videos at arbitrarily high framerates.
Munda \etal \cite{Munda18ijcv} cast intensity reconstruction as an energy minimization problem defined on a manifold induced by the event timestamps.
They combined direct event integration with total variation regularization and achieved real-time performance on the GPU.
Scheerlinck \etal \cite{Scheerlinck18accv} proposed to filter the events with a high-pass filter prior to integration.
They demonstrated video reconstruction results that are qualitatively comparable with \cite{Munda18ijcv} while being computationally more efficient.
While these approaches currently define the state-of-the-art, both suffer from artifacts which are inherent to direct event integration.
The reconstructions suffer from ``bleeding edges'' caused by the fact that the contrast threshold (the minimum brightness change of a pixel to trigger an event) is neither constant nor uniform across the image plane.
Additionally, pure integration of the events can in principle only recover intensity up to an unknown initial image $\Image_0$ which causes ``ghosting'' effects where the trace of the initial image remains visible in the reconstructed sequence.

Barua \etal \cite{Barua16wacv} proposed a learning-based approach to reconstruct intensity images from events.
They used \mbox{K-SVD} \cite{Aharon06tsp} on simulated data to learn a dictionary that maps small patches of integrated events to an image gradient and used Poisson integration to recover the intensity image.
In contrast, we do not reconstruct individual intensity images from small windows of events, but synthesize a temporally consistent video from a long stream of events (several seconds) using a recurrent network.
Instead of mapping event patches to a dictionary of image gradients, we learn pixel-wise intensity estimation directly. %

Despite the body of work on events-to-video reconstruction, downstream vision applications based on the reconstructions have, to the best of our knowledge, never been demonstrated prior to our work.

\section{Video Reconstruction}
\label{sec:methodology}

\global\long\def\introPlotHeight{2.55cm}
\begin{figure}[t]
\centering
\setlength{\tabcolsep}{0.06cm}
\begin{tabular}{c}
  \includegraphics[height=3.0cm]{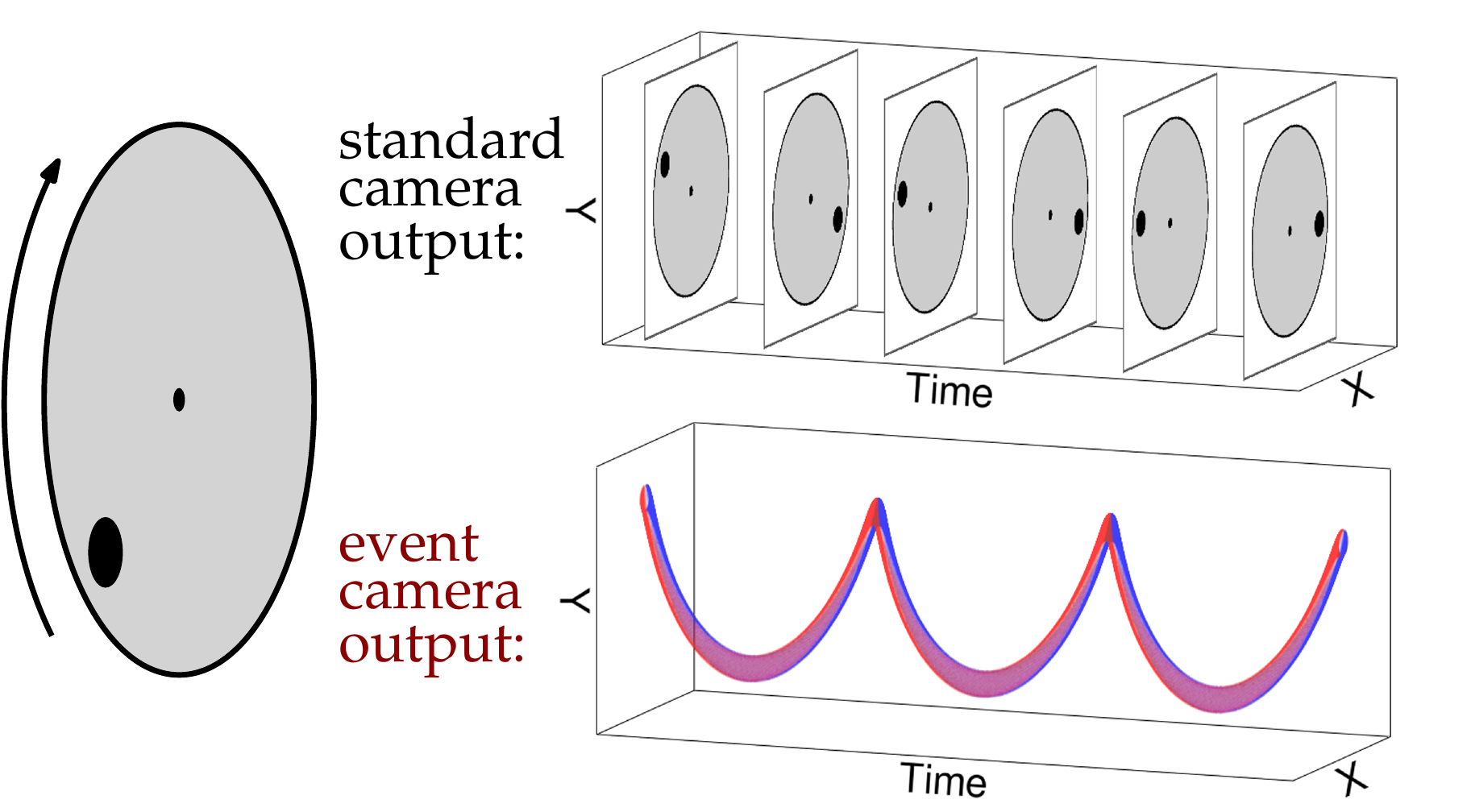}
\end{tabular}
\caption{Comparison of the output of a conventional camera and an event camera looking at a black disk on a rotating circle.
While a conventional camera captures frames at a fixed rate, an event camera transmits the brightness changes continuously in the form of a spiral of events in space-time (red: positive events, blue: negative events). Figure inspired by \cite{Mueggler14iros}.
}
\label{fig:principle_operation}
\end{figure}

An event camera consists of independent pixels that respond to changes in the spatio-temporal brightness signal $\Brightness(\xx,t)$\footnote{Event cameras respond in fact to logarithmic brightness changes, i.e. ${\Brightness=\log E}$ where $E$ is the \emph{irradiance}.} and transmit the changes in the form of a stream of asynchronous events~(Fig.~\ref{fig:principle_operation}).
For an ideal sensor, an event $\Event_i=(\Pixel_i,t_i,p_i)$ is triggered at pixel $\Pixel_i=(x_i,y_i)^T$ and time $t_i$ when the brightness change since the last event at the pixel reaches a threshold $\pm C$.
However, $C$ is in reality neither constant nor uniform across the image plane.
Rather, it strongly varies depending on factors such as the sign of the brightness change~\cite{Gallego17pami}, the event rate (because of limited pixel bandwidth)~\cite{Brandli14iscas}, and the temperature~\cite{Xu18ted}.
Consequently, events cannot by directly integrated to recover accurate intensity images in practice.

\subsection{Overview}

\begin{figure}
	\centering
	\begin{tabular}{c}
    \includegraphics[height=3.0cm]{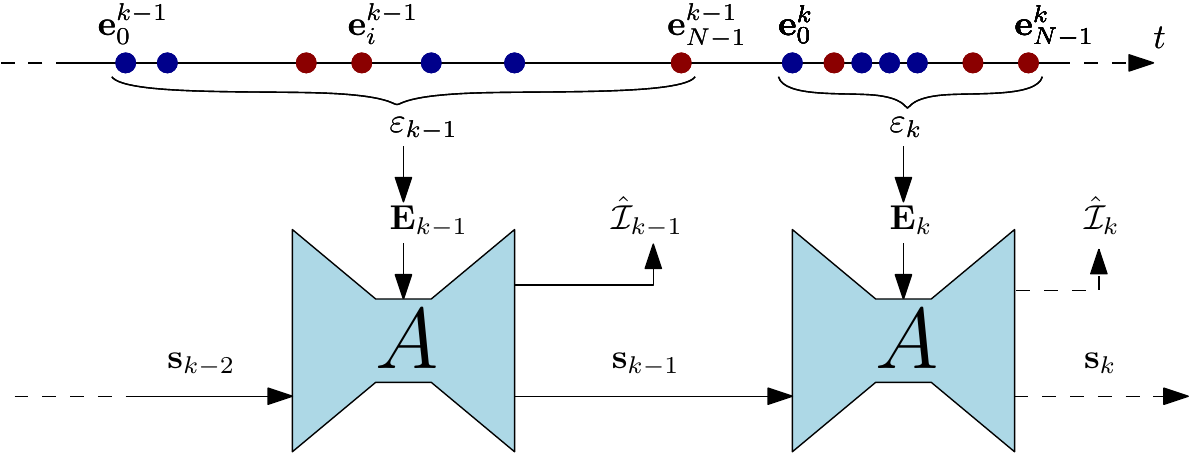}
    \end{tabular}
\caption{Overview of our approach. The event stream (depicted as red/blue dots on the time axis) is split into windows $\EventSequence_k$ containing multiple events.
Each window is converted into a 3D event tensor $\EventTensor_k$ and passed through the network, together with the previous state $\NetworkState_{k-1}$ to generate a new image reconstruction $\Reconstructed_k$  and updated state $\NetworkState_{k}$.
In this example, each window $\EventSequence_k$ contains a fixed number of events $\NumEvents=7$.
}
\label{fig:overview}
\end{figure}

Our goal is to translate a continuous stream of events
into a sequence of images $\lbrace \Reconstructed_k \rbrace$, where $\Reconstructed_k \in \left[0,1\right]^{W\timess H}$.
To achieve this, we partition the incoming stream of events into sequential (non-overlapping) spatio-temporal windows $\EventSequence_k=\left\lbrace \Event_i \right\rbrace$, for $i \in \left[ 0,\NumEvents-1 \right]$,
each containing a fixed number $\NumEvents$ of events.
The reconstruction function is implemented by a recurrent convolutional neural network, which maintains and updates an internal state $\NetworkState_k$ through time.
For each new event sequence $\EventSequence_k$, we generate a new image $\Reconstructed_k$ using the network state $\NetworkState_{k-1}$ (see~Fig.~\ref{fig:overview})
and update the state $\NetworkState_k$.
We train the network in supervised fashion, using a large amount of simulated event sequences with corresponding ground-truth images.

\subsection{Event Representation}
\label{sec:event_representation}

In order to be able to process the event stream using the convolutional recurrent network, we need to convert $\EventSequence_k$ into a fixed-size tensor representation $\EventTensor_k$.
A natural choice is to encode the events in a spatio-temporal voxel grid~\cite{Zhu18eccvw}.
The duration ${\Delta T=t^k_{N-1}-t^k_0}$ spanned by the events in $\EventSequence_k$ is discretized into $\NumBins$ temporal bins.
Every event distributes its polarity $p_i$ to the two closest spatio-temporal voxels as follows:
\begin{equation}
  \VoxelGrid(x_l,y_m,t_n) = \sum_{\substack{x_i = x_l\\y_i = y_m}}{p_i \max(0,1-|t_n-t^{*}_i|)},
\end{equation}
where
$t^{*}_i \triangleq \frac{\NumBins-1}{\Delta T}(t_i-t_0)$ is the normalized event timestamp.
We use $\NumBins=5$ temporal bins.

\subsection{Training Data}
\label{sec:training_data_generation}

Our network requires training data in the form of event sequences with corresponding ground-truth image sequences.
However, there exists no large-scale dataset with event data and corresponding ground-truth images.
Furthermore, images acquired by a conventional camera would provide poor ground truth in scenarios where event cameras excel, namely high dynamic range and high-speed scenes.
For these reasons, we propose to train the network on synthetic event data, and show subsequently (in Section~\ref{sec:evaluation}) that our network generalizes to real event data.

We use the event simulator ESIM \cite{Rebecq18corl}, which allows simulating a large amount of event data reliably.
ESIM renders images along the camera trajectory at high framerate, and interpolates the brightness signal at each pixel to approximate the continuous intensity signal needed to simulate an event camera.
Consequently, ground-truth images $\Image$ are readily available.
We map MS-COCO images~\cite{TsungYi14eccv} to a 3D plane and simulate the events triggered by random camera motion within this simple 3D scene.
Using \mbox{MS-COCO} images allows capturing a much larger variety of scenes than is available in any existing event camera dataset.
We set the camera sensor size to $240 \times 180$ pixels (to match the resolution of the DAVIS240C sensor used in our evaluation \cite{Brandli14ssc}). %
Note that inference can be performed at arbitrary resolutions since we will use a fully-convolutional network.
Examples of generated synthetic event sequences are presented in the supplement. %

We further enrich the training data by simulating a different set of positive and negative contrast thresholds for each simulated scene (sampled according to a normal distribution with mean $0.18$ and standard deviation $0.03$, values based on \cite{Kim14bmvc}).
This data augmentation prevents the network from learning to naively integrate events, which would work well on noise-free, simulated data, but would generalize poorly to real event data (for which the assumption of a fixed contrast threshold does not hold).

We generate $1@000$ sequences of 2 seconds each, which results in approximately 35 minutes of simulated event data. %
Note that the simulated sequences contain only globally homographic motion (i.e.\ there is no independent motion in the simulated sequences).
Nevertheless, our network generalizes surprisingly well to scenes with arbitrary motions, as will be shown in Sections~\ref{sec:evaluation}~and~\ref{sec:applications}.

\subsection{Network Architecture}
\label{sec:network_architecture}

\begin{figure*}
	\centering
    \includegraphics[width=\linewidth]{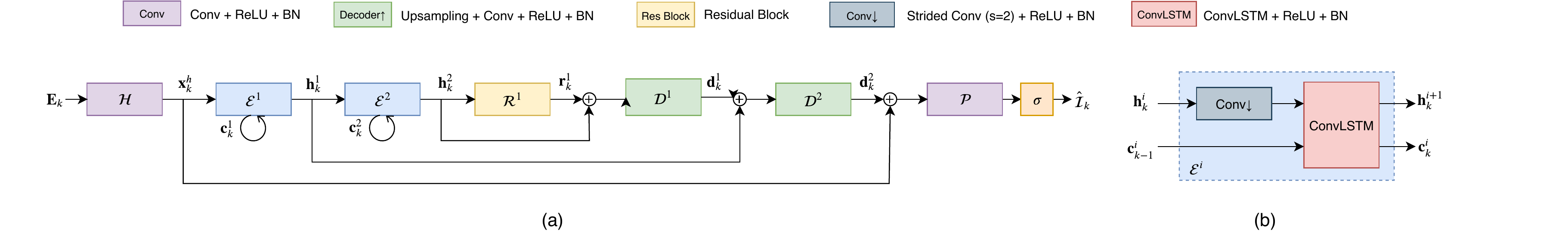}
\caption{We use a fully convolutional, UNet-like \cite{Ronneberger15icmicci} architecture (a), composed of $\NumEncoders$ recurrent encoder layers (b),
followed by $\NumResBlocks$ residual blocks and $\NumDecoders$ decoder layers, with skip connections between symmetric layers.
Encoders are composed of a strided convolution (stride $2$) followed by a \ConvLSTM~\cite{Shi15nips}.
Decoder blocks perform bilinear upsampling followed by a convolution.
ReLU activations and batch normalization \cite{Ioffe15icml} are used after each layer (except the last prediction layer, for which a sigmoid activation is used).
In this diagram, $\NumEncoders=2$ and $\NumResBlocks=1$.}
\label{fig:network_architecture}
\end{figure*}

Our neural network is a recurrent, fully convolutional network that was inspired by the UNet~\cite{Ronneberger15icmicci} architecture. An overview is shown in Fig.~\ref{fig:network_architecture}.
It is composed of a head layer ($\Head$), followed by $\NumEncoders$ recurrent encoder layers ($\Encoder^i$),
$\NumResBlocks$ residual blocks ($\ResBlock^j$),
$\NumDecoders$ decoder layers ($\Decoder^l$),
and a final image prediction layer ($\Prediction$).
Following \cite{Zhu18rss}, we use skip connections between symmetric encoder and decoder layers.
The number of output channels is $\NumBaseFeatureChannels$ for the head layer $\Head$, and is doubled after each encoder layer (thus, the final encoder has $\NumBaseFeatureChannels \times 2^{\NumEncoders}$ output channels).
The prediction layer performs a depthwise convolution (one output channel, kernel size $5$), followed by a sigmoid layer to produce an image prediction.
Encoder layers $\Encoder^i$ (Fig.~\ref{fig:network_architecture}(b)) consist of a 2D downsampling convolution (kernel size: $5$, stride: $2$) followed by a ConvLSTM \cite{Shi15nips}, with a kernel size of $3$,
 and whose number of input and hidden layers is the same as the preceding downsampling convolution.
Each encoder maintains a state $\EncoderState_k^i$ which is updated at every iteration, and initialized to zero at the first iteration ($k=0$).
The intermediate residual blocks \cite{He16cvpr} use a kernel size of $3$.
Each decoder layer consists of bilinear upsampling followed by a convolution with kernel size $5$.
Finally, we use the ReLU activation (for every layer except the final prediction) and batch normalization~\cite{Ioffe15icml}.

We used $\NumEncoders=3$, $\NumResBlocks=2$, $\NumBaseFeatureChannels=32$ and element-wise sum for the skip connection.
In Section~\ref{sec:architecture_search}, we motivate these choices of hyperparameters as the result of a search over multiple network architectures.

During training we unroll the network for $L$ steps.
We use $L=40$.
Note that this is in contrast to \cite{Rebecq19cvpr} which trains on significantly shorter event sequences ($L=8$).
The architecture in \cite{Rebecq19cvpr} is based on a vanilla recurrent (RNN) architecture, which suffers from vanishing gradients during back propagation through time on long sequences.
By contrast, our network uses stacked ConvLSTM gates which prevent these issues and allows us to train on longer sequences.
In Section~\ref{sec:network_analysis}, we show that our architecture based on LSTM improves the temporal stability of the network.

\vspace{0.2cm}
\subsection{Loss}
\label{sec:loss}

We use a combination of an image reconstruction loss and a temporal consistency loss.

\mypara{Image Reconstruction Loss.}
The image reconstruction loss ensures that the reconstructed image is similar to the target image.
While a direct pixel-wise loss such as the mean squared error (MSE) could be used, such losses are known to produce blurry images \cite{Johnson16eccv}.
Instead, we use a perceptual loss (specifically, the calibrated perceptual loss LPIPS \cite{Zhang18cvprLPIPS}).
The perceptual loss passes the reconstructed image and the target image through a VGG network \cite{Simonyan15iclr} that was trained on ImageNet~\cite{Russakovsky15ijcv}, and averages the distances between VGG features across multiple layers.
By minimizing LPIPS, our network effectively learns to endow the reconstructed images with natural statistics (i.e.\ with features close to those of natural images).
Our reconstruction loss is computed as $\ReconstructionLoss_k = d(\Reconstructed_k, \Image_k)$, where $d$ denotes the LPIPS distance \cite{Zhang18cvprLPIPS}.

\mypara{Temporal Consistency Loss.}
In our previous work \cite{Rebecq19cvpr}, the network relied on the recurrent connection to naturally enforce temporal consistency between successive reconstructions.
However, some temporal artifacts remained, notably some slight blinking that was especially noticeable in homogeneous image regions.
To address this issue, we introduce an explicit temporal consistency loss, the beneficial effect of which will be demonstrated in Section~\ref{sec:network_analysis}.
Our temporal consistency loss is based on \cite{Lai18eccv}.
Given optical flow maps $\OpticFlowMap{}_{k-1}^{k}$ between successive frames, the temporal loss is computed as the warping error
between two successive reconstructions:
\begin{equation}
    \label{eq:temporal_loss}
    \TemporalLoss_k = M_{k-1}^k \norm{\Reconstructed_k - \Warp_{k-1}^{k}(\Reconstructed_{k-1})}_1,
\end{equation}
where $\Warp_{k-1}^{k}(\Reconstructed_{k-1})$ is the result of warping the reconstruction $\Reconstructed_{k-1}$ to $\Reconstructed_k$ using the optical flow $\OpticFlowMap^k_{k-1}$,
and \hbox{$M_{k-1}^k = \exp (-\alpha \norm{\Image_k - \Warp_{k-1}^{k}(\Image_{k-1})}^2_2)$} is a weighting term that helps to mitigate the effect of occlusions (this term is small when the warping error in the ground truth
images $\Image_k$ is high, which happens predominantly at occlusions).
We set $\alpha=50$ in our experiments.

Note that the optical flow maps are only required at training time, but not at inference time.
The final loss is a weighted sum of the reconstruction and temporal losses:
\begin{equation}
    \Loss = \sum_{k=0}^{L}{\ReconstructionLoss_k} + \WeightTemporalLoss \sum_{k=L_0}^{L}{\TemporalLoss_k},
\end{equation}
where $\WeightTemporalLoss = 5$ (this value was chosen empirically to balance the range of values taken by both losses), and $L_0=2$ (the first few samples of each sequence are ignored in the computation of the temporal loss to leave time for the reconstruction to converge).

\subsection{Training Procedure}
\label{sec:training_modalities}

We split the synthetic sequences into $950$ training sequences and $50$ validation sequences.
The input event tensors are normalized such that the mean and standard deviation of the nonzero values in each tensor is $0$ and $1$, respectively.
We augment the training data using random 2D rotations (in the range of $\pm 20$ degrees), horizontal and vertical flips, and random cropping (with a crop size of $128 \times 128$).

We implement our network using PyTorch \cite{Paszke17nipsw} and use ADAM~\cite{Kingma15iclr} with a learning rate of $0.0001$.
We use a batch size of $2$ and train for $160$ epochs ($320@000$ iterations).%

\subsection{Post-processing}

While the sigmoid activation guarantees that the resulting image prediction $\Reconstructed$ takes values between $0$ and $1$,
we observe that the range of output values often does not span the entire range, i.e.\ the reconstructions can have low contrast.
To remedy this, we rescale the image intensities using robust min/max normalization to get a final reconstruction $\Reconstructed^f$:
\begin{equation}
    \Reconstructed^f = \frac{\Reconstructed - m}{M - m},
\end{equation}
where $m$ and $M$ are the 1\% and 99\% percentiles of $\Reconstructed$.
Finally, $\Reconstructed^f$ is clipped to the range $[0, 1]$.

\section{Evaluation}
\label{sec:evaluation}

\setlength{\tabcolsep}{0.3ex} %
\global\long\def\heightplot{2.1cm} %
\global\long\def\widthplot{2.805cm} %
\global\long\def\vspacecols{0.15ex} %
\begin{figure*}[t]
	\centering
    \begin{tabular}{cccccc}
    \includegraphics[width=\widthplot,height=\heightplot]{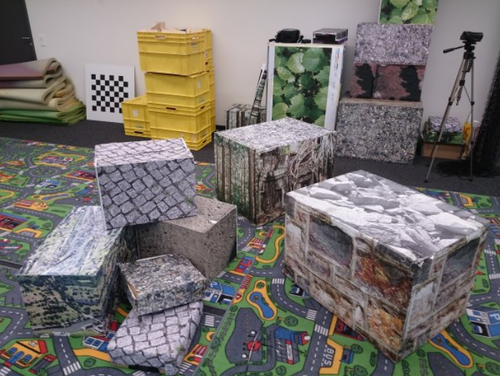}
    & \includegraphics[width=\widthplot,height=\heightplot]{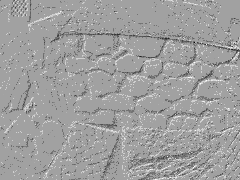}
    & \includegraphics[width=\widthplot,height=\heightplot]{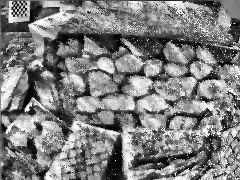}
    & \includegraphics[width=\widthplot,height=\heightplot]{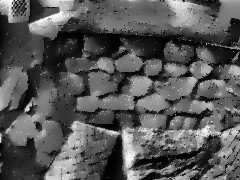}
    & \includegraphics[width=\widthplot,height=\heightplot]{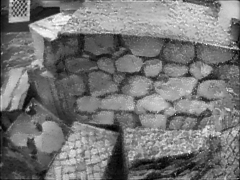}
    & \includegraphics[width=\widthplot,height=\heightplot]{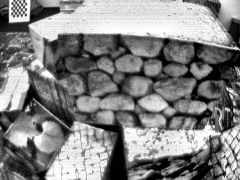}\\[\vspacecols]

    \includegraphics[width=\widthplot,height=\heightplot]{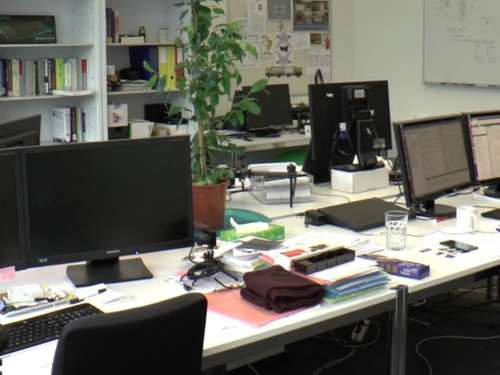}
    & \includegraphics[width=\widthplot,height=\heightplot]{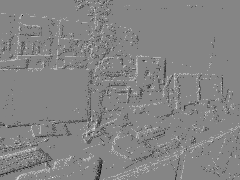}
    & \includegraphics[width=\widthplot,height=\heightplot]{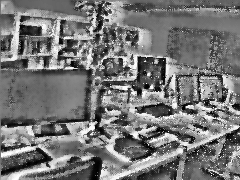}
    & \includegraphics[width=\widthplot,height=\heightplot]{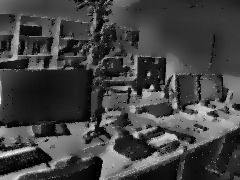}
    & \includegraphics[width=\widthplot,height=\heightplot]{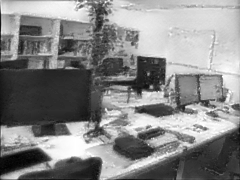}
    & \includegraphics[width=\widthplot,height=\heightplot]{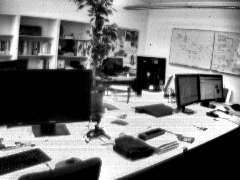}\\[\vspacecols]
    \includegraphics[width=\widthplot,height=\heightplot]{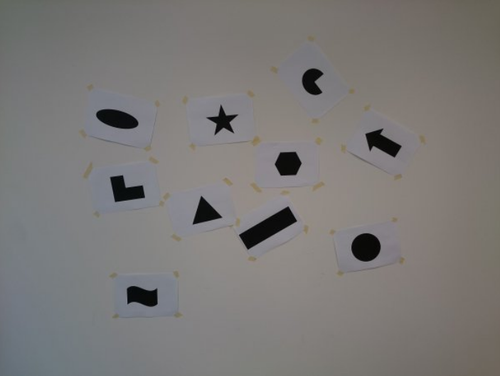}
    & \includegraphics[width=\widthplot,height=\heightplot]{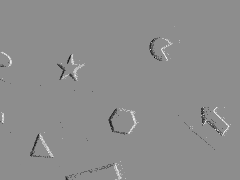}
    & \includegraphics[width=\widthplot,height=\heightplot]{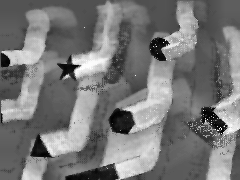}
    & \includegraphics[width=\widthplot,height=\heightplot]{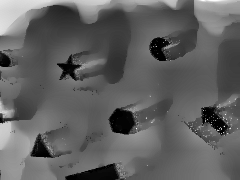}
    & \includegraphics[width=\widthplot,height=\heightplot]{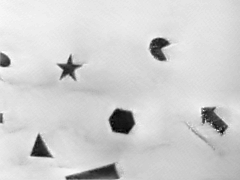}
    & \includegraphics[width=\widthplot,height=\heightplot]{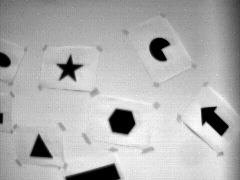}\\[\vspacecols]
    (a) Scene overview & (b) Events & (c) HF & (d) MR & (e) Ours & (f) Ground truth\\
    \end{tabular}
\caption{Comparison of our method with MR and HF on sequences from \cite{Mueggler17ijrr}. %
Our network is able to reconstruct fine details well (textures in the first row), while avoiding common artifacts (e.g. the ``bleeding edges'' in the third row).
}
\label{fig:comp_event_camera_dataset}
\end{figure*}

In this section, we present both quantitative and qualitative results on the fidelity of our reconstructions, and compare to recent methods \cite{Bardow16cvpr,Munda18ijcv,Scheerlinck18accv}.
We focus our evaluation on real event data. An evaluation on synthetic data can be found in supplementary material.

We use event sequences from the Event Camera Dataset~\cite{Mueggler17ijrr}. These sequences were recorded using a DAVIS240C sensor \cite{Brandli14ssc} moving in various environments.
It contains events as well as ground-truth grayscale frames at a rate of \SI{20}{\Hz}.
We remove redundant sequences (e.g.\ ones captured in the same scene) and those for which the frame quality is poor, leaving seven sequences in total that amount to $1@670$ ground-truth frames.
For each sequence, we reconstruct a video from the events with our method and each baseline.
For each ground-truth frame, we query the reconstructed image with the closest timestamp (tolerance of $\pm \SI{1}{\ms}$).

Each reconstruction is then compared to the corresponding ground-truth frame according to several quality metrics.
We apply local histogram equalization to every ground-truth frame and reconstructed frame prior to computing the error metrics (this way the intensity values lie in the same intensity range and are thus comparable).
Note that the camera speed gradually increases in each sequence, leading to significant motion blur on the ground-truth frames towards the end of the sequences; we therefore exclude these fast sections in our quantitative evaluation.
We also omit the first few seconds from each sequence, which leaves enough time for the baseline methods that are based on event integration to converge.
Note that this works in favor of the baselines, as our method converges almost immediately (more details in Section~\ref{sec:network_analysis}).

We compare our approach against several state-of-the-art methods: \cite{Bardow16cvpr} (which we denote as SOFIE for ``Simultaneous Optic Flow and Intensity Estimation''), \cite{Scheerlinck18accv} (HF for ``High-pass Filter''), and \cite{Munda18ijcv} (MR for ``Manifold Regularization''),
both in terms of image reconstruction quality and temporal consistency.
For HF and MR, we used the code that was provided by the authors and manually tuned the parameters on the evaluated sequences to get the best results possible.
For HF, we also applied a bilateral filter to the reconstructed images (with filter size $d=5$ and $\sigma=25$) in order to remove high-frequency noise, which improves the results of HF in all metrics.
For SOFIE, we report qualitative results instead of quantitative results since we were not able to obtain satisfying reconstructions on our datasets using the code provided by the authors.
We report three image quality metrics: mean squared error (MSE; lower is better), structural similarity (SSIM; higher is better)~\cite{Wang04tip}, and the calibrated perceptual loss (LPIPS; lower is better)~\cite{Zhang18cvprLPIPS}.
In addition, we measure the temporal consistency of the reconstructed videos using the temporal loss introduced in Eq.~\eqref{eq:temporal_loss}.
Note that computing the temporal loss requires optical flow maps between successive DAVIS frames, which we obtain with FlowNet2 \cite{Ilg17cvpr}.

\mypara{Results and Discussion.} The main quantitative results are presented in Table~\ref{tab:image_quality_comparison}, and are supported by qualitative results in Figs.~\ref{fig:comp_event_camera_dataset} and~\ref{fig:comp_bardow}.
Additional results are available in the supplementary material. %
We also encourage the reader to watch the supplementary video, which conveys these results better than still images.

Our reconstruction method outperforms the state of the art by a large margin, with an average 24\% increase in SSIM and a 22\% decrease in LPIPS.
Qualitatively, our method reconstructs small details remarkably well compared to the baselines (see the boxes in the first row of Fig.~\ref{fig:comp_event_camera_dataset}, for example).
Furthermore, our method does not suffer from ``ghosting'' or ``bleeding edges'' artifacts that are present in other methods (particularly visible in the third row of Fig.~\ref{fig:comp_event_camera_dataset}).
These artifacts result from (i) incorrectly estimated contrast thresholds and (ii) the fact that these methods can only estimate the image intensity up to some unknown initial intensity $\Image_0$, the ghost of which can remain visible.
We also compare our method to HF, MR, and SOFIE qualitatively using datasets and image reconstructions directly provided by the authors of \cite{Bardow16cvpr}, in Fig.~\ref{fig:comp_bardow}.
Once again, our network generates higher quality reconstructions, with finer details and less noise.
In Section~\ref{sec:applications}, we provide many more qualitative reconstruction results,
and in particular show that our network is able to leverage the outstanding properties of events to reconstruct images in high-speed and high dynamic range scenarios.

\setlength{\tabcolsep}{0.13ex} %
\global\long\def\heightplot{1.60cm} %
\global\long\def\widthplot{1.60cm} %
\global\long\def\vspacecols{0.2ex} %
\begin{figure}
	\centering
    \begin{tabular}{ccccc}
    \includegraphics[width=\widthplot,height=\heightplot]{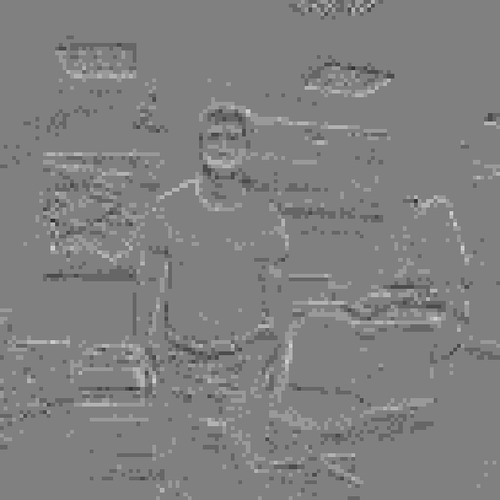}
    & \includegraphics[width=\widthplot,height=\heightplot]{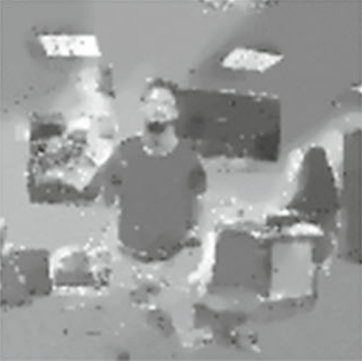}
    & \includegraphics[width=\widthplot,height=\heightplot]{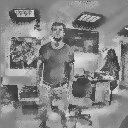}
    & \includegraphics[width=\widthplot,height=\heightplot]{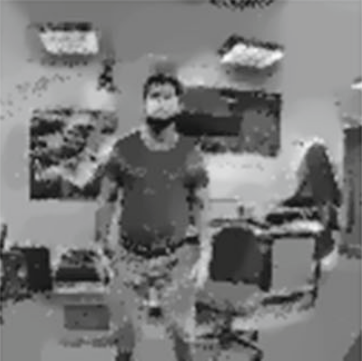}
    & \includegraphics[width=\widthplot,height=\heightplot]{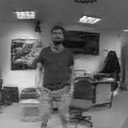}\\
    \includegraphics[width=\widthplot,height=\heightplot]{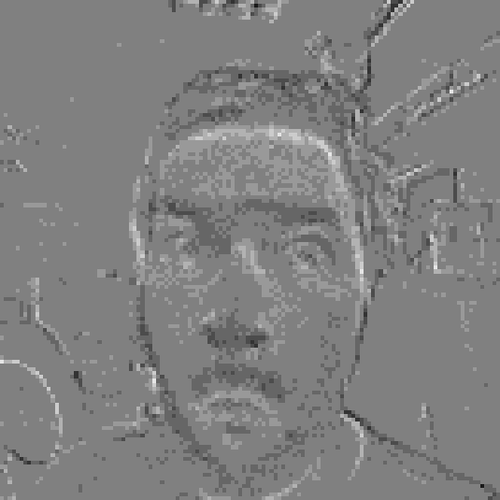}
    & \includegraphics[width=\widthplot,height=\heightplot]{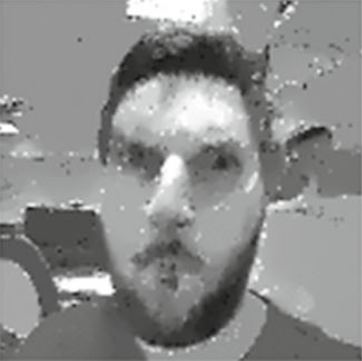}
    & \includegraphics[width=\widthplot,height=\heightplot]{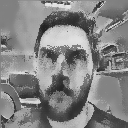}
    & \includegraphics[width=\widthplot,height=\heightplot]{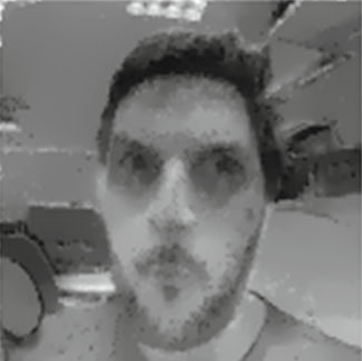}
    & \includegraphics[width=\widthplot,height=\heightplot]{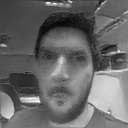}\\
    \footnotesize (a) Events & \footnotesize (b) SOFIE & \footnotesize (c) HF & \footnotesize (d) MR & \footnotesize (e) Ours\\
    \end{tabular}
\caption{Qualitative comparison on the dataset introduced by \cite{Bardow16cvpr}. Our method produces cleaner and more detailed results.
}
\label{fig:comp_bardow}
\end{figure}
\begin{table*}
\caption{Comparison to state-of-the-art image reconstruction methods on the Event Camera Dataset \cite{Mueggler17ijrr}.
Our approach outperforms prior methods on almost all datasets and metrics by a large margin, with an average 24\% increase
in structural similarity (SSIM) and a 22\% decrease in perceptual distance (LPIPS) compared to the best prior methods.}
\label{tab:image_quality_comparison}
\newcolumntype{Z}{S[table-format=2.2,table-auto-round]}
\centering
\setlength{\tabcolsep}{3mm}
\ra{1.05}
\small
\begin{tabular}{@{}lcZZZcZZZcZZZcr@{}}
  \toprule
  \multirow{2}[3]{*}{Dataset} && \multicolumn{3}{c}{MSE}  &&  \multicolumn{3}{c}{SSIM}  &&\multicolumn{3}{c}{LPIPS} \\
  \cmidrule(l{3mm}r{3mm}){3-5} \cmidrule(l{3mm}r{3mm}){7-9} \cmidrule(l{3mm}r{3mm}){11-13}
  && {HF} & {MR} & {Ours} && {HF} & {MR} & {Ours} && {HF} & {MR} & {Ours}  \\
  \midrule

dynamic\_6dof && 0.0997 & \bfseries 0.0506 & 0.1381 && 0.3944 & \bfseries 0.5227 & 0.4590 && 0.5416 & 0.4953 & \bfseries 0.4551 \\
boxes\_6dof && 0.0785 & 0.0993 & \bfseries 0.0364 && 0.4874 & 0.4532 & \bfseries 0.6203 && 0.5006 & 0.5345 & \bfseries 0.3809 \\
poster\_6dof && 0.0684 & \bfseries 0.0453 & 0.0564 && 0.4925 & 0.5377 & \bfseries 0.6192 && 0.4485 & 0.5202 & \bfseries 0.3462 \\
shapes\_6dof && 0.0885 & 0.1921 & \bfseries 0.0352 && 0.5048 & 0.5108 & \bfseries 0.7950 && 0.6066 & 0.6437 & \bfseries 0.4731 \\
office\_zigzag && 0.0895 & 0.0869 & \bfseries 0.0295 && 0.3846 & 0.4464 & \bfseries 0.5425 && 0.5364 & 0.4967 & \bfseries 0.4086 \\
slider\_depth && 0.0646 & 0.0702 & \bfseries 0.0524 && 0.4991 & 0.5002 & \bfseries 0.5776 && 0.5036 & 0.5493 & \bfseries 0.4377 \\
calibration && 0.0912 & 0.0700 & \bfseries 0.0171 && 0.4837 & 0.5357 & \bfseries 0.6965 && 0.4840 & 0.4711 & \bfseries 0.3629 \\
\midrule
Mean && 0.0829 & 0.0878 & \bfseries 0.0522 && 0.4638 & 0.5009 & \bfseries 0.6157 && 0.5173 & 0.5301 & \bfseries 0.4092 \\

   \bottomrule
\end{tabular}
\end{table*}

Finally, Table~\ref{tab:temporal_consistency_comparison} shows the temporal error (lower error means higher temporal consistency) for all methods, as well as for the ground-truth sequences for reference.
Note that the temporal loss is greater than zero on the ground-truth sequences because of small errors in optical flow estimation and occlusions.
(It is still significantly lower than for all reconstruction methods.)
Our method outperforms the competing approaches in terms of temporal consistency.
We attribute this mostly to the temporal loss introduced in Section~\ref{sec:loss}.
In Section~\ref{sec:network_analysis}, we evaluate the effect of the temporal loss in an ablation study. %

\begin{table}
\caption{
  Comparison of the temporal error (Eq.~\eqref{eq:temporal_loss}, lower is better) between HF, MR and our method.
  Our video reconstructions have higher temporal consistency than the baselines.
  }
  \label{tab:temporal_consistency_comparison}
\newcolumntype{Z}{S[table-format=2.2,table-auto-round]}
\centering
\setlength{\tabcolsep}{2.5mm}
\ra{1.05}
\small
\begin{tabular}{@{}lcZZZZcr@{}}
  \toprule
  \multirow{2}[4]{*}{Dataset} &&\multicolumn{4}{c}{Temporal Error} \\
  \cmidrule(l{3mm}r{3mm}){3-6}
  && {HF} & {MR} & {Ours} & {Ground truth} \\
  \midrule

dynamic\_6dof && 3.3246 & 1.9071 & \bfseries 1.6380 & 0.5251 \\
boxes\_6dof && 3.3709 & 1.7922 & \bfseries 1.3215 & 0.5879 \\
poster\_6dof && 3.6284 & 2.1489 & \bfseries 1.7718 & 0.5663 \\
shapes\_6dof && 3.4954 & 1.8020 & \bfseries 1.4822 & 0.8932 \\
office\_zigzag && 3.1834 & 1.5831 & \bfseries 1.3769 & 0.4846 \\
slider\_depth && 2.1427 & 1.6192 & \bfseries 1.3669 & 0.7043 \\
calibration && 2.7240 & 1.5237 & \bfseries 1.0224 & 0.6224 \\
\midrule
Mean && 3.1242 & 1.7680 & \bfseries 1.4257 & 0.6263 \\

   \bottomrule
\end{tabular}
\end{table}

\section{Applications}
\label{sec:applications}

We now present some applications of our method.
We synthesize high framerate videos of fast physical phenomena (Section~\ref{sec:highspeed_reconstructions}), high dynamic range videos (Section~\ref{sec:hdr_reconstructions}), and color video (Section~\ref{sec:color_reconstruction}).
Finally, we evaluate the effectiveness of our reconstructions as an intermediate representation that enables direct application of conventional vision algorithms to event data (Section~\ref{sec:downstream_applications}).

\subsection{High Speed Video Reconstruction}
\label{sec:highspeed_reconstructions}

Event cameras have a higher temporal resolution than conventional sensors (\SI{}{\SIUnitSymbolMicro s} vs \SI{}{\ms}). %
In addition, the event stream is sparse by nature, which saves bandwidth. %
We now show that our method can decompress the event stream to reconstruct videos of fast motions with high framerate (thousands of frames per second).

\mypara{Datasets.}
Since there exists no public event dataset containing fast physical phenomena, we recorded our own.
We used the Samsung DVS Gen3 sensor \cite{Son17isscc}, with VGA resolution. %
We recorded four sequences at daytime under bright sunlight (Fig.~\ref{fig:high_speed_reconstructions}).
The first two feature objects (plaster garden gnome, ceramic mug) being shot with a rifle (approximate muzzle velocity: 376 m/s).
The last two sequences feature two balloons (filled with water and air, respectively) being popped with a needle.
In order to reconstruct the background, each sequence starts with the camera being moved slightly, after which it is kept steady.
We additionally recorded the first two sequences with a high-end mobile phone camera (Huawei P20 Pro) operating at 240 FPS\footnote{While the Huawei P20 Pro can in principle record at 960 FPS,
it can only do so for a very short amount of time (\SI{0.2}{\second}), which made a synchronized recording impractical.}.

\mypara{High Framerate Video Synthesis from Events.}
We used a fixed number of events $\NumEvents \simeq 10^4$ per window (exact values in Fig.~\ref{fig:high_speed_reconstructions_framerates}), resulting in video reconstructions at only a few hundred FPS.
While it would be possible to retrain the network with smaller window sizes to address the specific case of extremely high framerate video synthesis, it is in fact possible to arbitrarily increase the output framerate without retraining.
To achieve that, we run multiple reconstructions in parallel, introducing a slight temporal shift (of $D$ events) between each.
We thus obtain a set of videos with different temporal offsets, then merged by reordering the frames from the individual videos.
This yields a video with an arbitrarily high framerate, reaching multiple thousand FPS (Fig.~\ref{fig:high_speed_reconstructions_framerates}).
This temporal upsampling process may introduce some slight flickering, which is easily reduced using a simple filter \cite{ReduceFlicker}.

\mypara{Results and Discussion.}
In the supplementary video, we show the synthesized, high framerate videos and compare them with the 240 FPS reference videos.
Fig.~\ref{fig:high_speed_reconstructions} shows a few still frames from each sequence to convey the motion.
At these extreme speeds, the event sensor is pushed to its limits: the events suffer from high noise and have many artefacts (e.g.\ readout artefacts). %
Nonetheless, our network performs well, revealing details that are invisible to the naked eye or a consumer camera.
The output framerate (which varies with the event rate) is shown in Fig.~\ref{fig:high_speed_reconstructions_framerates}, and consistently stays in the range of thousands of FPS: at least an order of magnitude above conventional consumer cameras.

\global\long\def\heightplot{2.55cm} %
\global\long\def\widthplot{3.4cm} %
\begin{figure*}
	\centering
    \begin{tabular}{ccccc}

    \rotatebox{90}{Gnome Shooting}
    \scalebox{-1}[1]{\includegraphics[width=\widthplot,height=\heightplot]{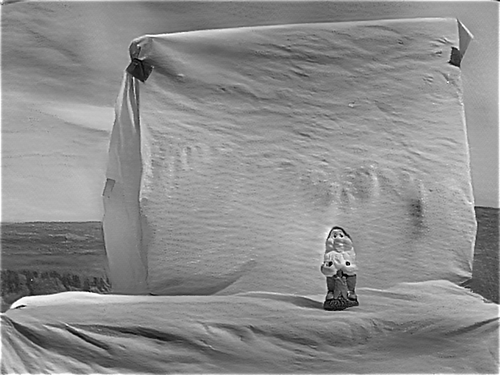}}
    & \scalebox{-1}[1]{\includegraphics[width=\widthplot,height=\heightplot]{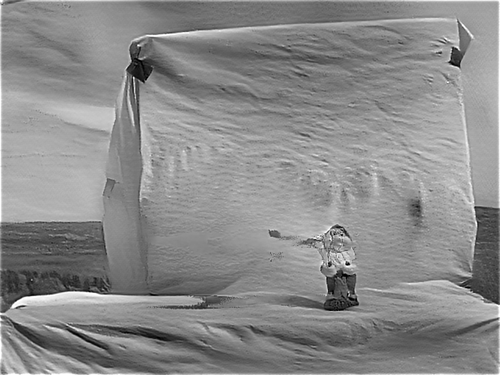}}
    & \scalebox{-1}[1]{\includegraphics[width=\widthplot,height=\heightplot]{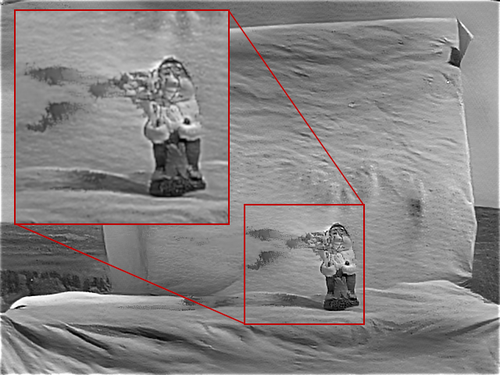}}
    & \scalebox{-1}[1]{\includegraphics[width=\widthplot,height=\heightplot]{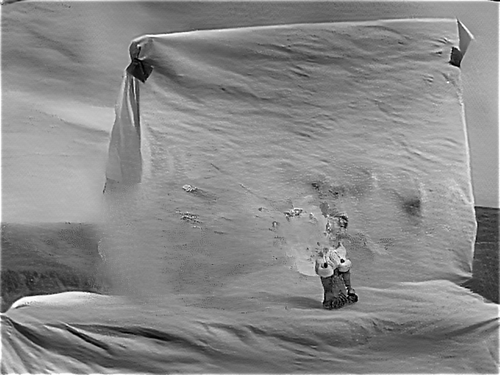}}
    & \scalebox{-1}[1]{\includegraphics[width=\widthplot,height=\heightplot]{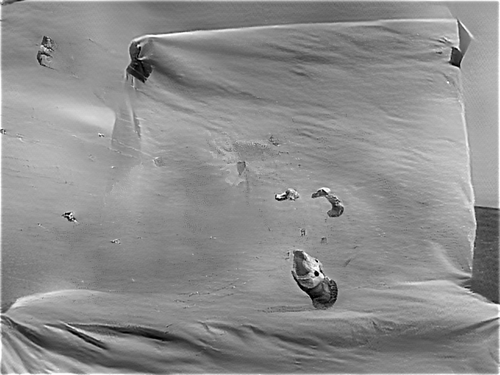}}\\

    \rotatebox{90}{Mug Shooting}
    \scalebox{-1}[1]{\includegraphics[width=\widthplot,height=\heightplot]{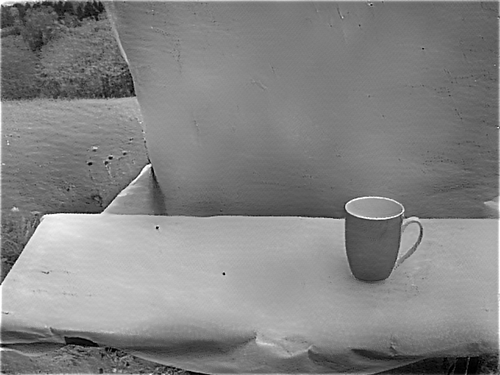}}
    & \scalebox{-1}[1]{\includegraphics[width=\widthplot,height=\heightplot]{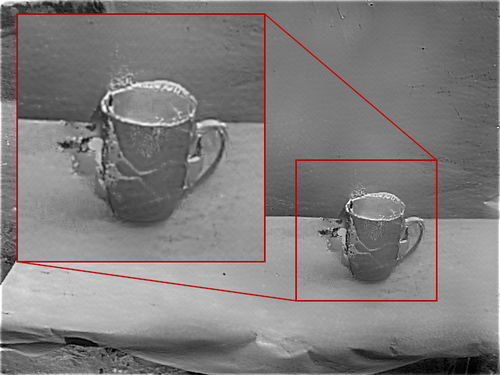}}
    & \scalebox{-1}[1]{\includegraphics[width=\widthplot,height=\heightplot]{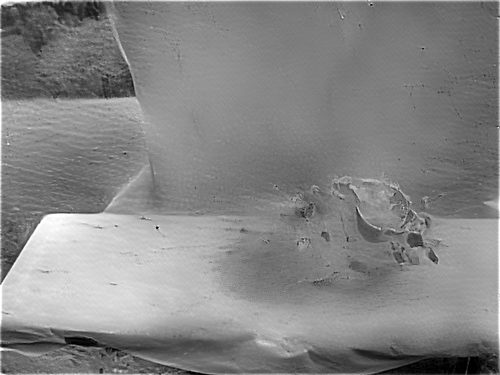}}
    & \scalebox{-1}[1]{\includegraphics[width=\widthplot,height=\heightplot]{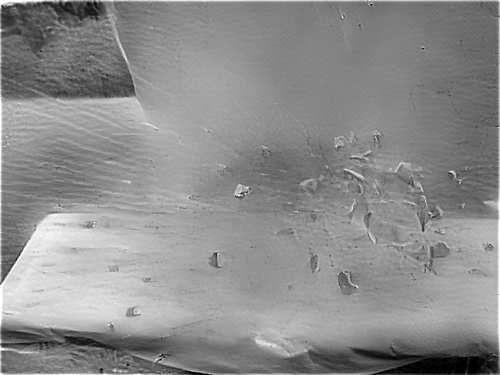}}
    & \scalebox{-1}[1]{\includegraphics[width=\widthplot,height=\heightplot]{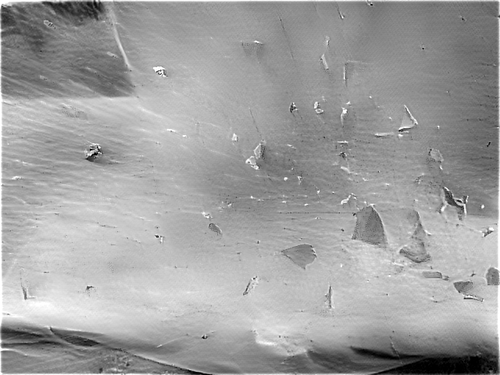}}\\

    \rotatebox{90}{Water Balloon}
    \includegraphics[trim={3.9cm 0cm 0cm 2.5cm},clip,width=\widthplot,height=\heightplot]{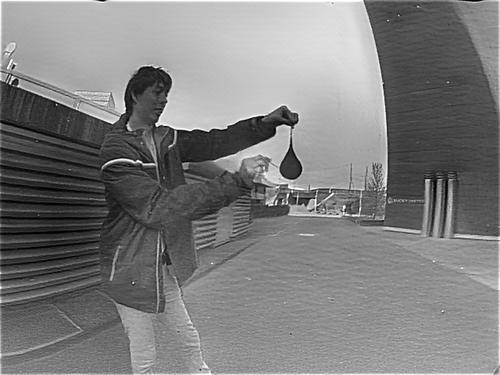}
    & \includegraphics[trim={0.95cm 0cm 0cm 0.61cm},clip,width=\widthplot,height=\heightplot]{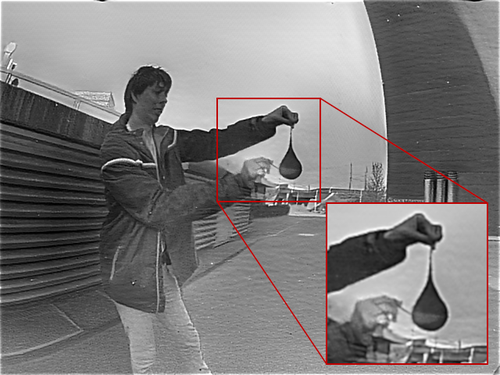}
    & \includegraphics[trim={3.9cm 0cm 0cm 2.5cm},clip,width=\widthplot,height=\heightplot]{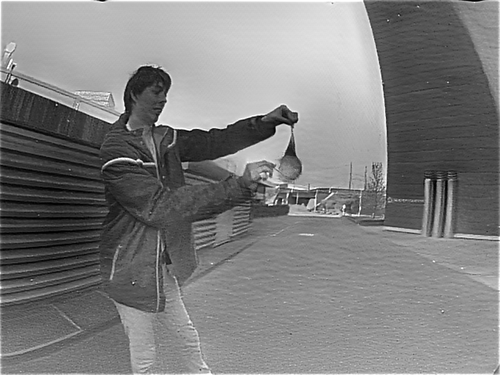}
    & \includegraphics[trim={3.9cm 0cm 0cm 2.5cm},clip,width=\widthplot,height=\heightplot]{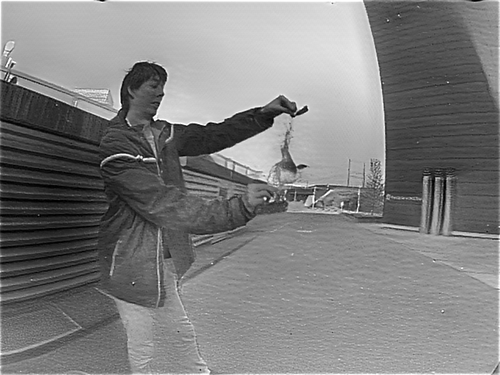}
    & \includegraphics[trim={3.9cm 0cm 0cm 2.5cm},clip,width=\widthplot,height=\heightplot]{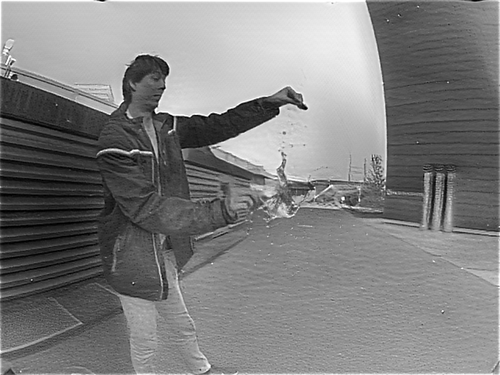}\\

    \rotatebox{90}{Air Balloon}
    \includegraphics[width=\widthplot,height=\heightplot]{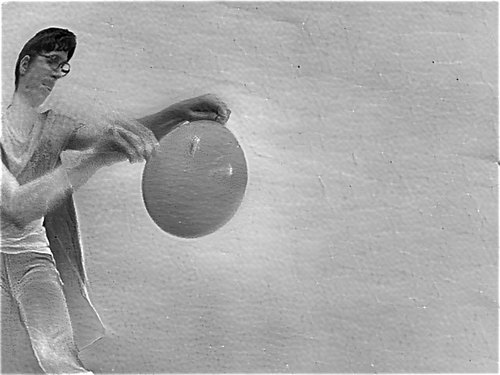}
    & \includegraphics[width=\widthplot,height=\heightplot]{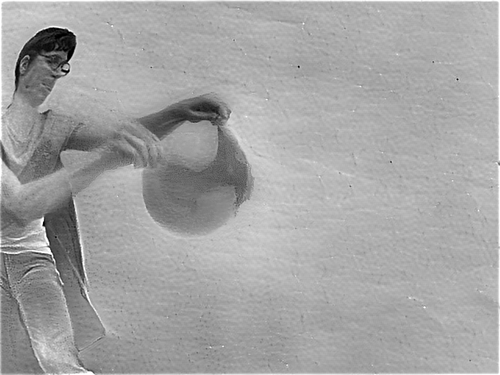}
    & \includegraphics[width=\widthplot,height=\heightplot]{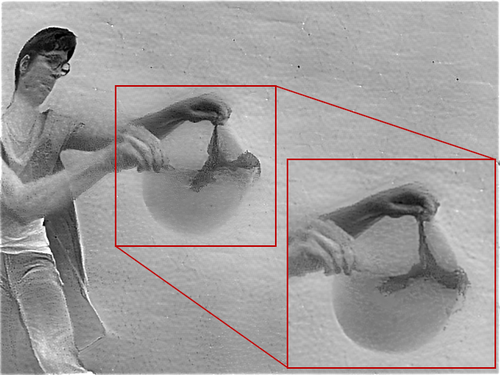}
    & \includegraphics[width=\widthplot,height=\heightplot]{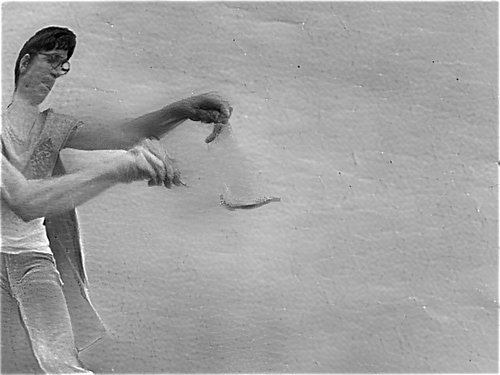}
    & \includegraphics[width=\widthplot,height=\heightplot]{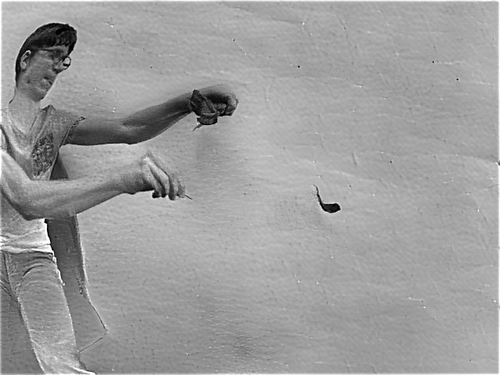}\\
    \end{tabular}

\caption{Video reconstructions of high speed physical phenomena, synthesized at $>5@000$ FPS with our approach.
First two rows: shooting a garden gnome and a mug with a rifle.
Last two rows: popping a water balloon and an air balloon with a needle.
Our reconstructions reveal details invisible to the naked eye or a conventional consumer camera.
In the first two rows, the trace of the bullet is clearly visible (the bullet itself was too fast for the event sensor to catch),
and cracks in both objects are visible before the pieces fly apart.
In the last two rows, the membrane of the balloons contracting away from the point where the needle hit is clearly visible.
}
\label{fig:high_speed_reconstructions}
\end{figure*}

\global\long\def\widthplot{0.18\linewidth} %
\setlength{\tabcolsep}{0.5cm} %
\begin{figure*}
	\centering
    \begin{tabular}{cccc}
    \includegraphics[width=\widthplot]{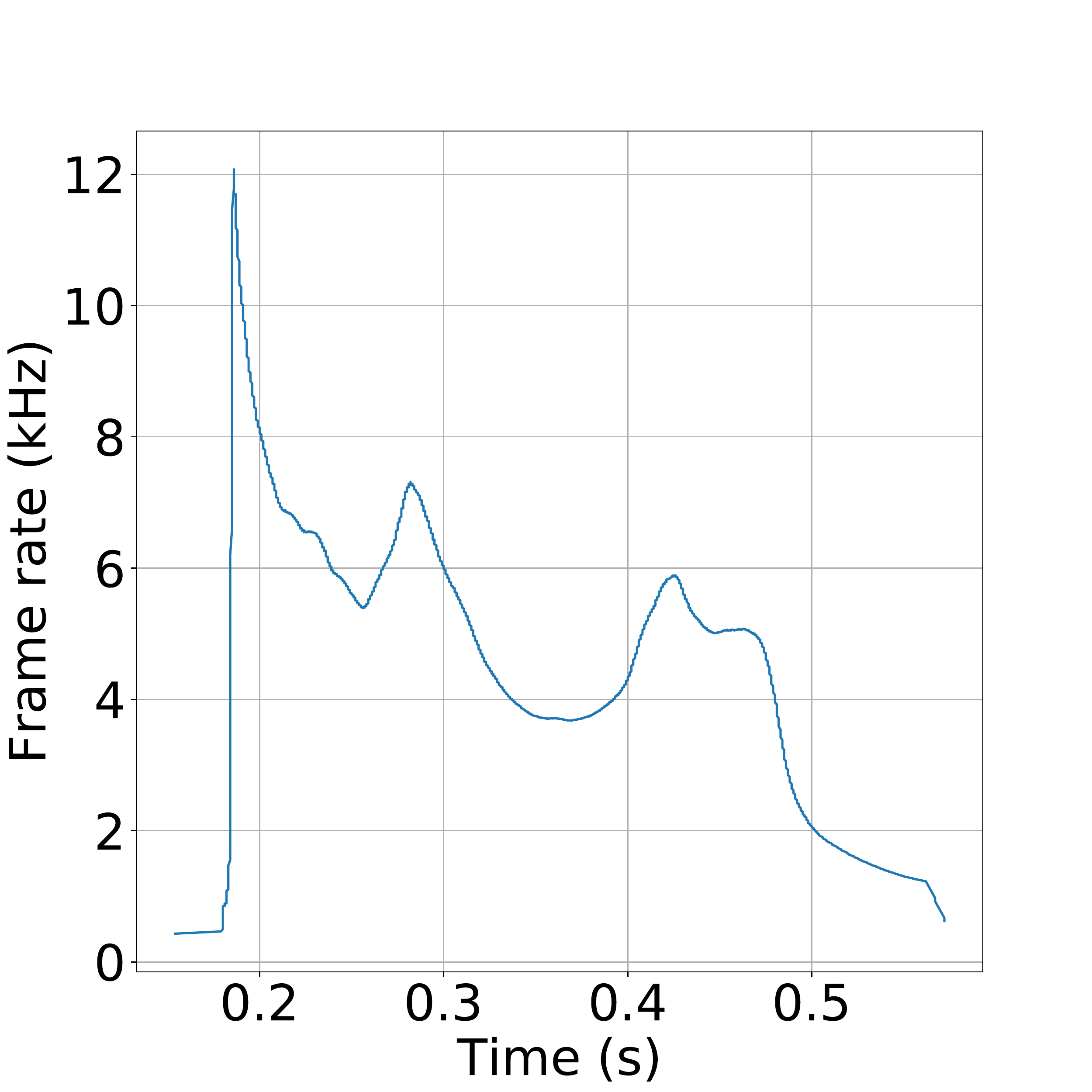}
    & \includegraphics[width=\widthplot]{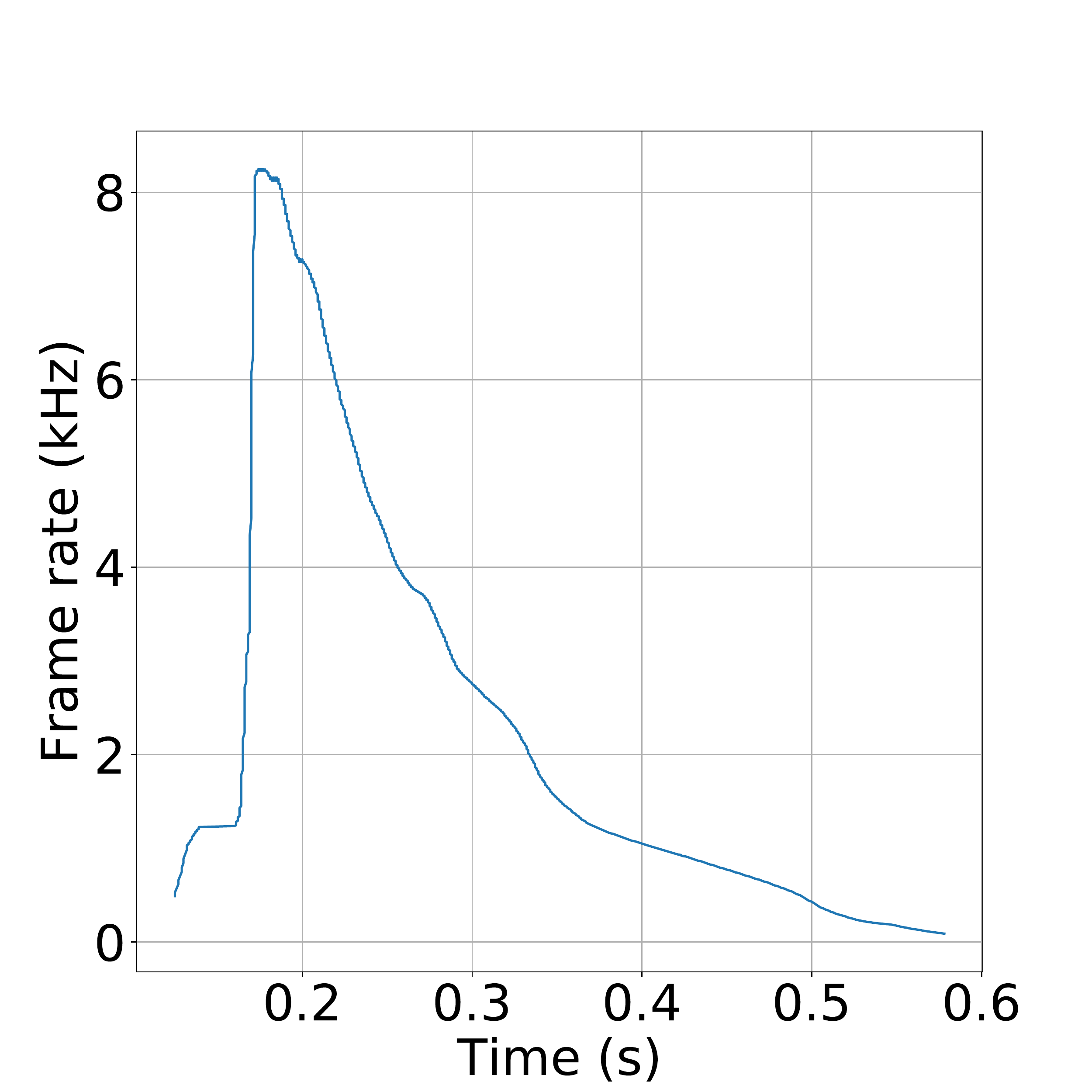}
    & \includegraphics[width=\widthplot]{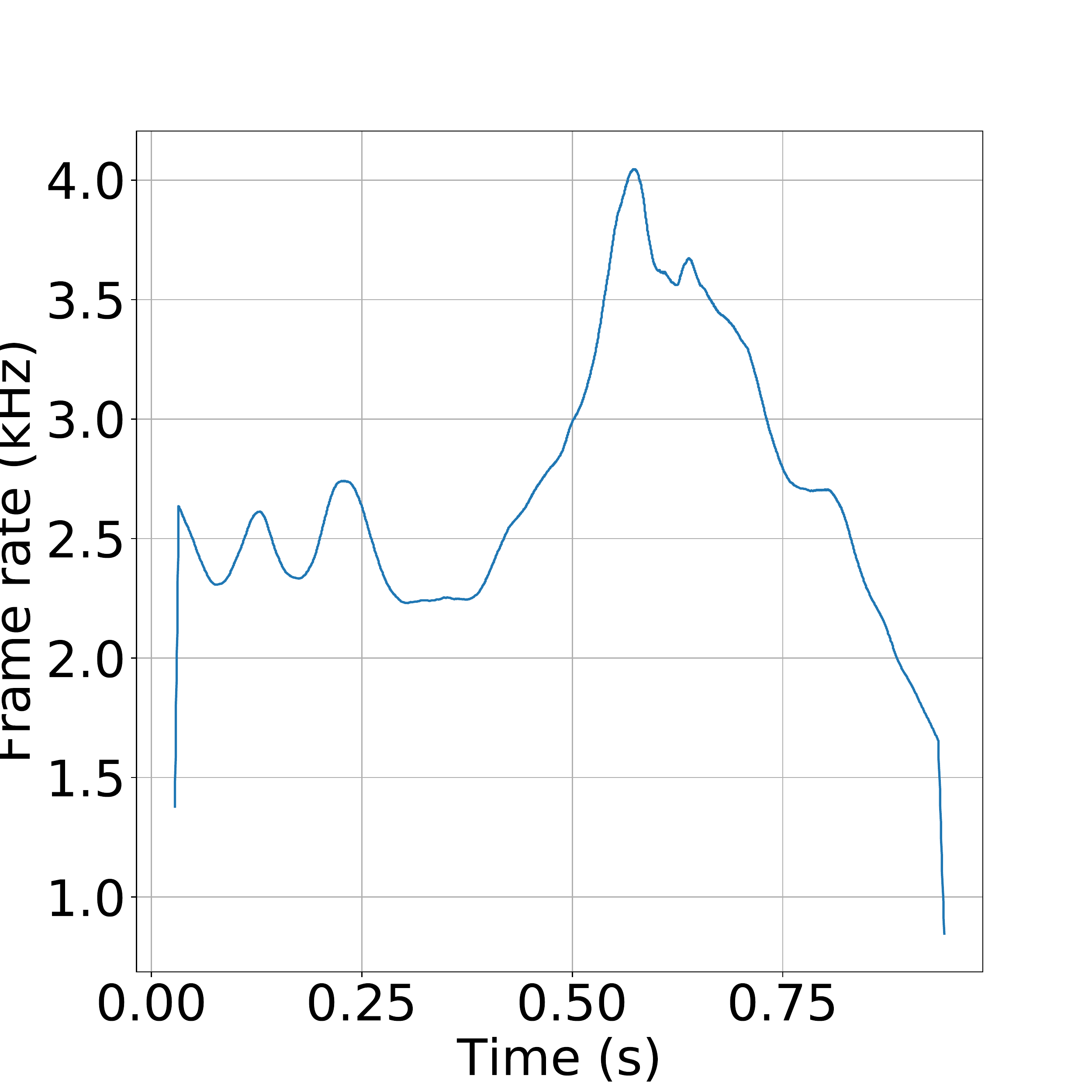}
    & \includegraphics[width=\widthplot]{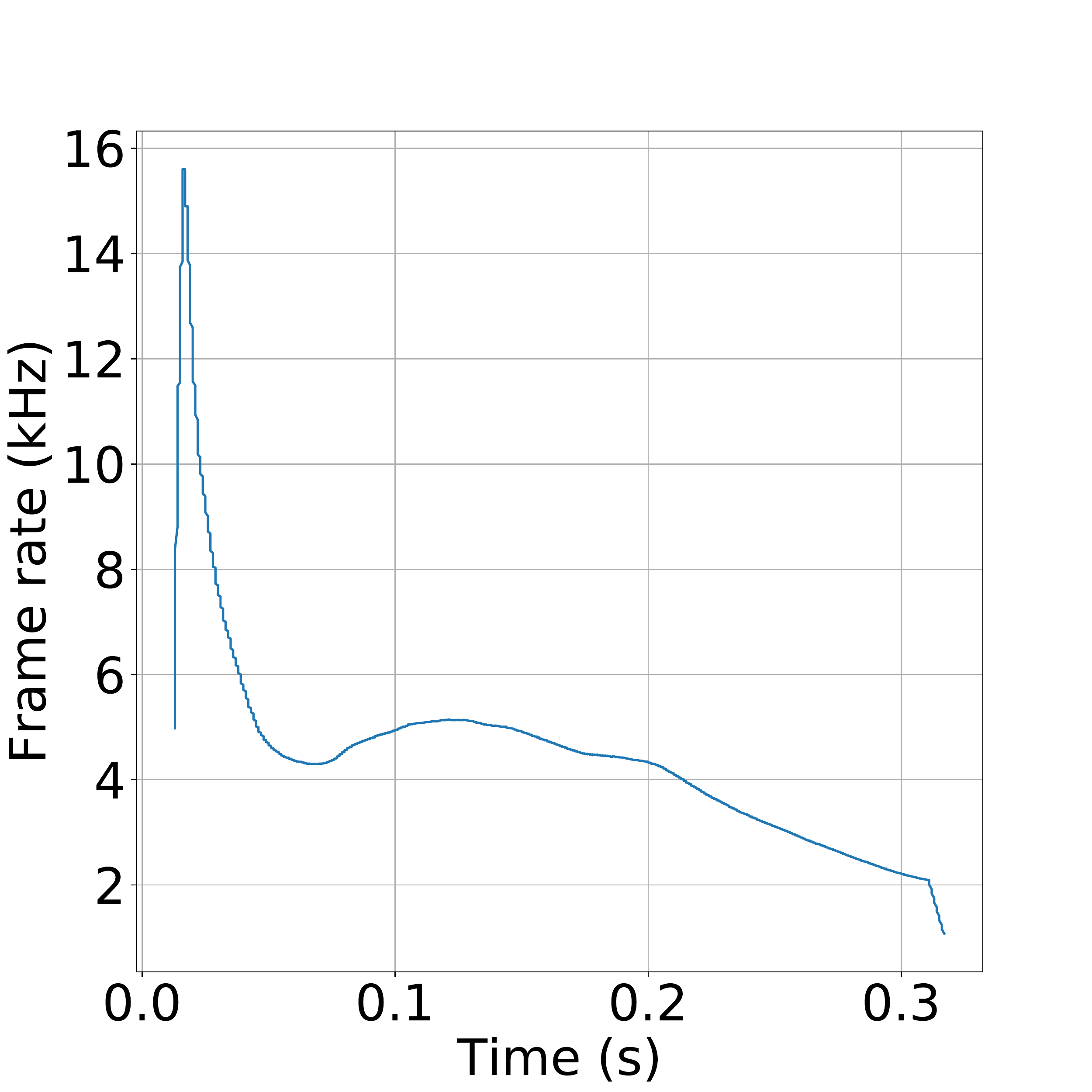}\\

    (a) Gnome & (b) Mug & (c) Water balloon  & (d) Air balloon \\
    \small $\NumEvents=10@000$, $D=500$ &  \small $\NumEvents=12@000$, $D=1000$ & \small $\NumEvents=12@000$, $D=1000$ & \small $N=20@000$, $D=500$\\

    \end{tabular}

\caption{Reconstruction framerate for the high-speed sequences. %
The output framerate grows with the event rate and value of $D$, and varies between 1 kHz and 15 kHz.}
\label{fig:high_speed_reconstructions_framerates}
\end{figure*}

\subsection{High Dynamic Range Reconstruction}
\label{sec:hdr_reconstructions}

Event cameras react to changes in log intensity \cite{Lichtsteiner08ssc}, which endows them with much higher dynamic range than conventional cameras (\SI{140}{\decibel} versus \SI{60}{\decibel}).
The videos that our method synthesizes preserve the high dynamic range of the events. %
In Fig.~\ref{fig:hdr_reconstructions}, we support this claim by showing qualitative reconstruction results in a variety of challenging HDR scenes and compare our reconstructions to the corresponding frame from a conventional camera.

\mypara{Datasets.}
There are many publicly available event camera datasets featuring HDR scenes~\cite{Mueggler17ijrr,Zhu18ral,Scheerlinck18accv}.
However, the reference images in these datasets were taken by the DAVIS sensor~\cite{Brandli14ssc}, the quality of which falls short of state-of-the-art consumer cameras.
To support a fair and up-to-date comparison, we recorded our own sequences in HDR scenes (indoors and outdoors), using a recent event camera (Samsung DVS Gen3) and a high-end smartphone for reference (Huawei P20 Pro).
The two sensors were rigidly mounted to each other for recording, and the corresponding footage was geometrically and temporally aligned manually in post processing.

\mypara{Results and Discussion.}
A comparison of our reconstruction results from event data and the frames from the conventional camera is presented in Fig.~\ref{fig:hdr_reconstructions}. %
The phone camera provides color images, which we converted to grayscale for easier visual comparison with our reconstructions.
The first row of Fig.~\ref{fig:hdr_reconstructions} shows a ``selfie'' sequence, recorded indoors with the sensors hand-held.
While the window behind the main subject appears severely overexposed in the conventional frame (b), the events (a) capture the full dynamic range, which our network successfully leverages to reconstruct the entire scene.
In addition, because of the camera shaking induced by the hand-held motion, the phone frame suffers from motion blur, which is not present in our reconstruction.
The second row shows a driving sequence, recorded with both sensors placed on the windshield of a car driving out of a tunnel.
Once again, the area outside of the tunnel is saturated in the conventional frame, while the events capture details both indoors and outdoors, which our reconstructions recover.
Finally, the third row shows an outdoor example recorded with the sensors pointing directly at the sun on a bright day.
In this extreme case, the events suffer from unusually high levels of noise (flickering events), which cause reconstruction artefacts such as the dark stain around the sun.
Nonetheless, our reconstruction does not suffer from glare, and reveals the circular shape of the sun, which is lost in the frame.
The video sequences, along with additional results on sequences from previously released datasets~\cite{Scheerlinck18accv,Zhu18ral}, are available in the supplementary material.

\global\long\def\heightplot{2.03cm} %
\global\long\def\widthplot{2.7cm} %
\global\long\def\vspacecols{0.05cm} %
\setlength{\tabcolsep}{0.05cm} %
\global\long\def\vspacecols{0.12cm} %

\begin{figure}
	\centering
    \begin{tabular}{ccc}

    \rotatebox{90}{\kern0.65cm \footnotesize Selfie}
    \includegraphics[width=\widthplot,height=\heightplot]{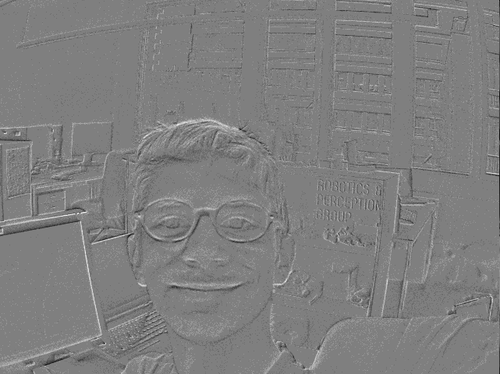}
    & \includegraphics[width=\widthplot,height=\heightplot]{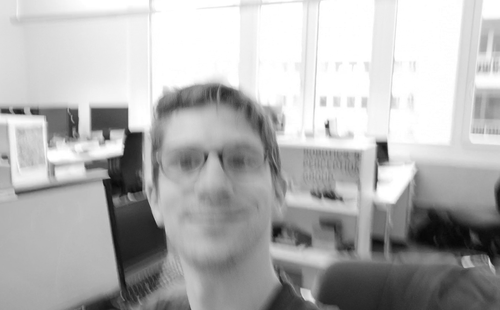}
    & \includegraphics[width=\widthplot,height=\heightplot]{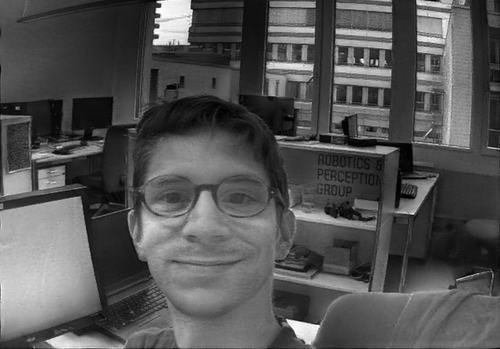}\\[\vspacecols]

    \rotatebox{90}{\kern0.25cm \footnotesize Tunnel Drive}
    \includegraphics[width=\widthplot,height=\heightplot]{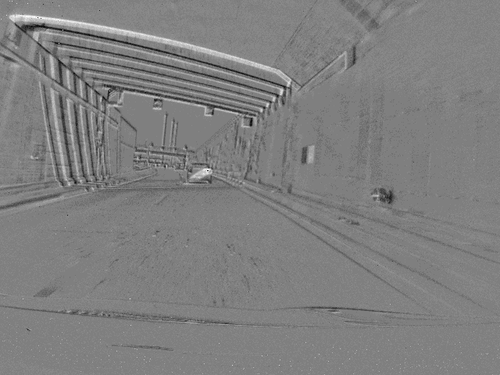}
    & \includegraphics[width=\widthplot,height=\heightplot]{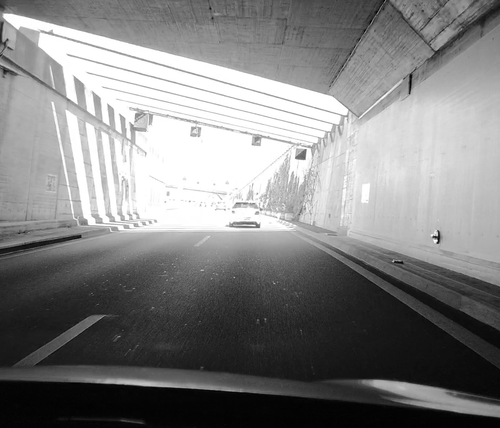}
    & \includegraphics[width=\widthplot,height=\heightplot]{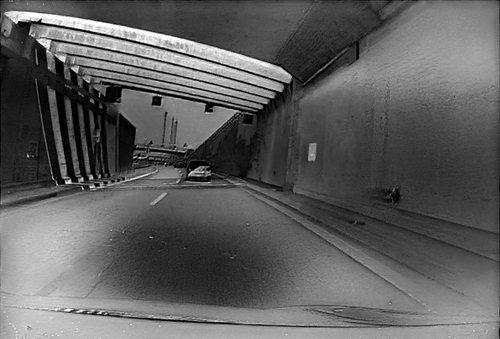}\\[\vspacecols]

    \rotatebox{90}{\kern0.12cm \footnotesize Direct Sunlight}
    \includegraphics[width=\widthplot,height=\heightplot]{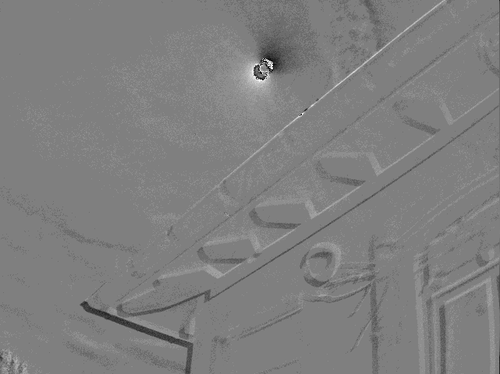}
    & \includegraphics[height=\heightplot]{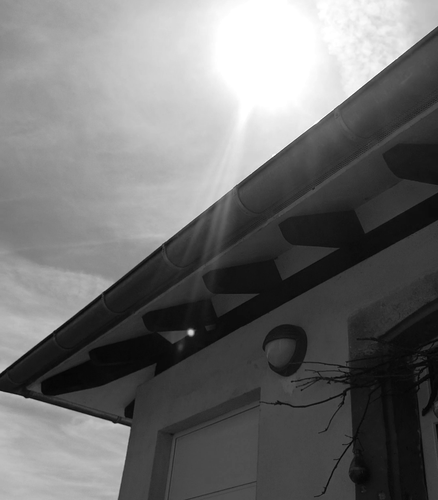}
    & \includegraphics[width=\widthplot,height=\heightplot]{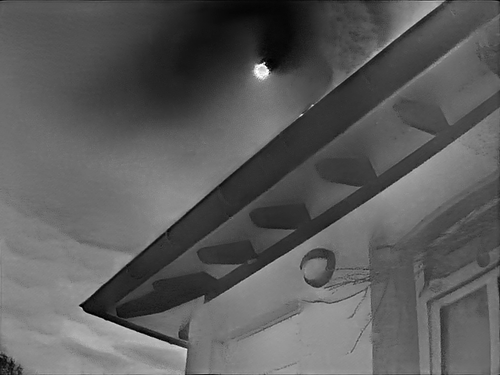}\\[\vspacecols]

    (a) \footnotesize Events & (b) \footnotesize Frame & (c) \footnotesize Reconstruction\\
    \end{tabular}
\caption{Video reconstruction under challenging lighting.
First row: hand-held, indoor ``selfie'' sequence.
Second row: driving sequence recorded while driving out of a tunnel.
Third row: outdoor sequence recorded with the sensors pointing directly at the sun on a bright day.
The frames from the consumer camera (Huawei P20 Pro) (b) suffer from under- or over-exposure, while the events (a) capture the whole dynamic range of the scene, which our method successfully recovers (c).
}
\label{fig:hdr_reconstructions}
\end{figure}

\subsection{Color Video Reconstruction}
\label{sec:color_reconstruction}

Until recently, event cameras were mostly monochrome, producing events based on the variations in luminance \cite{Lichtsteiner08ssc}, and discarding color information.
This has changed with the recent introduction of a sensor that can perceive ``color events'', the Color-DAVIS346 \cite{Taverni18tcsii}.
The Color-DAVIS346 consists of an 8$\times$6mm CMOS chip equipped with a color array filter (CFA), forming an RGBG filter pattern. %
The pixels in the CFA are sensitive to the variation of their specific color filter, producing events that encode color information.
Color reconstruction from such events was first shown by \cite{Moeys17iscas}, where a single color image was recovered from a large set of events using a method similar to \cite{Kim14bmvc}.
Later, \cite{Scheerlinck19cvprw} adapted existing monochrome video reconstruction methods \cite{Munda18ijcv,Rebecq19cvpr} to color by reconstructing the individual color channels independently, resulting in videos that have a quarter of the resolution of the sensor.

\setlength{\tabcolsep}{0.15ex} %
\global\long\def\heightplot{2.254cm} %
\global\long\def\widthplot{3cm} %
\begin{figure}
	\centering
    \begin{tabular}{ccc}

        \includegraphics[width=\widthplot,height=\heightplot]{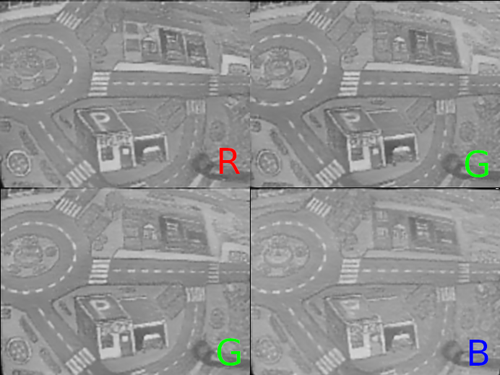}
        & \includegraphics[width=\widthplot,height=\heightplot]{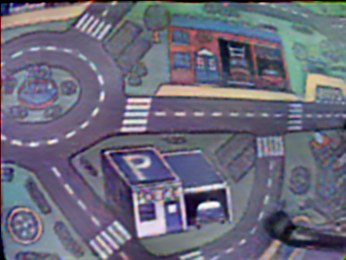}
        & \includegraphics[height=\heightplot]{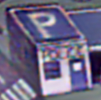}\\
        (a) \footnotesize Individual channels & (b) \footnotesize \cite{Scheerlinck19cvprw} (upsampled, color) & \footnotesize (c) \cite{Scheerlinck19cvprw} (details)\\

        \includegraphics[width=\widthplot,height=\heightplot]{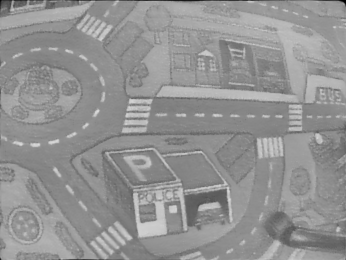}
        & \includegraphics[width=\widthplot,height=\heightplot]{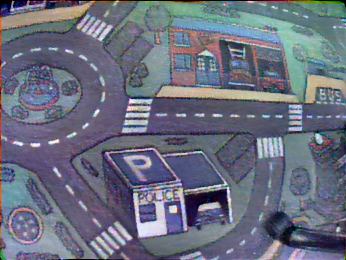}
        & \includegraphics[height=\heightplot]{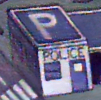}\\
        (d) \footnotesize Grayscale (full res) & (e) \footnotesize Ours (full res, color) & \footnotesize (f) Ours (details)\\

    \end{tabular}
\caption{Our color reconstruction approach.
Color channels are reconstructed independently at quarter resolution (a), then upsampled and recombined into a low-quality color image (b,c).
The latter is combined with a high-quality grayscale image (d) reconstructed using all the events (ignoring the CFA).
The resulting color image (e,f) preserves fine details that are lost in the quarter resolution reconstruction \cite{Scheerlinck19cvprw} (compare (c) and (f)).
}
\label{fig:color_upsampling}
\end{figure}

We now describe a simple method to perform color reconstruction from color event data at full resolution with our network, and then present qualitative results.
Following \cite{Scheerlinck19cvprw}, we reconstruct the four color channels independently at quarter resolution (Fig.~\ref{fig:color_upsampling}(a)), upsample with bicubic interpolation,
and recombine them into a low-quality color image (Fig.~\ref{fig:color_upsampling}(b)). %
We then combine the latter with a full-resolution grayscale image obtained by running our network on all the events (ignoring the CFA).
To do this, we project the (upsampled) color image in the LAB colorspace, and replace the luminance channel with the high-quality grayscale reconstruction (Fig.~\ref{fig:color_upsampling}(d)).
This exploits the human visual system's lower acuity to color differences than to luminance, a widely known phenomenon commonly used to compress videos (chroma subsampling \cite{Poynton02}).

\setlength{\tabcolsep}{0.15ex} %
\global\long\def\heightplot{2.255cm} %
\global\long\def\widthplot{3.3cm} %
\begin{figure*}
	\centering
    \begin{tabular}{ccccc}

        \rotatebox{90}{\kern0.235cm Flying Room}
        \includegraphics[width=\widthplot,height=\heightplot]{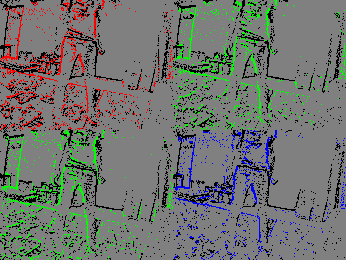}
        & \includegraphics[width=\widthplot,height=\heightplot]{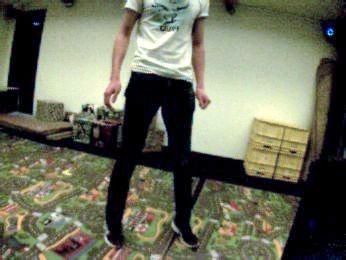}
        & \includegraphics[width=\widthplot,height=\heightplot]{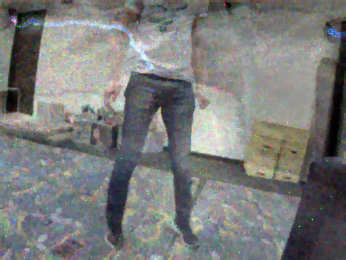}
        & \includegraphics[width=\widthplot,height=\heightplot]{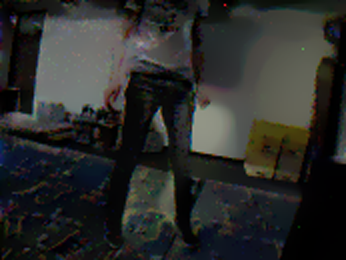}
        & \includegraphics[width=\widthplot,height=\heightplot]{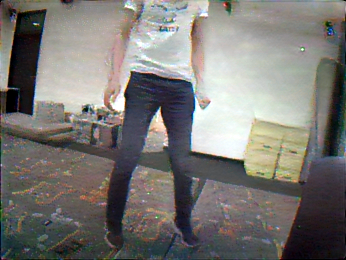}\\

        \rotatebox{90}{\kern0.66cm Jenga}
        \includegraphics[width=\widthplot,height=\heightplot]{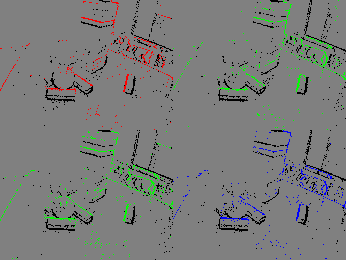}
        & \includegraphics[width=\widthplot,height=\heightplot]{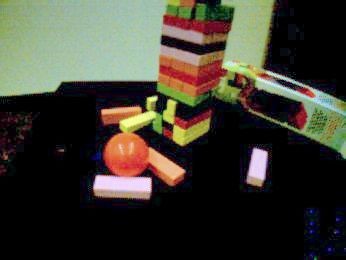}
        & \includegraphics[width=\widthplot,height=\heightplot]{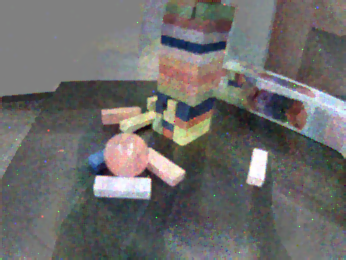}
        & \includegraphics[width=\widthplot,height=\heightplot]{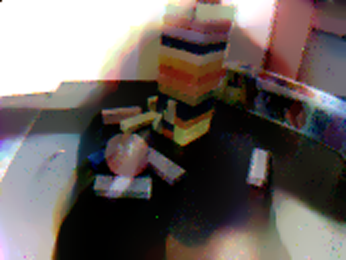}
        & \includegraphics[width=\widthplot,height=\heightplot]{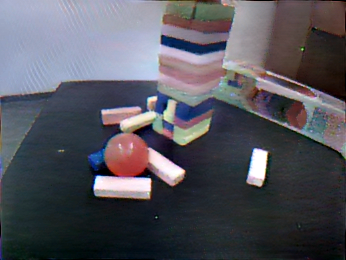}\\

        \rotatebox{90}{\kern0.755cm HDR}
        \includegraphics[width=\widthplot,height=\heightplot]{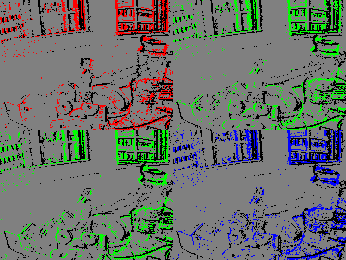}
        & \includegraphics[width=\widthplot,height=\heightplot]{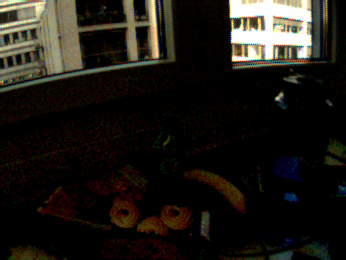}
        & \includegraphics[width=\widthplot,height=\heightplot]{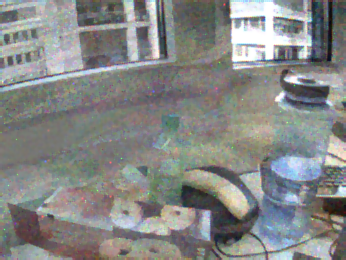}
        & \includegraphics[width=\widthplot,height=\heightplot]{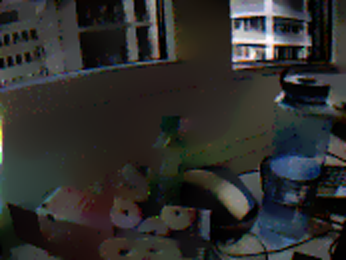}
        & \includegraphics[width=\widthplot,height=\heightplot]{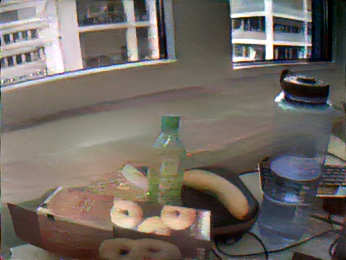}\\

        \rotatebox{90}{\kern0.42cm Low light}
        \includegraphics[width=\widthplot,height=\heightplot]{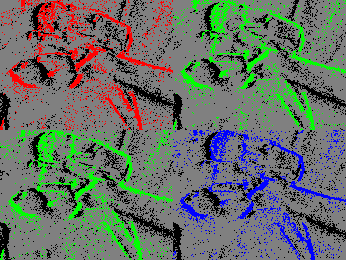}
        & \includegraphics[width=\widthplot,height=\heightplot]{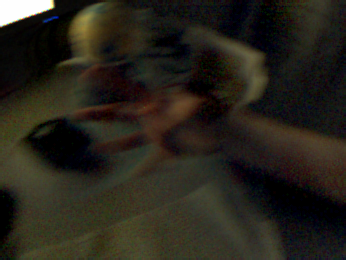}
        & \includegraphics[width=\widthplot,height=\heightplot]{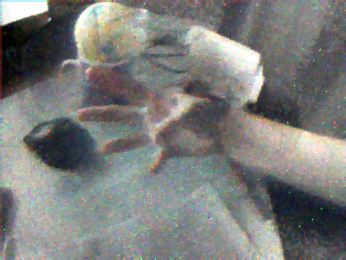}
        & \includegraphics[width=\widthplot,height=\heightplot]{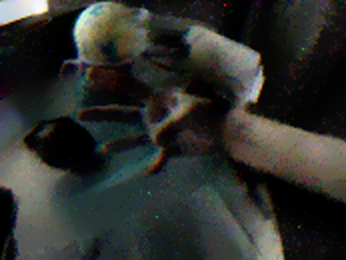}
        & \includegraphics[width=\widthplot,height=\heightplot]{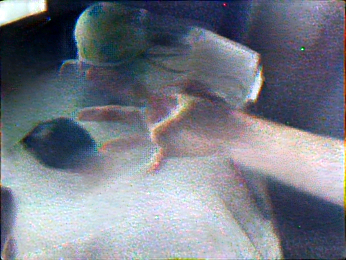}\\

    \footnotesize (a) Color events & \footnotesize (b) DAVIS frame & (c) \footnotesize HF & (d) \footnotesize MR & \footnotesize (e) Our reconstruction
    \end{tabular}
\caption{Color reconstruction results from color events (a) using datasets from \cite{Scheerlinck19cvprw}, comparing a conventional frame (b) with multiple reconstructions from events only: HF (c), MR (d), and ours (d).
For visualization, the color events (a) are split into each color channel.
Positive (ON) events are colored by the corresponding filter color, and negative (OFF) events are black.}
\label{fig:color_reconstructions}
\end{figure*}

Fig.~\ref{fig:color_reconstructions} shows color reconstruction results from our method, compared to HF and MR.
Unlike MR and our method, HF preserves the Bayer pattern, thus color images can simply be obtained by applying a demosaicing algorithm to raw image reconstructions.
For MR, we applied the same technique as our method to obtain color reconstructions.
Qualitatively, the results are consistent with those obtained for grayscale reconstructions (Fig.~\ref{fig:comp_event_camera_dataset}): our method produces cleaner reconstructions than HF (less noise and ``bleeding edges'' artifacts),
and richer reconstructions than MR which tends to smooth out details.
The last two rows of Fig.~\ref{fig:color_reconstructions} show two scenarios with challenging lighting conditions: an HDR scene and a low-light scene.

\subsection{Downstream Applications}
\label{sec:downstream_applications}

We now investigate the possibility of using our reconstructions as an intermediate representation that facilitates direct application of conventional computer vision algorithms to event data.

\mypara{Representations for Event Data.}
Because the output of an event camera is an asynchronous stream of events (a representation that is fundamentally different from images), existing computer vision techniques cannot be directly applied to events.
To address this problem, many algorithms have been specifically tailored to leverage event data for a wide range of applications (a good survey is provided by \cite{Gallego19Arxiv}).
Despite strong differences between these methods, we argue that they follow the same paradigm, relying on two ingredients:
(i) a mechanism to build an internal representation of past event data, and (ii) an inference mechanism to decode new  events given the current internal representation.
For example, \cite{Lagorce17pami,Sironi18cvpr} tackle the task of object classification from event data as follows.
As internal representation, they use a \emph{time surface}: essentially an image recording the timestamp of the last event fired at each pixel (with some temporal decay to increase the influence of recent events), which is updated with every event.
They train a supervised classifier (linear SVM) to predict an object class from this surface.
Finally, object prediction (the inference mechanism) consists in evaluating the classifier, which can either be done with every new event (if fast enough), or at regular intervals \cite{Sironi18cvpr}.

The image reconstructions from our method can also be viewed as a representation for event data.
Similarly to other representations, our network uses, at any given time, all past events to produce an image (in other words, to update the representation).
Unlike other representations, however, our image reconstructions live in the space of natural images.
As such, they are \emph{transferrable}: any computer vision algorithm operating on regular images can be used as the inference mechanism, enabling the application of pre-existing vision algorithms to event data.
We acknowledge, however, that reconstructing an image with our network is slow compared to other methods that use simpler and more efficient representations that may be tailored to a task (\eg time surfaces for optic flow estimation \cite{Benosman14tnnls}). %
Nevertheless, for the remainder of this section, we will show the effectiveness of our reconstructions as event representations on two different downstream tasks: object classification from events and camera pose estimation with events and inertial measurements.
In both cases, we achieve state-of-the-art accuracy.

\mypara{Object Classification.}\label{sec:exp_classification}
Pattern recognition from event data is an active research topic.
While one line of work focuses on spiking neural architectures (SNNs) to recognize patterns from a stream of events with minimal latency (H-FIRST \cite{Orchard15pami}), conventional machine learning techniques combined with novel event representations such as time surfaces (HOTS~\cite{Lagorce17pami}) have shown the most promising results so far.
Recently, HATS~\cite{Sironi18cvpr} addressed the problem of object classification from a stream of events.
They proposed several modifications to HOTS, and achieved major improvements in classification accuracy, outperforming all prior approaches by a large margin.

We propose an alternative approach to object classification based on a stream of events.
Instead of using a hand-crafted event representation, we directly apply a classification network (trained on image data) to images reconstructed from events.
We compare our approach against several recent methods: HOTS, and the state-of-the-art HATS, using the datasets and metric (classification accuracy) used in the HATS paper.
The N-MNIST (Neuromorphic-MNIST) and N-Caltech101 datasets \cite{Orchard15fns} are event-based versions of the MNIST \cite{Lecun98ieee} and Caltech101 \cite{Li06pami} datasets.
To convert the images to event sequences, an event camera was placed on a motor, and automatically moved while pointing at images from MNIST (respectively Caltech101) that were projected onto a white wall.
The N-CARS dataset \cite{Sironi18cvpr}
proposes a binary classification task: deciding whether a car is visible or not using a \SI{100}{\ms} sequence of events.
Fig.~\ref{fig:preview_classification_datasets} shows a sample event sequence from each of the three datasets.

Our approach follows the same methodology for each dataset.
First, for each event sequence in the training set, we use our network to reconstruct an image from the events (Fig.~\ref{fig:preview_classification_datasets}, bottom row).
We then train an off-the-shelf CNN for object classification using the reconstructed images from the training set.
For \mbox{N-MNIST}, we use a simple CNN (details in supplementary material) and train it from scratch.
For N-Caltech101 and N-CARS, we use ResNet-18 \cite{He16cvpr}, initialized with weights pretrained on ImageNet \cite{Russakovsky15ijcv}, and fine-tune the network for the dataset at hand.
Once trained, we evaluate each network on the test set (images reconstructed from the events in the test set) and report the classification accuracy.
Furthermore, we perform a transfer learning experiment for the N-MNIST and N-Caltech101 datasets (for which corresponding images are available for every event sequence): we train the CNN on the conventional image datasets, and evaluate the network directly on images reconstructed from events without fine-tuning.

\setlength{\tabcolsep}{0.25cm} %
\global\long\def\heightplot{1.7cm} %
\begin{figure}
	\centering
    \begin{tabular}{ccc}
    \includegraphics[height=\heightplot]{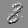}\hspace{1ex}
    & \includegraphics[height=\heightplot]{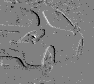}\hspace{1ex}
    & \includegraphics[height=\heightplot]{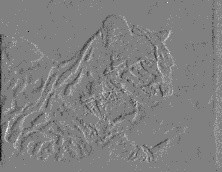}\\
    \includegraphics[height=\heightplot]{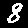}\hspace{1ex}
    & \includegraphics[height=\heightplot]{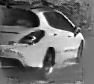}\hspace{1ex}
    & \includegraphics[height=\heightplot]{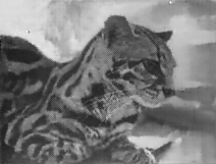}\\
    \footnotesize (a) N-MNIST & \footnotesize (b) N-CARS & \footnotesize (c) N-Caltech101 \\
    \end{tabular}
\caption{Samples from each dataset used in the evaluation of our object classification approach based on events (Section~\ref{sec:exp_classification}). Top: preview of the event sequence. Bottom: our image reconstruction.
}
\label{fig:preview_classification_datasets}
\end{figure}

\begin{table}
\caption{Classification accuracy compared to recent approaches, including HATS \cite{Sironi18cvpr}, the state-of-the-art.
}
\label{tab:classification_results}
\centering
\ra{1.05}
\resizebox{1.0\linewidth}{!}{
	\small
	\begin{tabular}{@{}l@{\hspace{6mm}}*{12}{c@{\hspace{4mm}}}c@{\hspace{8mm}}r@{\hspace{3mm}}r@{}}
		\toprule
		       & N-MNIST & N-CARS & N-Caltech101 \\
		\midrule
		HOTS & 0.808 & 0.624 & 0.210 \\
		HATS/linear SVM & \textbf{0.991} & 0.902 & 0.642 \\
		HATS/ResNet-18 & n.a. & 0.904 & 0.700 \\
		Ours (transfer learning) & 0.807 & n.a. & 0.821 \\
		Ours (fine-tuned) & 0.983 & \textbf{0.910} & \textbf{0.866} \\
		\bottomrule
	\end{tabular}
}
\end{table}

For the baselines, we directly report the accuracy provided in \cite{Sironi18cvpr}. %
To make the comparison with HATS as fair as possible, we also provide results of classifying HATS features with a ResNet-18 network (instead of the linear SVM that was used originally).
The results are presented in Table~\ref{tab:classification_results}, where the datasets are presented in increasing order of difficulty from left to right.
Despite the simplicity of our approach, it outperforms all baselines.
The gap between our method and the state-of-the-art increases with the difficulty of the datasets.
While we perform slightly worse than HATS on N-MNIST (98.3\% versus 99.1\%), this can be attributed to the synthetic nature of N-MNIST, for which our approach does not bring substantial advantages compared to a hand-crafted feature representation such as HATS.
Note that, in contrast to HATS, we did not perform any hyperparameter tuning.
On N-CARS (binary classification task with natural event data), our method slightly outperforms the baseline  (91\% versus 90.4\% for HATS).
However, N-CARS is almost saturated in terms of accuracy. %

On N-Caltech101 (the most challenging dataset, requiring classification of natural event data into 101 object classes), our method outperforms HATS by a large margin (86.6\% versus 70.0\%).
This significant gap can be explained by the fact that our approach leverages decades of computer vision research and datasets.
Lifting the event stream into the image domain with our events-to-video approach allows us to use a mature CNN architecture that was pretrained on existing labeled datasets. We can thus use powerful hierarchical features learned on a large body of image data -- something that is not possible with event data, for which labeled datasets are scarce.
Strikingly, our approach, in a pure transfer learning setting (i.e.\ feeding images reconstructed from events to a network trained on real image data) performs better than all other methods, while not using the event sequences from the training set.
To the best of our knowledge, this is the first time that direct transfer learning between image data and event data has been achieved.

While the proposed approach reaches state-of-the-art accuracy, alternative approaches such as HATS are computationally more efficient, mostly because updating the internal representation (time surface) requires less operations
than generating a new image with our neural network.
Nonetheless, our approach is real-time capable.
On N-Caltech101, end-to-end classification takes less than \SI{10}{\ms} (sequence reconstruction: $\leq\!\SI{8}{\ms}$, object classification: $\leq\!\SI{2}{\ms}$) on an NVIDIA RTX 2080 Ti GPU.
More details can be found in Section~\ref{sec:network_analysis}.

\mypara{Visual-Inertial Odometry.}\label{sec:exp_vio}
\setlength{\tabcolsep}{0.15cm} %
\global\long\def\heightplot{2.625cm} %
\global\long\def\widthplot{3.5cm} %
\begin{figure}[t]
	\centering
    \begin{tabular}{cc}
    \includegraphics[width=\widthplot,height=\heightplot]{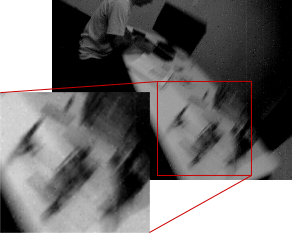}
    &\includegraphics[width=\widthplot,height=\heightplot]{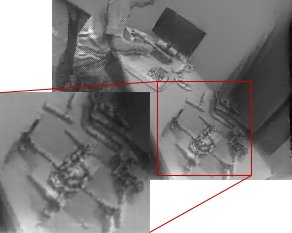}\\
    (a) DAVIS frame & (b) Our reconstruction\\
    \end{tabular}
\caption{Comparison of DAVIS frames and reconstructed frames on a high-speed portion of the `dynamic\_6dof' sequence.
Our reconstructions from events do not suffer from motion blur, which leads to increased pose estimation accuracy (Table~\ref{tab:vio_results}).
}
\vspace{-3mm}
\label{fig:vio_motion_blur_frames}
\end{figure}
\begin{table}[t]
\caption{Mean translation error (in meters) on the sequences from~\cite{Mueggler17ijrr}.
Our method outperforms all other methods that use events and IMU, including UltimateSLAM~(E+I).
Surprisingly, it even performs on par with UltimateSLAM (E+F+I), while not using additional frames. Methods for which the mean translation error exceeds $\SI{5}{\meter}$ are marked as ``failed''. %
}
\label{tab:vio_results}
\centering
\ra{1.05}
\resizebox{1.0\linewidth}{!}{
	\small
	\begin{tabular}{@{}l@{\hspace{5mm}}*{12}{c@{\hspace{3mm}}}c@{\hspace{8mm}}r@{\hspace{2mm}}r@{}}
		\toprule
		          & Ours & U.SLAM & U.SLAM & HF & MR & VINS-Mono \\
	    Inputs    & \small E+I & \small E+I & \small E+F+I & \small E+I & \small E+I & \small F+I \\
		\midrule
		shapes\_translation & 0.18 & 0.32 & \textbf{0.17} & failed & 2.00 & 0.93 \\
		poster\_translation & \textbf{0.05} & 0.09 & 0.06 & 0.49 & 0.15 & failed\\
		boxes\_translation & \textbf{0.15} & 0.81 & 0.26 & 0.70 & 0.45 & 0.22\\
		dynamic\_translation & \textbf{0.08} & 0.23 & 0.09 & 0.58 & 0.17 & 0.13\\
		shapes\_6dof & 1.09 & 0.09 & \textbf{0.06} & failed & 3.00 & 1.99\\
		poster\_6dof & \textbf{0.12} & 0.20 & 0.22 & 0.45 & 0.17 & 1.99\\
		boxes\_6dof & 0.62 & 0.41 & \textbf{0.34} & 1.71 & 1.17 & 0.94\\
		dynamic\_6dof & 0.15 & 0.27 & \textbf{0.11} & failed & 0.55 & 0.76\\
		hdr\_boxes & 0.34 & 0.44 & 0.37 & 0.64 & 0.66 & \textbf{0.32}\\
		\midrule
		Mean & 0.31 & 0.32 & \textbf{0.19} & 0.76 & 0.92 & 0.91\\
		Median & \textbf{0.15} & 0.27 & 0.17 & 0.61 & 0.55 & 0.84\\
		\bottomrule
	\end{tabular}
}
\end{table}
The task of visual-inertial odometry (VIO) is to recover the 6-degrees-of-freedom (6-DOF) pose of a camera from a set of visual measurements (images or events) and inertial measurements from an inertial measurement unit (IMU) that is rigidly attached to the camera.
Because of its importance in augmented/virtual reality and mobile robotics, VIO has been extensively studied in the last decade and is relatively mature today \cite{Mourikis07icra,Leutenegger15ijrr,Blosch15iros,Forster17troOnmanifold,Qin18tro}.
Yet systems based on conventional cameras fail in challenging conditions such as high-speed motions and high-dynamic-range environments. This has recently motivated the development of VIO systems with event data (EVIO) \cite{Zhu17cvpr,Rebecq17bmvc,Rosinol18ral}.

The state-of-the-art EVIO system, UltimateSLAM~\cite{Rosinol18ral}, operates by independently tracking visual features from pseudo-images reconstructed from events using motion compensation \cite{Rebecq17bmvc} (i.e.\ the internal representation) plus optional images from a conventional camera, and fusing the tracks with inertial measurements using an existing optimization backend \cite{Leutenegger15ijrr}.

Here, we go one step further and directly apply an off-the-shelf VIO system (specifically, VINS-Mono \cite{Qin18tro}, which is state-of-the-art \cite{Delmerico18icra}) to videos reconstructed from events using either our approach, MR, or HF, and evaluate against UltimateSLAM.
As is standard \cite{Zhu17cvpr,Rebecq17bmvc,Rosinol18ral}, we use sequences from the Event Camera Dataset \cite{Mueggler17ijrr}, which contain events, frames, and IMU measurements from a DAVIS240C \cite{Brandli14ssc} sensor.
Each sequence is 60 seconds long, and contains data from a hand-held event camera undergoing a variety of motions in several environments.
All sequences feature extremely fast motions (angular velocity up to \SI{880}{\degree/\second} and linear velocity up to \SI{3.5}{\meter/\second}), which leads to severe motion blur on the frames (Fig.~\ref{fig:vio_motion_blur_frames}).
We compare our approach against the two operating modes of UltimateSLAM: UltimateSLAM (E+I) which uses only events and IMU, and UltimateSLAM (E+F+I) that uses the events, the IMU, and additional frames. %
We run a publicly available VIO evaluation toolbox \cite{Zhang18iros} on raw trajectories provided by the authors of UltimateSLAM, which ensures that the trajectories estimated by all methods are evaluated in the exact same manner.
For completeness, we also report results from running VINS-Mono directly on the frames from the DAVIS sensor.

Table \ref{tab:vio_results} presents the mean translation error of each method, for all datasets (additional results are presented in the supplement).
First, we note that our method performs better than UltimateSLAM (E+I) on all sequences, with the exception of the `shapes\_6dof' sequence.
This sequence features a few synthetic shapes with very few features ($\leq\!10$), which cause VINS-Mono to not properly initialize, leading to high error (note that this is a problem with VINS-Mono and not our image reconstructions).
Overall, the median error of our method is \SI{0.15}{\meter}, which is almost half the error of UltimateSLAM (E+I) (\SI{0.27}{\meter}) which uses the exact same data. %
Indeed, while UltimateSLAM (E+I) uses coarse pseudo-images created from a single, small window of events, our network is able to reconstruct images with finer details and higher temporal consistency -- both of which lead to better feature tracks and thus better pose estimates.
Even more strikingly, our approach performs on par with UltimateSLAM (E+F+I), while the latter requires additional frames which we do not need.
The median error of both methods is comparable (\SI{0.15}{\meter} for ours versus \SI{0.17}{\meter} for UltimateSLAM (E+F+I)).%

Finally, we point out that running the same VIO (VINS-Mono) on competing image reconstructions (MR and HF) yields significantly larger tracking errors (e.g.\ median error three times larger for MR), which further highlights the superiority of our image reconstructions for downstream vision applications.
We acknowledge that our approach is not as fast as UltimateSLAM. %
Since the main difference between both approaches is how they build the internal event representation, a rough estimate of the performance gap can be obtained by comparing the time it takes for each method to synthesize a new image.
UltimateSLAM takes about $\SI{1}{\ms}$ on a CPU whereas our method takes  $\leq\!\SI{4}{\ms}$ on a high-end GPU.
Nevertheless, our events-to-video network allows harnessing the outstanding properties of events for VIO and exceeds the accuracy of state-of-the-art EVIO algorithms that were designed specifically for event data.

\section{Analysis}
\label{sec:network_analysis}

In this section, we provide a thorough analysis of our network.
First, we measure the computational efficiency of our approach and compare it against several baselines.
Second, we perform some ablation studies to justify the choice of some components.
Finally, we analyze some of the interesting properties of our network and contrast these with prior work.

\subsection{Computational Efficiency}

Here we analyze the performance (i.e.\ computational efficiency) of the network.
We also justify the specific hyperparameters of our architecture as the result of a simple search over these parameters.

We compare the performance of our method against HF \cite{Scheerlinck18accv} and MR \cite{Munda18ijcv}.
Due to fundamental differences in the way each of these methods processes event data, it is difficult to provide a direct and fair performance comparison.
HF processes the event stream in an event-by-event fashion, providing (in theory) a new image reconstruction with every incoming event.
However, the raw image reconstructions from HF need to be filtered (for example, using a bilateral filter) to obtain results with reasonable quality.
While MR can in principle also operate in an event-by-event fashion, its best quality results are obtained when it processes batches of events. %
Our method also operates on batches of events.
In Table~\ref{tab:performance_comparison}, we report the mean ``frame synthesis time'', which we define as the time it takes to process $N=10@000$ events
(this number was chosen because it yields good-quality images for all three methods).
We ran our method and MR on an NVIDIA GeForce RTX 2080 Ti GPU, and HF on an Intel Core i9-9900K @ 3.60 GHz CPU.

\begin{table}[h!]
\caption{Frame synthesis time (defined as the time it takes to process a batch of $\NumEvents=10@000$ events here) for our method, HF, and MR.
HF works best when some filtering (e.g.\ a bilateral filter) is applied ($^*$), which increases the per-frame computation time.}
\label{tab:performance_comparison}
\centering
\ra{1.05}
	\begin{tabular}{@{}l@{\hspace{6mm}}*{12}{l@{\hspace{4mm}}}l@{\hspace{8mm}}l@{\hspace{3mm}}l@{}}
		\toprule
		       & Frame synthesis time (ms) \\
		\midrule
		HF & $0.70$ / $1.45^*$ \\
		MR & $0.84$ \\
	    Ours &  $5.53$ \\
		\bottomrule
	\end{tabular}
\end{table}

\mypara{Discussion.}
Our method is not as fast as HF or MR, which can process event data roughly 5 times faster.
While high performance is not the main focus of the present work, our network can nonetheless easily run in real-time, providing state-of-the-art reconstructions in terms of video quality.
We believe there is ample room for performance improvements.
First, the performance of our approach may improve significantly when implemented on hardware specifically optimized to perform fast and efficient inference in neural networks.
Second, exploiting the sparsity of the event tensors (most values of which are zeros) could additionally improve the computational efficiency by a large margin.
One promising direction in that regard would be to use sparse convolutions \cite{Graham18cvpr} or hardware accelerators designed to efficiently process sparse inputs \cite{Aimar18tnnls}.
Finally, we believe one of the most alluring characteristics of our method is its ability to summarize a large number of events into one high-quality image.
Since the network is robust to the number of events in each window (Section~\ref{sec:network_analysis}), it can be used with large windows of events when online operation is required (for example, for generating a live video preview),
and run again offline with smaller windows to generate a high framerate video a posterori (as shown for example in Section~\ref{sec:highspeed_reconstructions}).
Alternatively, our network could also be used to generate high-quality images at low framerates (i.e.\ using large batches of events), which could be fused with event data using a complementary filter \cite{Scheerlinck18accv}
to synthesize high-quality videos at very high framerate more efficiently.

\mypara{Searching for a Lightweight Architecture.}
\label{sec:architecture_search}
In this section, we show that our choice of network architecture parameters provides a good trade-off between quality and performance.
We performed a simple architecture search over important hyperparameters:
\begin{itemize}
  \item number of encoders $\NumEncoders$ in the range $\left\{ 2, 3, 4 \right\}$,
  \item number of residual blocks $\NumResBlocks$ (in $\left\{ 0, 1, 2 \right\}$),
  \item type of skip connection $\oplus$ (concatenation or sum),
  \item number of feature channels $\NumBaseFeatureChannels$ ($\left\{8, 16, 32, 64\right\}$).
\end{itemize}

\begin{figure}
\centering
  \includegraphics[width=0.85\linewidth]{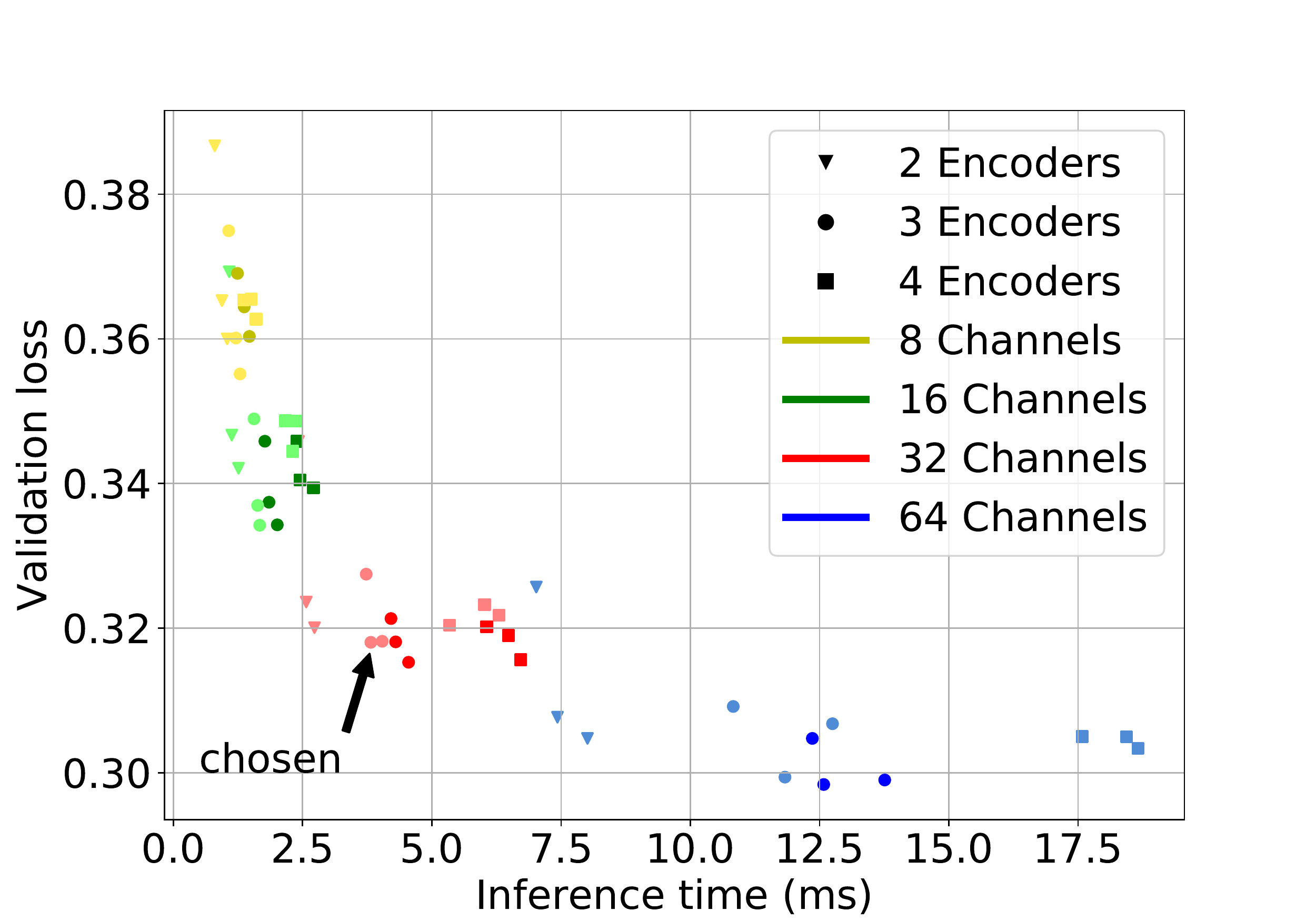}
\caption{Identifying a good trade-off between accuracy and efficiency.
Each point in this graph corresponds to a different architecture variant.
The color intensity indicates the type of skip connection: light colors correspond to element-wise sum and normal colors to concatenation.
The distribution forms an ``elbow'' that suggests a good trade-off between inference time and reconstruction quality.}
\label{fig:neural_architecture_search}
\end{figure}

We then trained a simplified version of the network (with the recurrent connection disabled) for each parameter combination ($72$ combinations in total) on a small subset of the full dataset to convergence (10 epochs),
monitoring the validation loss as well as the mean inference time per event tensor (evaluated on the full network).
The results are presented in Fig.~\ref{fig:neural_architecture_search}. %
Notably, the distribution of the architectures form an ``elbow'', which indicates that there exists a good trade-off between model complexity (i.e.\ inference time) and reconstruction quality (validation loss).
Based on these results, we choose $\NumEncoders=3$, $\NumResBlocks=2$, $\NumBaseFeatureChannels=32$, and element-wise sum for the skip connection.

\subsection{Ablation Studies}

We now present some ablation studies to highlight the impact of some of the key features of our architecture.

\mypara{Temporal Loss.}
To measure the impact of the temporal loss (Section~\ref{sec:loss}, Eq.~\eqref{eq:temporal_loss}), we trained the network without it (using $\WeightTemporalLoss=0$).
Table~\ref{tab:ablation_study_temporal_loss} compares the resulting network with the full network, in terms of temporal consistency and image reconstruction quality.
We use the same sequences as in our quantitative evaluation (Section~\ref{sec:evaluation}), and report the mean values over all datasets, for each metric.
Unsurprisingly, the network that was trained with the temporal loss achieves better temporal consistency (31\% decrease of the temporal error on average).
More interestingly, the full network also performs better in terms of image quality overall (average improvement of about 5\%), suggesting that the temporal loss also acts
as a regularizer, driving the optimizer to converge to a better local optimum.

\mypara{Recurrent Connection.}
Table~\ref{tab:ablation_study_remove_recurrent_connection} compares the image quality and temporal consistency when the recurrent connection of our network is removed.
It highlights the role that the recurrent connection plays in achieving good video reconstruction quality.
The recurrent connection increases temporal video consistency by a large margin (65\% decrease in temporal error), removing the high-frequency blinking present in the reconstructions produced without the recurrent connection (see supplementary video).
The recurrent connection also increases the image quality (15\%), suggesting that the network can effectively leverage its short-term memory to reconstruct accurate images.

\mypara{ConvLSTM vs.\ vanilla RNN.}
We now compare our network (based on stacked \ConvLSTM s \cite{Shi15nips}) to the preliminary version of this work \cite{Rebecq19cvpr} based on a vanilla RNN, focusing on the generalization ability of both networks
to the duration of the event windows used (or, equivalently, the number of events in each window).
We train our network and \cite{Rebecq19cvpr} with the same training data, and %
evaluate both networks at two different inference rates, feeding non-overlapping event windows of a fixed duration $\tau$.
In the first experiment, we use $\tau=\SI{50}{\ms}$, which is close to the average duration of each event tensor in the training data.
In the second experiment, we use $\tau=\SI{5}{\ms}$ to assess the generalization ability of both networks to a window size that is significantly different from the training data.
Note that the second experiment is harder since the networks see a much smaller number of events at each time step (i.e.\ incomplete information), and thus must rely more strongly on their internal memory to recall the missing data necessary to produce good quality reconstructions.
The results are reported in Table~\ref{tab:comp_framerate}.
When the window length is close to the training conditions ($\tau=\SI{50}{\ms}$), the networks perform similarly.
However, with $\tau=\SI{5}{\ms}$, the image quality drops drastically for the vanilla RNN \cite{Rebecq19cvpr}, while degrading only slightly with our network (6\% decrease in SSIM).
Our network is thus more robust to varying window sizes, maintaining intensity forward in time in a more stable fashion.

\begin{table}[t]
\caption{Ablation study: effect of the temporal loss. The temporal loss improves the temporal consistency as well as the image quality.}
\label{tab:ablation_study_temporal_loss}
\newcolumntype{Z}{S[table-format=2.2,table-auto-round]}
\centering

\setlength{\tabcolsep}{1.8mm}
\ra{1.05}
\small
\begin{tabular}{c*{4}Z}
  \toprule
  & {MSE} & {SSIM} & {LPIPS} & {Temporal Error} \\
  \midrule
  w/o temporal loss & 0.0730 & 0.5935 & 0.4349 & 2.0810 \\
  w/ temporal loss & \bfseries 0.0522 & \bfseries 0.6157 & \bfseries 0.4092 & \bfseries 1.4257 \\
  \bottomrule
\end{tabular}

\end{table}

\begin{table}[t]
\caption{Ablation study: effect of the recurrent connection.
The quality and temporal consistency improve when using the recurrent connection, validating that our network is able to successfully propagate information through time.
}
\label{tab:ablation_study_remove_recurrent_connection}
\newcolumntype{Z}{S[table-format=2.2,table-auto-round]}
\centering
\setlength{\tabcolsep}{2mm}
\ra{1.05}
\small
\begin{tabular}{c*{4}Z}
  \toprule
  & {MSE}  & {SSIM}  & {LPIPS}  & {Temporal Error}  \\
  \midrule
  w/o recurrent & 0.0769 & 0.5364 & 0.4736 & 3.9964 \\
  w/ recurrent & \bfseries 0.0522 & \bfseries 0.6157 & \bfseries 0.4092 & \bfseries 1.4257 \\
  \bottomrule
\end{tabular}

\end{table}

\begin{table}[t]
\caption{Reconstruction quality with $\tau=\SI{50}{\ms}$ versus $\tau=\SI{5}{\ms}$ windows. %
The RNN \cite{Rebecq19cvpr} does not generalize to $\tau=\SI{5}{\ms}$, yielding poor quality reconstructions (high MSE and LPIPS, low SSIM).
In contrast, the reconstruction quality of our network degrades only slightly ($\leq$6\% decrease in SSIM), thanks to its more stable recurrent connection.
}
\label{tab:comp_framerate}
\newcolumntype{Z}{S[table-format=2.2,table-auto-round]}
\centering
\setlength{\tabcolsep}{5mm}
\ra{1.05}
\small
\begin{tabular}{c*{4}Z}
  \toprule
  & {MSE} & {SSIM} & {LPIPS} \\
  \midrule
  RNN ($\tau=\SI{50}{\ms}$)  & 0.0542 & 0.6100 & \bfseries 0.4038 \\
  RNN ($\tau=\SI{5}{\ms}$) & 0.1436 & 0.2685 & 0.7065 \\
  Ours ($\tau=\SI{50}{\ms}$) & \bfseries 0.0522 & \bfseries 0.6157 & 0.4092 \\
  Ours ($\tau=\SI{5}{\ms}$) & 0.0605 & 0.5800 & 0.4355 \\
  \bottomrule
\end{tabular}

\end{table}

\setlength{\tabcolsep}{0.3ex} %
\global\long\def\heightplot{2,55cm} %
\global\long\def\widthplot{3.4cm} %
\global\long\def\vspacecols{0.3ex} %
\begin{figure*}
	\centering
    \begin{tabular}{ccccc}
    \includegraphics[width=\widthplot,height=\heightplot]{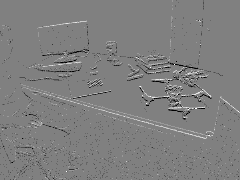}
    & \includegraphics[width=\widthplot,height=\heightplot]{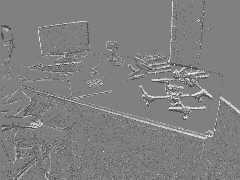}
    & \includegraphics[width=\widthplot,height=\heightplot]{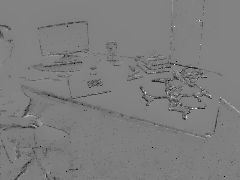}
    & \includegraphics[width=\widthplot,height=\heightplot]{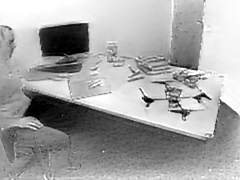}
    & \includegraphics[width=\widthplot,height=\heightplot]{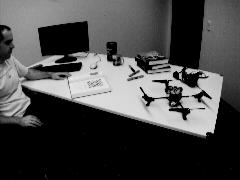}\\[\vspacecols]
    \includegraphics[width=\widthplot,height=\heightplot]{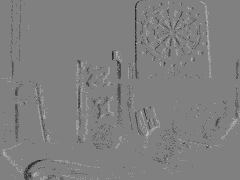}
    & \includegraphics[width=\widthplot,height=\heightplot]{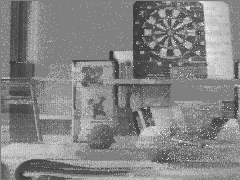}
    & \includegraphics[width=\widthplot,height=\heightplot]{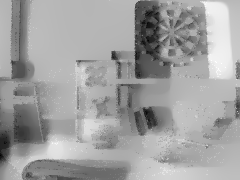}
    & \includegraphics[width=\widthplot,height=\heightplot]{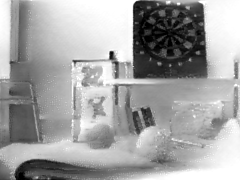}
    & \includegraphics[width=\widthplot,height=\heightplot]{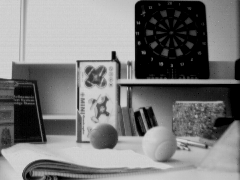}\\[\vspacecols]
    (a) Events & (b) HF & (c) MR & (d) Ours & (e) Ground truth\\
    \end{tabular}
\caption{Analysis of the initialization phase (reconstruction from few events). This figure shows image reconstructions from each method, 0.5 seconds after the sensor was started. HF \cite{Scheerlinck18accv} and MR \cite{Munda18ijcv}, which are based on event integration, cannot recover the intensity correctly, resulting in ``edge'' images (first and second row) or severe ``ghosting'' effects (third row, where the trace of the dartboard is clearly visible).
In contrast, our network successfully reconstructs most of the scene accurately, even with a low number of events.
}
\label{fig:analysis_init_phase}
\end{figure*}

\global\long\def\heightplot{2.55cm} %
\global\long\def\widthplot{3.4cm} %
\begin{figure*}
	\centering
    \begin{tabular}{ccccc}

    \includegraphics[width=\widthplot,height=\heightplot]{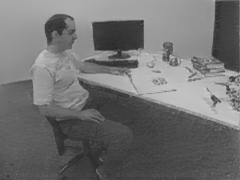}
    & \includegraphics[width=\widthplot,height=\heightplot]{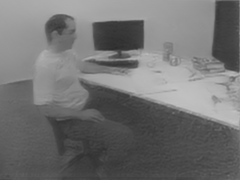}
    & \includegraphics[width=\widthplot,height=\heightplot]{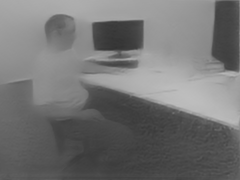}
    & \includegraphics[width=\widthplot,height=\heightplot]{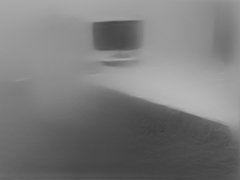}
    & \includegraphics[width=\widthplot,height=\heightplot]{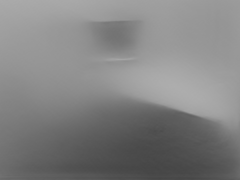}\\

    \includegraphics[width=\widthplot,height=\heightplot]{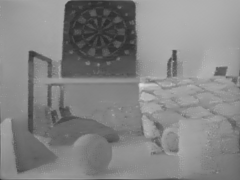}
    & \includegraphics[width=\widthplot,height=\heightplot]{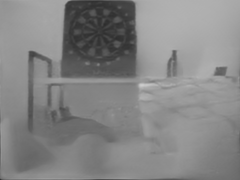}
    & \includegraphics[width=\widthplot,height=\heightplot]{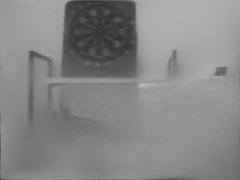}
    & \includegraphics[width=\widthplot,height=\heightplot]{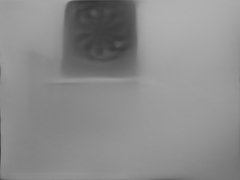}
    & \includegraphics[width=\widthplot,height=\heightplot]{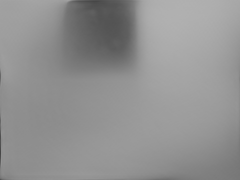}\\

    (a) Initial time & (b) After 1 iteration & (c) After 3 iterations & (d) After 10 iterations & (e) After 20 iterations\\
    \end{tabular}
\caption{In this experiment, the events are artificially stopped at some time $t$, i.e.\ the network is fed empty event tensors in all subsequent iterations.
The correct thing to do would be to simply copy the first image to all subsequent predictions.
The network has instead learned to gradually decay the image.
}
\label{fig:network_memory_short}
\end{figure*}

\subsection{Edge Cases}

We now present two interesting edge cases that shed light on some characteristics of our approach.
We first analyze the initialization phase, when few events have been observed.
We then perform a simple experiment to estimate the effective size of our network's memory, to better understand its behavior in regions with low event rates.

\mypara{Initialization.}
By analyzing the initialization phase (i.e.\ when few events have been previously triggered) we gain interesting insight into how our network operates.
We see significantly different behaviour when compared to prior approaches that are based on direct event integration.
Fig.~\ref{fig:analysis_init_phase} compares image reconstructions from our approach, HF, and MR during the initialization phase.
We specifically examine the interval from $\SI{0}{\second}$ to $\SI{0.5}{\second}$ from the beginning of capture.
HF and MR, which rely on event integration, can only recover the intensity up to the initial (unknown) image $\Image_0$ (i.e.\ they can only recover $\Reconstructed \approx \Image-\Image_0$), which results in an ``edge'' image which does not capture the appearance of the scene correctly.
In contrast, our method successfully leverages deep priors to reconstruct the scene despite the low number of events.

\mypara{Network Memory.}
For how long can our network remember intensity information?
To answer this question -- in other words, to measure the effective size of the temporal receptive field of our network -- we perform a simple experiment. %
We take a given sequence of events and artificially stop the events at a fixed time $t$ (this is achieved in practice by zeroing out all the event tensors with timestamps $\geq t$).
We feed the resulting empty event tensors to our network and present the evolution of the reconstructions as a function of the number of iterations in Fig.~\ref{fig:network_memory_short}.
In the complete absence of events, the current reconstruction should be left untouched, i.e.\ the network should simply copy all the pixel values forward.
As shown in Fig.~\ref{fig:network_memory_short}, this is in fact what the network has learned to do in the first few iterations, although this was never hardcoded in the network design; rather, the recurrent network discovered this pattern from the training data.
However, as can be observed in Fig.~\ref{fig:network_memory_short}, the network gradually decays the image intensity when forced to perform inference with tensors containing no events.
Interestingly, the image decay is not isotropic: the regions with high contrast (e.g.\ the dart board in the first row) tend to be preserved for a higher number of iterations.
While this experiment may seem artificial -- such a situation cannot happen in practice since no event tensor is generated when no events fire -- it sheds light
on the behavior of the network in regions where very few events are fired. %
We believe the length of the memory is strongly tied to the distribution of optical flows in the training data.
Indeed, our synthetic training datasets (aimed at general-purpose reconstruction) contain few ``pauses'' (i.e.\ regions with low event rate), thus our network does not need to retain information for long time periods.
This suggests that our network could be further improved for specific applications by generating training data with a similar distribution of optical flow than the target application.
In autonomous driving for example, the network may learn to retain intensity information for a long time in the center of the image (focus of expansion, low event rate), and for a shorter amount of time on the sides of the image (higher event rate).

\section{Conclusion}

We presented a novel events-to-video reconstruction framework based on a recurrent convolutional network trained on simulated event data.
In addition to outperforming state-of-the-art reconstruction methods on real event data by a large margin ($>\!20\%$ improvement), we showed the applicability of our method to
synthesize high framerate, high dynamic range, and color video reconstructions from event data only.
Finally, we demonstrated the effectiveness of our reconstructions as as intermediate representation that bridges event cameras and mainstream computer vision. %

\ifCLASSOPTIONcompsoc
  \section*{Acknowledgments}
\else
  \section*{Acknowledgment}
\fi

The authors would like to thank Cedric Scheerlinck, Matthias Faessler and his family for valuable discussions and for help recording data, as well as iniVation and Samsung for providing the event cameras that were used for this research.

\ifCLASSOPTIONcaptionsoff
  \newpage
\fi

\bibliographystyle{IEEEtran}
\bibliography{all}

\begin{thebibliography}{10}
\providecommand{\url}[1]{#1}
\csname url@samestyle\endcsname
\providecommand{\newblock}{\relax}
\providecommand{\bibinfo}[2]{#2}
\providecommand{\BIBentrySTDinterwordspacing}{\spaceskip=0pt\relax}
\providecommand{\BIBentryALTinterwordstretchfactor}{4}
\providecommand{\BIBentryALTinterwordspacing}{\spaceskip=\fontdimen2\font plus
\BIBentryALTinterwordstretchfactor\fontdimen3\font minus
  \fontdimen4\font\relax}
\providecommand{\BIBforeignlanguage}[2]{{%
\expandafter\ifx\csname l@#1\endcsname\relax
\typeout{** WARNING: IEEEtran.bst: No hyphenation pattern has been}%
\typeout{** loaded for the language `#1'. Using the pattern for}%
\typeout{** the default language instead.}%
\else
\language=\csname l@#1\endcsname
\fi
#2}}
\providecommand{\BIBdecl}{\relax}
\BIBdecl

\bibitem{Lichtsteiner08ssc}
P.~Lichtsteiner, C.~Posch, and T.~Delbruck, ``{A 128$\times$128 120 dB 15
  $\mu$s latency asynchronous temporal contrast vision sensor},'' \emph{{IEEE}
  J. Solid-State Circuits}, vol.~43, no.~2, pp. 566--576, 2008.

\bibitem{Bardow16cvpr}
P.~Bardow, A.~J. Davison, and S.~Leutenegger, ``Simultaneous optical flow and
  intensity estimation from an event camera,'' in \emph{{IEEE} Conf. Comput.
  Vis. Pattern Recog. (CVPR)}, 2016, pp. 884--892.

\bibitem{Barua16wacv}
S.~Barua, Y.~Miyatani, and A.~Veeraraghavan, ``Direct face detection and video
  reconstruction from event cameras,'' in \emph{{IEEE} Winter Conf. Appl.
  Comput. Vis. (WACV)}, 2016, pp. 1--9.

\bibitem{Munda18ijcv}
G.~Munda, C.~Reinbacher, and T.~Pock, ``Real-time intensity-image
  reconstruction for event cameras using manifold regularisation,'' \emph{Int.
  J. Comput. Vis.}, vol. 126, no.~12, pp. 1381--1393, Jul. 2018.

\bibitem{Scheerlinck18accv}
C.~Scheerlinck, N.~Barnes, and R.~Mahony, ``Continuous-time intensity
  estimation using event cameras,'' in \emph{Asian Conf. Comput. Vis. (ACCV)},
  2018.

\bibitem{Taverni18tcsii}
G.~Taverni, D.~P. Moeys, C.~Li, C.~Cavaco, V.~Motsnyi, D.~S.~S. Bello, and
  T.~Delbruck, ``Front and back illuminated {D}ynamic and {A}ctive {P}ixel
  {V}ision {S}ensors comparison,'' \emph{{IEEE} Trans. Circuits Syst. {II}},
  vol.~65, no.~5, pp. 677--681, 2018.

\bibitem{Conradt09iscas}
J.~Conradt, M.~Cook, R.~Berner, P.~Lichtsteiner, R.~J. Douglas, and
  T.~Delbruck, ``A pencil balancing robot using a pair of {AER} dynamic vision
  sensors,'' in \emph{{IEEE} Int. Symp. Circuits Syst. (ISCAS)}, 2009, pp.
  781--784.

\bibitem{Cook11ijcnn}
M.~Cook, L.~Gugelmann, F.~Jug, C.~Krautz, and A.~Steger, ``Interacting maps for
  fast visual interpretation,'' in \emph{Int. Joint Conf. Neural Netw.
  (IJCNN)}, 2011, pp. 770--776.

\bibitem{Benosman14tnnls}
R.~Benosman, C.~Clercq, X.~Lagorce, S.-H. Ieng, and C.~Bartolozzi,
  ``Event-based visual flow,'' \emph{{IEEE} Trans. Neural Netw. Learn. Syst.},
  vol.~25, no.~2, pp. 407--417, 2014.

\bibitem{Mueggler14iros}
E.~Mueggler, B.~Huber, and D.~Scaramuzza, ``Event-based, 6-{DOF} pose tracking
  for high-speed maneuvers,'' in \emph{IEEE/RSJ Int. Conf. Intell. Robot. Syst.
  (IROS)}, 2014, pp. 2761--2768.

\bibitem{Kim16eccv}
H.~Kim, S.~Leutenegger, and A.~J. Davison, ``Real-time {3D} reconstruction and
  6-{DoF} tracking with an event camera,'' in \emph{Eur. Conf. Comput. Vis.
  (ECCV)}, 2016, pp. 349--364.

\bibitem{Gallego17pami}
G.~Gallego, J.~E.~A. Lund, E.~Mueggler, H.~Rebecq, T.~Delbruck, and
  D.~Scaramuzza, ``Event-based, 6-{DOF} camera tracking from photometric depth
  maps,'' \emph{{IEEE} Trans. Pattern Anal. Mach. Intell.}, vol.~40, no.~10,
  pp. 2402--2412, Oct. 2018.

\bibitem{Lagorce17pami}
X.~Lagorce, G.~Orchard, F.~Gallupi, B.~E. Shi, and R.~Benosman, ``{HOTS}: A
  hierarchy of event-based time-surfaces for pattern recognition,''
  \emph{{IEEE} Trans. Pattern Anal. Mach. Intell.}, vol.~39, no.~7, pp.
  1346--1359, Jul. 2017.

\bibitem{Sironi18cvpr}
A.~Sironi, M.~Brambilla, N.~Bourdis, X.~Lagorce, and R.~Benosman, ``{HATS}:
  Histograms of averaged time surfaces for robust event-based object
  classification,'' in \emph{{IEEE} Conf. Comput. Vis. Pattern Recog. (CVPR)},
  2018, pp. 1731--1740.

\bibitem{Zhu18rss}
A.~Z. Zhu, L.~Yuan, K.~Chaney, and K.~Daniilidis, ``{EV-FlowNet}:
  Self-supervised optical flow estimation for event-based cameras,'' in
  \emph{Robotics: Science and Systems (RSS)}, 2018.

\bibitem{Zhou18eccv}
Y.~Zhou, G.~Gallego, H.~Rebecq, L.~Kneip, H.~Li, and D.~Scaramuzza,
  ``Semi-dense {3D} reconstruction with a stereo event camera,'' in \emph{Eur.
  Conf. Comput. Vis. (ECCV)}, 2018, pp. 242--258.

\bibitem{Kim14bmvc}
H.~Kim, A.~Handa, R.~Benosman, S.-H. Ieng, and A.~J. Davison, ``Simultaneous
  mosaicing and tracking with an event camera,'' in \emph{British Mach. Vis.
  Conf. (BMVC)}, 2014.

\bibitem{Gehrig18eccv}
D.~Gehrig, H.~Rebecq, G.~Gallego, and D.~Scaramuzza, ``Asynchronous,
  photometric feature tracking using events and frames,'' in \emph{Eur. Conf.
  Comput. Vis. (ECCV)}, 2018.

\bibitem{Aharon06tsp}
M.~Aharon, M.~Elad, and A.~M. Bruckstein, ``{K}-{SVD}: An algorithm for
  designing overcomplete dictionaries for sparse representation,'' \emph{{IEEE}
  Trans. Signal Process.}, vol.~54, no.~11, pp. 4311--4322, 2006.

\bibitem{Brandli14iscas}
C.~Brandli, L.~Muller, and T.~Delbruck, ``Real-time, high-speed video
  decompression using a frame- and event-based {DAVIS} sensor,'' in
  \emph{{IEEE} Int. Symp. Circuits Syst. (ISCAS)}, 2014, pp. 686--689.

\bibitem{Xu18ted}
J.~Xu, J.~Zou, and Z.~Gao, ``Comment on temperature and parasitic photocurrent
  effects in {D}ynamic {V}ision {S}ensors,'' \emph{{IEEE} Trans. Electron
  Devices}, vol.~65, no.~7, pp. 3081--3082, Jul. 2018.

\bibitem{Zhu18eccvw}
A.~Z. Zhu, L.~Yuan, K.~Chaney, and K.~Daniilidis, ``Unsupervised event-based
  optical flow using motion compensation,'' in \emph{Eur. Conf. Comput. Vis.
  Workshops (ECCVW)}, 2018.

\bibitem{Rebecq18corl}
H.~Rebecq, D.~Gehrig, and D.~Scaramuzza, ``{ESIM}: an open event camera
  simulator,'' in \emph{Conf. on Robotics Learning (CoRL)}, 2018.

\bibitem{TsungYi14eccv}
T.~Lin, M.~Maire, S.~J. Belongie, L.~D. Bourdev, R.~B. Girshick, J.~Hays,
  P.~Perona, D.~Ramanan, P.~Doll{\'{a}}r, and C.~L. Zitnick, ``Microsoft
  {COCO}: Common objects in context,'' in \emph{Eur. Conf. Comput. Vis.
  (ECCV)}, 2014.

\bibitem{Brandli14ssc}
C.~Brandli, R.~Berner, M.~Yang, S.-C. Liu, and T.~Delbruck, ``A 240x180 130{dB}
  3us latency global shutter spatiotemporal vision sensor,'' \emph{{IEEE} J.
  Solid-State Circuits}, vol.~49, no.~10, pp. 2333--2341, 2014.

\bibitem{Ronneberger15icmicci}
O.~Ronneberger, P.~Fischer, and T.~Brox, ``{U}-net: Convolutional networks for
  biomedical image segmentation,'' in \emph{International Conference on Medical
  Image Computing and Computer-Assisted Intervention}, 2015.

\bibitem{Shi15nips}
X.~Shi, Z.~Chen, H.~Wang, D.~Yeung, W.~Wong, and W.~Woo, ``Convolutional {LSTM}
  network: {A} machine learning approach for precipitation nowcasting,'' in
  \emph{Conf. Neural Inf. Process. Syst. (NIPS)}, 2015.

\bibitem{Ioffe15icml}
S.~Ioffe and C.~Szegedy, ``Batch normalization: Accelerating deep network
  training by reducing internal covariate shift,'' in \emph{Proc. Int. Conf.
  Mach. Learning (ICML)}, 2015.

\bibitem{He16cvpr}
K.~He, X.~Zhang, S.~Ren, and J.~Sun, ``Deep residual learning for image
  recognition,'' in \emph{{IEEE} Conf. Comput. Vis. Pattern Recog. (CVPR)},
  2016, pp. 770--778.

\bibitem{Rebecq19cvpr}
H.~Rebecq, R.~Ranftl, V.~Koltun, and D.~Scaramuzza, ``Events-to-video: Bringing
  modern computer vision to event cameras,'' in \emph{{IEEE} Conf. Comput. Vis.
  Pattern Recog. (CVPR)}, 2019.

\bibitem{Johnson16eccv}
J.~Johnson, A.~Alahi, and F.~Li, ``Perceptual losses for real-time style
  transfer and super-resolution,'' in \emph{Eur. Conf. Comput. Vis. (ECCV)},
  2016.

\bibitem{Zhang18cvprLPIPS}
R.~Zhang, P.~Isola, A.~A. Efros, E.~Shechtman, and O.~Wang, ``The unreasonable
  effectiveness of deep features as a perceptual metric,'' in \emph{{IEEE}
  Conf. Comput. Vis. Pattern Recog. (CVPR)}, 2018.

\bibitem{Simonyan15iclr}
K.~Simonyan and A.~Zisserman, ``Very deep convolutional networks for
  large-scale image recognition,'' in \emph{Int. Conf. Learn. Representations
  ({ICLR})}, 2015.

\bibitem{Russakovsky15ijcv}
O.~Russakovsky, J.~Deng, H.~Su, J.~Krause, S.~Satheesh, S.~Ma, Z.~Huang,
  A.~Karpathy, A.~Khosla, M.~Bernstein, A.~C. Berg, and F.-F. Li, ``{ImageNet}
  large scale visual recognition challenge,'' \emph{Int. J. Comput. Vis.}, vol.
  115, no.~3, pp. 211--252, Apr. 2015.

\bibitem{Lai18eccv}
W.~Lai, J.~Huang, O.~Wang, E.~Shechtman, E.~Yumer, and M.~Yang, ``Learning
  blind video temporal consistency,'' in \emph{Eur. Conf. Comput. Vis. (ECCV)},
  2018.

\bibitem{Paszke17nipsw}
A.~Paszke, S.~Gross, S.~Chintala, G.~Chanan, E.~Yang, Z.~DeVito, Z.~Lin,
  A.~Desmaison, L.~Antiga, and A.~Lerer, ``Automatic differentiation in
  {PyTorch},'' in \emph{NIPS Workshops}, 2017.

\bibitem{Kingma15iclr}
D.~P. Kingma and J.~L. Ba, ``Adam: A method for stochastic optimization,''
  \emph{Int. Conf. Learn. Representations ({ICLR})}, 2015.

\bibitem{Mueggler17ijrr}
E.~Mueggler, H.~Rebecq, G.~Gallego, T.~Delbruck, and D.~Scaramuzza, ``The
  event-camera dataset and simulator: Event-based data for pose estimation,
  visual odometry, and {SLAM},'' \emph{Int. J. Robot. Research}, vol.~36, pp.
  142--149, 2017.

\bibitem{Wang04tip}
Z.~Wang, A.~C. Bovik, H.~R. Sheikh, and E.~P. Simoncelli, ``Image quality
  assessment: From error visibility to structural similarity,'' \emph{{IEEE}
  Trans. Image Process.}, vol.~13, no.~4, pp. 600--612, Apr. 2004.

\bibitem{Ilg17cvpr}
E.~Ilg, N.~Mayer, T.~Saikia, M.~Keuper, A.~Dosovitskiy, and T.~Brox,
  ``{FlowNet} 2.0: Evolution of optical flow estimation with deep networks,''
  in \emph{{IEEE} Conf. Comput. Vis. Pattern Recog. (CVPR)}, 2017, pp.
  1647--1655.

\bibitem{Son17isscc}
B.~Son, Y.~Suh, S.~Kim, H.~Jung, J.-S. Kim, C.~Shin, K.~Park, K.~Lee, J.~Park,
  J.~Woo, Y.~Roh, H.~Lee, Y.~Wang, I.~Ovsiannikov, and H.~Ryu, ``A 640x480
  dynamic vision sensor with a 9um pixel and {300Meps} address-event
  representation,'' in \emph{{IEEE} Intl. Solid-State Circuits Conf. (ISSCC)},
  2017.

\bibitem{ReduceFlicker}
\BIBentryALTinterwordspacing
``{R}educe{F}licker.'' [Online]. Available:
  \url{http://avisynth.nl/index.php/ReduceFlicker}
\BIBentrySTDinterwordspacing

\bibitem{Zhu18ral}
A.~Z. Zhu, D.~Thakur, T.~Ozaslan, B.~Pfrommer, V.~Kumar, and K.~Daniilidis,
  ``The multivehicle stereo event camera dataset: An event camera dataset for
  {3D} perception,'' \emph{{IEEE} Robot. Autom. Lett.}, vol.~3, no.~3, pp.
  2032--2039, Jul. 2018.

\bibitem{Moeys17iscas}
D.~P. Moeys, C.~Li, J.~N.~P. Martel, S.~Bamford, L.~Longinotti, V.~Motsnyi,
  D.~S.~S. Bello, and T.~Delbruck, ``Color temporal contrast sensitivity in
  dynamic vision sensors,'' in \emph{{IEEE} Int. Symp. Circuits Syst. (ISCAS)},
  2017, pp. 1--4.

\bibitem{Scheerlinck19cvprw}
C.~Scheerlinck, H.~Rebecq, T.~Stoffregen, N.~Barnes, R.~Mahony, and
  D.~Scaramuzza, ``{CED:} color event camera dataset,'' in \emph{{IEEE} Conf.
  Comput. Vis. Pattern Recog. Workshops (CVPRW)}, 2019.

\bibitem{Poynton02}
\BIBentryALTinterwordspacing
C.~Poynton, ``Chroma subsampling notation,'' 2002. [Online]. Available:
  \url{http://vektor.theorem.ca/graphics/ycbcr/Chroma_subsampling_notation.pdf}
\BIBentrySTDinterwordspacing

\bibitem{Gallego19Arxiv}
\BIBentryALTinterwordspacing
G.~Gallego, T.~Delbruck, G.~Orchard, C.~Bartolozzi, B.~Taba, A.~Censi,
  S.~Leutenegger, A.~Davison, J.~Conradt, K.~Daniilidis, and D.~Scaramuzza,
  ``Event-based vision: {A} survey,'' \emph{ar{X}iv e-prints}, vol.
  abs/1904.08405, 2019. [Online]. Available:
  \url{http://arxiv.org/abs/1904.08405}
\BIBentrySTDinterwordspacing

\bibitem{Orchard15pami}
G.~Orchard, C.~Meyer, R.~Etienne-Cummings, C.~Posch, N.~Thakor, and
  R.~Benosman, ``{HFirst}: A temporal approach to object recognition,''
  \emph{{IEEE} Trans. Pattern Anal. Mach. Intell.}, vol.~37, no.~10, pp.
  2028--2040, 2015.

\bibitem{Orchard15fns}
G.~Orchard, A.~Jayawant, G.~K. Cohen, and N.~Thakor, ``Converting static image
  datasets to spiking neuromorphic datasets using saccades,'' \emph{Front.
  Neurosci.}, vol.~9, p. 437, 2015.

\bibitem{Lecun98ieee}
Y.~Lecun, L.~Bottou, Y.~Bengio, and P.~Haffner, ``Gradient-based learning
  applied to document recognition,'' \emph{Proc. {IEEE}}, vol.~86, no.~11, pp.
  2278--2324, 1998.

\bibitem{Li06pami}
L.~Fei-Fei, R.~Fergus, and P.~Perona, ``One-shot learning of object
  categories,'' \emph{{IEEE} Trans. Pattern Anal. Mach. Intell.}, vol.~28,
  no.~4, pp. 594--611, 2006.

\bibitem{Mourikis07icra}
A.~I. Mourikis and S.~I. Roumeliotis, ``A multi-state constraint {K}alman
  filter for vision-aided inertial navigation,'' in \emph{{IEEE} Int. Conf.
  Robot. Autom. (ICRA)}, Apr. 2007, pp. 3565--3572.

\bibitem{Leutenegger15ijrr}
S.~Leutenegger, S.~Lynen, M.~Bosse, R.~Siegwart, and P.~Furgale,
  ``Keyframe-based visual-inertial {SLAM} using nonlinear optimization,''
  \emph{Int. J. Robot. Research}, 2015.

\bibitem{Blosch15iros}
M.~Bloesch, S.~Omari, M.~Hutter, and R.~Siegwart, ``Robust visual inertial
  odometry using a direct {EKF}-based approach,'' in \emph{IEEE/RSJ Int. Conf.
  Intell. Robot. Syst. (IROS)}, 2015.

\bibitem{Forster17troOnmanifold}
C.~Forster, L.~Carlone, F.~Dellaert, and D.~Scaramuzza, ``On-manifold
  preintegration for real-time visual-inertial odometry,'' \emph{{IEEE} Trans.
  Robot.}, vol.~33, no.~1, pp. 1--21, 2017.

\bibitem{Qin18tro}
T.~Qin, P.~Li, and S.~Shen, ``{VINS}-{M}ono: A robust and versatile monocular
  visual-inertial state estimator,'' \emph{{IEEE} Trans. Robot.}, vol.~34, pp.
  1004--1020, 2018.

\bibitem{Zhu17cvpr}
A.~Z. Zhu, N.~Atanasov, and K.~Daniilidis, ``Event-based visual inertial
  odometry,'' in \emph{{IEEE} Conf. Comput. Vis. Pattern Recog. (CVPR)}, 2017,
  pp. 5816--5824.

\bibitem{Rebecq17bmvc}
H.~Rebecq, T.~Horstschaefer, and D.~Scaramuzza, ``Real-time visual-inertial
  odometry for event cameras using keyframe-based nonlinear optimization,'' in
  \emph{British Mach. Vis. Conf. (BMVC)}, 2017.

\bibitem{Rosinol18ral}
A.~{Rosinol Vidal}, H.~Rebecq, T.~Horstschaefer, and D.~Scaramuzza, ``Ultimate
  {SLAM}? combining events, images, and {IMU} for robust visual {SLAM} in {HDR}
  and high speed scenarios,'' \emph{{IEEE} Robot. Autom. Lett.}, vol.~3, no.~2,
  pp. 994--1001, Apr. 2018.

\bibitem{Delmerico18icra}
J.~Delmerico and D.~Scaramuzza, ``A benchmark comparison of monocular
  visual-inertial odometry algorithms for flying robots,'' \emph{{IEEE} Int.
  Conf. Robot. Autom. (ICRA)}, 2018.

\bibitem{Zhang18iros}
Z.~Zhang and D.~Scaramuzza, ``A tutorial on quantitative trajectory evaluation
  for visual(-inertial) odometry,'' in \emph{IEEE/RSJ Int. Conf. Intell. Robot.
  Syst. (IROS)}, 2018.

\bibitem{Graham18cvpr}
B.~Graham, M.~Engelcke, and L.~van~der Maaten, ``{3D} semantic segmentation
  with submanifold sparse convolutional networks,'' in \emph{{IEEE} Conf.
  Comput. Vis. Pattern Recog. (CVPR)}, 2018, pp. 9224--9232.

\bibitem{Aimar18tnnls}
A.~Aimar, H.~Mostafa, E.~Calabrese, A.~Rios{-}Navarro, R.~Tapiador{-}Morales,
  I.~Lungu, M.~B. Milde, F.~Corradi, A.~Linares{-}Barranco, S.~Liu, and
  T.~Delbr{\"{u}}ck, ``{N}ull{H}op: {A} flexible convolutional neural network
  accelerator based on sparse representations of feature maps,'' \emph{{IEEE}
  Trans. Neural Netw. Learn. Syst.}, 2018.

\bibitem{Redmon18arxiv}
\BIBentryALTinterwordspacing
J.~Redmon and A.~Farhadi, ``Yolov3: An incremental improvement,'' \emph{ar{X}iv
  e-prints}, vol. abs/1804.02767, 2018. [Online]. Available:
  \url{http://arxiv.org/abs/1804.02767}
\BIBentrySTDinterwordspacing

\bibitem{Li18cvpr}
Z.~Li and N.~Snavely, ``Megadepth: Learning single-view depth prediction from
  internet photos,'' in \emph{{IEEE} Conf. Comput. Vis. Pattern Recog. (CVPR)},
  2018.

\end{thebibliography}

\begin{IEEEbiography}[{\includegraphics[width=1in,height=1.25in,clip,keepaspectratio]{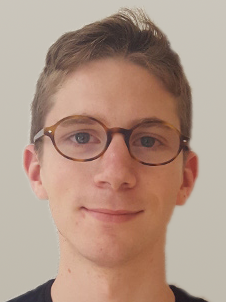}}]{Henri Rebecq}
    is a Ph.D. student in the Robotics and Perception Group at the University of Zurich and ETH Zurich, where he is working on on event-based vision.
    He received an M.Sc.Eng. degree from T\'el\'ecom ParisTech, and an M.Sc. degree from Ecole Normale Sup\'erieure de Cachan, France.
    His research interests include 3D reconstruction, SLAM, and computational photography with event cameras.
    He is a recipient of the Best Industry Paper award at the British Machine Vision Conference (2016), the Misha Mahowald Prize for Neuromorphic Engineering (2017), the Qualcomm Innovation Fellowship (2018),
    and the IEEE RA-L 2018 Best Paper (Honorable Mention).
    \end{IEEEbiography}

\begin{IEEEbiography}[{\includegraphics[width=1in,height=1.25in,clip,keepaspectratio]{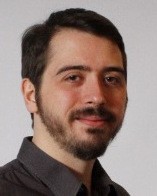}}]{Ren\'e Ranftl}
    is a Senior Research Scientist at the Intel Intelligent Systems Lab in Munich, Germany. He received a M.Sc. degree and a Ph.D. degree
    from Graz University of Technology, Austria, in 2010 and 2015, respectively. His research interest spans topics in computer vision, machine learning, and robotics.
    He is the recipient of several awards including a Best Systems Paper Award at the Conference on Robot Learning 2018
    and the Best Paper Award at the International Conference on Scale Space and Variational Methods~2015.

\end{IEEEbiography}

\begin{IEEEbiography}[{\includegraphics[width=1in,height=1.25in,clip,keepaspectratio]{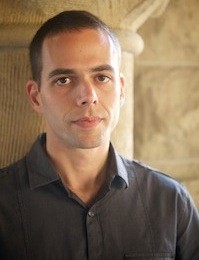}}]{Vladlen Koltun}
    is a Senior Principal Researcher and the director of the Intelligent Systems Lab at Intel. His lab conducts high-impact basic research on intelligent systems, with emphasis on computer vision, robotics, and machine learning. He has mentored more than 50 PhD students, postdocs, research scientists, and PhD student interns, many of whom are now successful research leaders.
\end{IEEEbiography}

\begin{IEEEbiography}[{\includegraphics[width=1in,height=1.25in,clip,keepaspectratio]{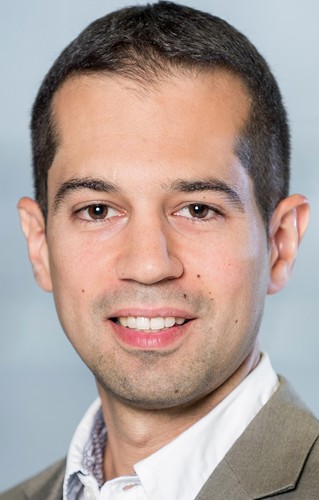}}]{Davide Scaramuzza}
    (born in 1980, Italian) is Associate Professor of Robotics and Perception at both departments of Informatics (University of Zurich) and Neuroinformatics (University of Zurich and ETH Zurich), where he does research at the intersection of robotics, computer vision, and neuroscience.
    He did his PhD in robotics and computer vision at ETH Zurich and a postdoc at the University of Pennsylvania.
    From 2009 to 2012, he led the European project “sFly”, which introduced the PX4 autopilot and pioneered visual-SLAM--based autonomous navigation of micro drones.
    For his research contributions, he was awarded the Misha Mahowald Neuromorphic Engineering Award, the IEEE Robotics and Automation Society Early Career Award, the SNSF-ERC Starting Grant, a Google Research Award, the European Young Research Award, and several conference paper awards.
    He coauthored the book “Introduction to Autonomous Mobile Robots” (published by MIT Press) and more than 100 papers on robotics and perception.
\end{IEEEbiography}

\clearpage
\appendices
\section{Supplementary Video \& Code}

As the main focus of the present work is video reconstruction, we strongly encourage the reader to view the supplementary video, which contains:

\begin{itemize}
  \item Video reconstructions from our method on various event datasets, with a visual comparison to several state of the art methods (Section~\ref{sec:evaluation}).
  \item High framerate videos of the high speed experiments (Section~\ref{sec:highspeed_reconstructions}).
  \item High dynamic range video reconstructions (Section~\ref{sec:hdr_reconstructions}).
  \item Color video reconstructions (Section~\ref{sec:color_reconstruction}).
  \item Video illustrations of the ablation studies (Section~\ref{sec:network_analysis}).
  \item Video of the VINS-Mono visual-inertial odometry algorithm \cite{Qin18tro} running on a video reconstruction from events (Section~\ref{sec:downstream_applications}).
  \item Qualitative results on two additional downstream applications that were not presented in the main paper: object detection (based on YOLOv3 \cite{Redmon18arxiv}), and monocular depth prediction (based on MegaDepth~\cite{Li18cvpr}). We point out that neither of these tasks have ever been shown with event data before this work.
  \end{itemize}

\mypara{Code Release.}
To spur further research, we release the reconstruction code and a pretrained model at: \url{http://rpg.ifi.uzh.ch/E2VID}.

\section{Formal Network Description}

An overview of the network architecture is presented in Fig.~\ref{fig:network_architecture}.
Given an event tensor $\EventTensor_k$ (at time step $k$) and the previous network state, defined as $\NetworkState_k = \left\{ \EncoderState^1_{k-1}, ..., \EncoderState^{\NumEncoders}_{k-1} \right\}$,
our network performs the following sequence of operations (omitting ReLU and batch normalization):

\begin{flalign}
  \xx_k^h &= \Head(\EventTensor_k)\\
  \hh_k^i, \EncoderState_k^i &= \Encoder^i(\hh_k^{i-1}, \EncoderState_{k-1}^i)\\
  \rr_k^j &= \ResBlock^j(\rr_k^{j-1})\\
  \dd_k^l &= \Decoder^l(\dd_k^{l-1} \oplus \hh_k^{\NumEncoders-l+1})\\
  \Reconstructed_k &= \sigma\left(\Prediction(\dd_k^{\NumDecoders} \oplus \xx_k^h)\right)
\end{flalign}

where $1 \leq i \leq \NumEncoders$, $1 \leq j \leq \NumResBlocks$ and $1 \leq l \leq \NumDecoders$, $\hh_k^0 = \xx_k^h$, $r_k^0 = \hh_k^{\NumEncoders}$ and $\dd_k^0 = \rr_k^{\NumResBlocks}$.
At the first iteration ($k=0$), the hidden states for each encoder layer are initialized to zero, \ie $\EncoderState^i_{0} = 0$, for $1 \leq i \leq \NumEncoders$.
The $\oplus$ operator denotes the skip connection function (element-wise sum), and $\sigma$ the sigmoid function.

\section{Why Use Synthetic Training Data?}

\begin{figure}[h!]
\begin{center}
   \includegraphics[width=0.49\linewidth]{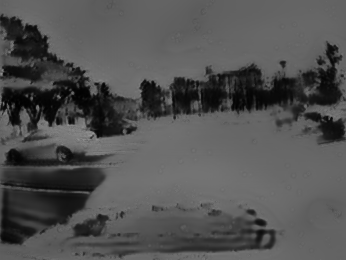}
   \includegraphics[width=0.49\linewidth]{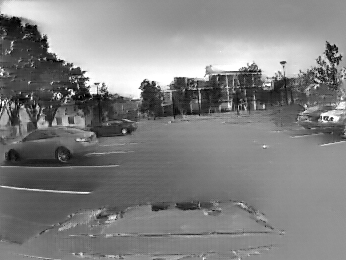}
\end{center}
   \caption{Reconstruction from (i) a network trained only on real event data from the DAVIS240C sensor (left), and (ii) a network trained only on simulated event data (right). This sequence is from the MVSEC dataset, and was recorded with a mDAVIS346 sensor.}
\label{fig:training_real_versus_sim}
\end{figure}

Here, we expand on the reasons that motivated us to train our reconstruction network using synthetic event data.
First, simulation allows to capture a large variety of scenes and motions at very little cost.
Second, a conventional camera (even a high quality one) would provide poor ground truth in high-speed conditions (motion blur) and HDR scenes, which are the conditions in which event sensors excel; by contrast, synthetic data does not suffer from these issues.
Last but not least, simulation allows to randomize the contrast thresholds of the event sensor, which increases the ability of the network to generalize to different sensor configurations (contrast sensitivity).
To illustrate this last point, we show in Fig.~\ref{fig:training_real_versus_sim} (left) what happens when training the network on real event data from an event  camera (specifically, the sequences from the Event Camera Dataset \cite{Mueggler17ijrr} already presented in the main paper, which were recorded with a DAVIS240C sensor), and evaluating the trained network on data coming from a different event sensor (specifically, the `outdoors\_day1` sequence from the MVSEC dataset \cite{Zhu18ral}, which was recorded with a mDAVIS346 sensor): the reconstruction suffers from many artefacts.
This can be explained by the fact that the events from the mDAVIS346 sensor have statistics that are quite different from the training events (DAVIS240C): the set of contrast thresholds are likely quite different between both sensors, and the illumination conditions are also different (outdoor lighting for the MVSEC dataset versus indoor lighting for the training event data).
By contrast, the network trained on simulated event data (Fig.~\ref{fig:training_real_versus_sim}, right) generalizes well to the event data from the mDAVIS346, producing a visually pleasing image reconstruction.

\section{Additional Results}
\label{sec:supplementary_materials}

\subsection{Results on Synthetic Event Data}
We show a quantitative comparison of the reconstruction quality of our method as well as  MR and HF on synthetic event sequences in Table~\ref{tab:image_quality_comparison_synthetic_supp}. We present qualitative reconstruction results on this dataset in Fig.~\ref{fig:comp_synthetic_supp}.
All methods perform better on synthetic data than real data. This is expected because simulated events are free of noise.
Nonetheless, the performance gap between our method and the state of the art is preserved, and even slightly increases (24\% improvement in SSIM, 56\% decrease in LPIPS).
We note that perfect reconstruction, even on noise-free event streams is not possible, since image reconstruction from events
is only possibly up the the quantization limit imposed by the contrast threshold of the event camera.

\begin{table*}
\caption{Comparison of image quality with respect to state of the art on synthetic event sequences.}
\label{tab:image_quality_comparison_synthetic_supp}
\newcolumntype{Z}{S[table-format=2.2,table-auto-round]}
\centering
\setlength{\tabcolsep}{3mm}
\ra{1.05}
\small
\begin{tabular}{@{}lcZZZcZZZcZZZcr@{}}
  \toprule
  \multirow{2}[3]{*}{Dataset} && \multicolumn{3}{c}{MSE}  &&  \multicolumn{3}{c}{SSIM}  &&\multicolumn{3}{c}{LPIPS} \\
  \cmidrule(l{3mm}r{3mm}){3-5} \cmidrule(l{3mm}r{3mm}){7-9} \cmidrule(l{3mm}r{3mm}){11-13}
  && {HF} & {MR} & {Ours} && {HF} & {MR} & {Ours} && {HF} & {MR} & {Ours}  \\
  \midrule
synthetic\_0 && 0.082 & 0.055 & \bfseries 0.015 && 0.536 & 0.608 & \bfseries 0.827 && 0.492 & 0.466 & \bfseries 0.260 \\
synthetic\_1 && 0.151 & 0.135 & \bfseries 0.062 && 0.384 & 0.448 & \bfseries 0.598 && 0.536 & 0.559 & \bfseries 0.366 \\
synthetic\_2 && 0.069 & 0.078 & \bfseries 0.022 && 0.603 & 0.676 & \bfseries 0.816 && 0.423 & 0.415 & \bfseries 0.257 \\
synthetic\_3 && 0.067 & 0.050 & \bfseries 0.038 && 0.574 & 0.661 & \bfseries 0.749 && 0.447 & 0.426 & \bfseries 0.327 \\
synthetic\_4 && 0.078 & 0.058 & \bfseries 0.021 && 0.617 & 0.667 & \bfseries 0.854 && 0.406 & 0.418 & \bfseries 0.250 \\
synthetic\_5 && 0.077 & 0.077 & \bfseries 0.024 && 0.501 & 0.606 & \bfseries 0.743 && 0.525 & 0.538 & \bfseries 0.357 \\
synthetic\_6 && 0.068 & 0.039 & \bfseries 0.025 && 0.557 & 0.645 & \bfseries 0.765 && 0.439 & 0.476 & \bfseries 0.298 \\
\midrule
Mean && 0.085 & 0.070 & \bfseries 0.030 && 0.539 & 0.616 & \bfseries 0.765 && 0.467 & 0.471 & \bfseries 0.302 \\
   \bottomrule
\end{tabular}
\end{table*}

\setlength{\tabcolsep}{0.3ex} %
\global\long\def\heightplot{2.55cm} %
\global\long\def\widthplot{3.4cm} %
\global\long\def\vspacecols{0.3ex} %
\begin{figure*}
	\centering
    \begin{tabular}{ccccc}
    \includegraphics[width=\widthplot,height=\heightplot]{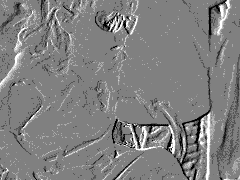}
    & \includegraphics[width=\widthplot,height=\heightplot]{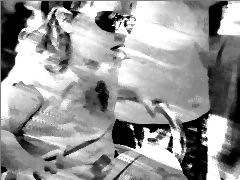}
    & \includegraphics[width=\widthplot,height=\heightplot]{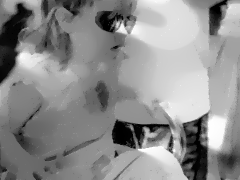}
    & \includegraphics[width=\widthplot,height=\heightplot]{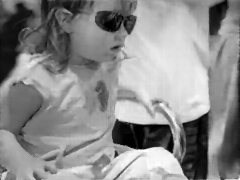}
    & \includegraphics[width=\widthplot,height=\heightplot]{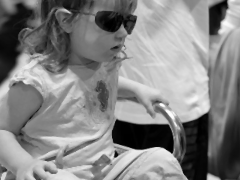}\\[\vspacecols]
    \includegraphics[width=\widthplot,height=\heightplot]{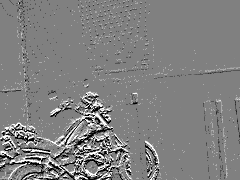}
    & \includegraphics[width=\widthplot,height=\heightplot]{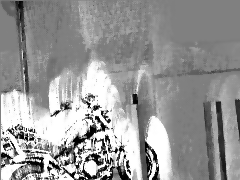}
    & \includegraphics[width=\widthplot,height=\heightplot]{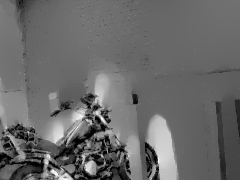}
    & \includegraphics[width=\widthplot,height=\heightplot]{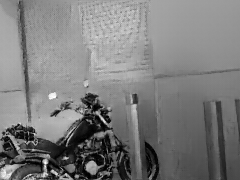}
    & \includegraphics[width=\widthplot,height=\heightplot]{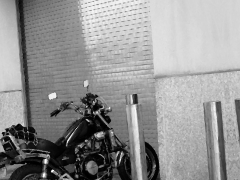}\\[\vspacecols]
    \includegraphics[width=\widthplot,height=\heightplot]{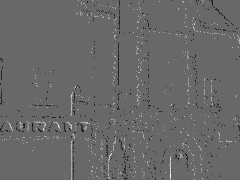}
    & \includegraphics[width=\widthplot,height=\heightplot]{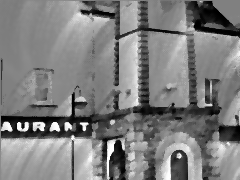}
    & \includegraphics[width=\widthplot,height=\heightplot]{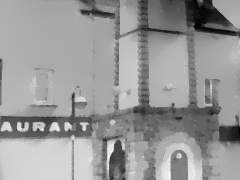}
    & \includegraphics[width=\widthplot,height=\heightplot]{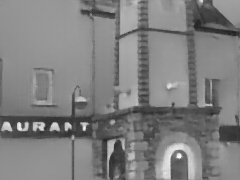}
    & \includegraphics[width=\widthplot,height=\heightplot]{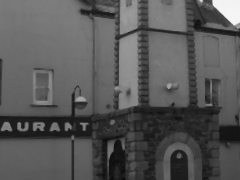}\\[\vspacecols]
    \includegraphics[width=\widthplot,height=\heightplot]{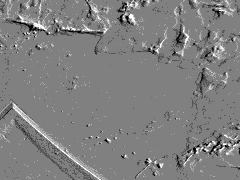}
    & \includegraphics[width=\widthplot,height=\heightplot]{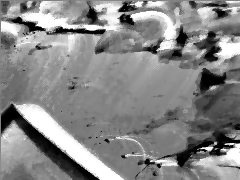}
    & \includegraphics[width=\widthplot,height=\heightplot]{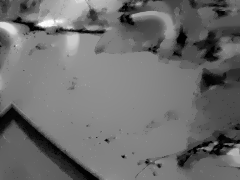}
    & \includegraphics[width=\widthplot,height=\heightplot]{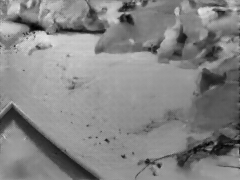}
    & \includegraphics[width=\widthplot,height=\heightplot]{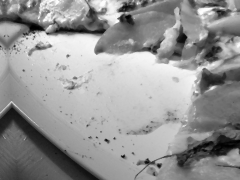}\\[\vspacecols]
    \includegraphics[width=\widthplot,height=\heightplot]{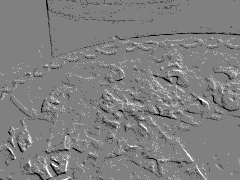}
    & \includegraphics[width=\widthplot,height=\heightplot]{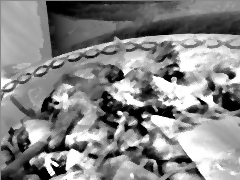}
    & \includegraphics[width=\widthplot,height=\heightplot]{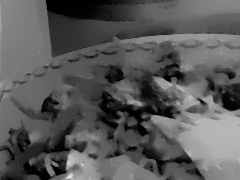}
    & \includegraphics[width=\widthplot,height=\heightplot]{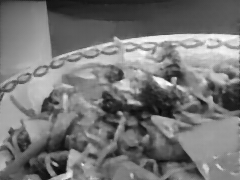}
    & \includegraphics[width=\widthplot,height=\heightplot]{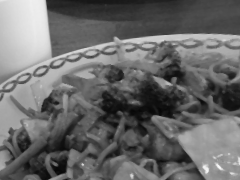}\\[\vspacecols]
    \includegraphics[width=\widthplot,height=\heightplot]{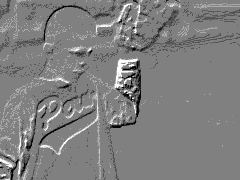}
    & \includegraphics[width=\widthplot,height=\heightplot]{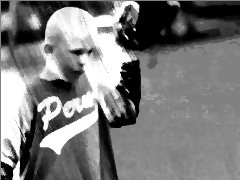}
    & \includegraphics[width=\widthplot,height=\heightplot]{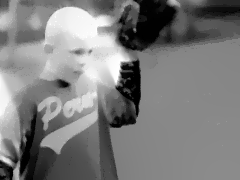}
    & \includegraphics[width=\widthplot,height=\heightplot]{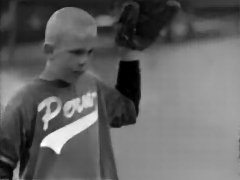}
    & \includegraphics[width=\widthplot,height=\heightplot]{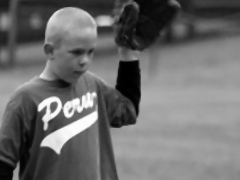}\\[\vspacecols]
    \includegraphics[width=\widthplot,height=\heightplot]{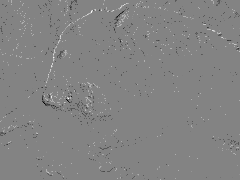}
    & \includegraphics[width=\widthplot,height=\heightplot]{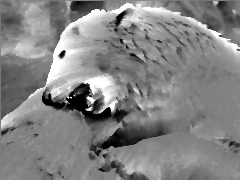}
    & \includegraphics[width=\widthplot,height=\heightplot]{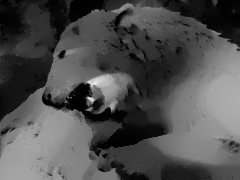}
    & \includegraphics[width=\widthplot,height=\heightplot]{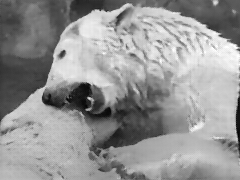}
    & \includegraphics[width=\widthplot,height=\heightplot]{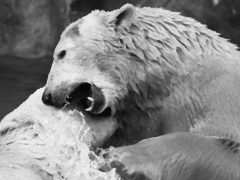}\\[\vspacecols]
    (a) Events & (b) HF & (c) MR & (d) Ours & (e) Ground truth\\
    \end{tabular}
\caption{Qualitative comparison of our reconstruction method with HF \cite{Scheerlinck18accv} and MR \cite{Munda18ijcv} on synthetic sequences from the validation set. Note our method is able to reconstruct fine details such as the bear's fur (last row), which competing methods are not able to preserve.
}
\label{fig:comp_synthetic_supp}
\end{figure*}

\subsection{Additional Qualitative Results on Real Data}

Fig.~\ref{fig:comp_event_camera_dataset_supp} shows qualitative results on sequences from the Event Camera Dataset \cite{Mueggler17ijrr} (which we used for our quantitative evaluation).
Fig~\ref{fig:comp_bardow_supp} shows qualitative results on the sequences introduced by Bardow \etal \cite{Bardow16cvpr}.
Figs.~\ref{fig:qualitative_hdr_supp},~\ref{fig:qualitative_night_supp}~and~\ref{fig:hdr_reconstructions_public_datasets} present HDR reconstruction results on various publicly available datasets \cite{Zhu18ral,Scheerlinck18accv,Scheerlinck19cvprw}.
Further results are  shown in the supplementary video which conveys these results in a better form than still images.

\setlength{\tabcolsep}{0.3ex} %
\global\long\def\heightplot{2.125cm} %
\global\long\def\widthplot{2.8cm} %
\global\long\def\vspacecols{0.3ex} %
\begin{figure*}
	\centering
    \begin{tabular}{cccccc}
    \includegraphics[width=\widthplot,height=\heightplot]{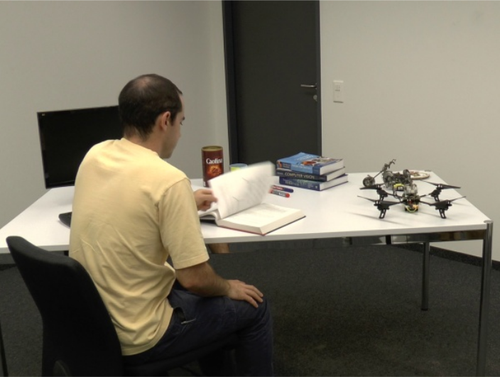}
    & \includegraphics[width=\widthplot,height=\heightplot]{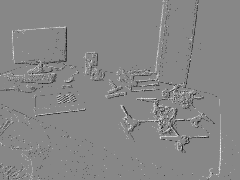}
    & \includegraphics[width=\widthplot,height=\heightplot]{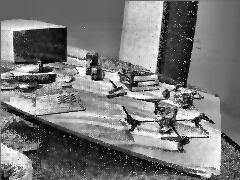}
    & \includegraphics[width=\widthplot,height=\heightplot]{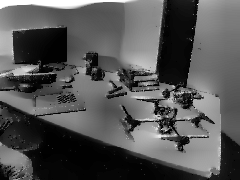}
    & \includegraphics[width=\widthplot,height=\heightplot]{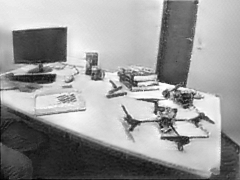}
    & \includegraphics[width=\widthplot,height=\heightplot]{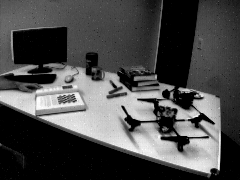}\\[\vspacecols]
    \includegraphics[width=\widthplot,height=\heightplot]{images/comp_event_camera_dataset/preview_boxes.png}
    & \includegraphics[width=\widthplot,height=\heightplot]{images/comp_event_camera_dataset/boxes_6dof/130_events.png}
    & \includegraphics[width=\widthplot,height=\heightplot]{images/comp_event_camera_dataset/boxes_6dof/processed_local/130_HF.png}
    & \includegraphics[width=\widthplot,height=\heightplot]{images/comp_event_camera_dataset/boxes_6dof/processed_local/130_MR.png}
    & \includegraphics[width=\widthplot,height=\heightplot]{images/comp_event_camera_dataset/boxes_6dof/processed_local/130_Ours_RT.png}
    & \includegraphics[width=\widthplot,height=\heightplot]{images/comp_event_camera_dataset/boxes_6dof/processed_local/130_groundtruth.png}\\[\vspacecols]
    \includegraphics[width=\widthplot,height=\heightplot]{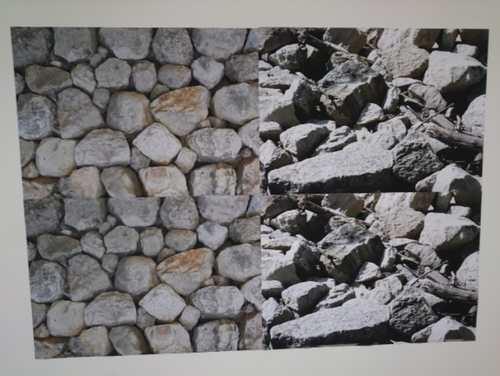}
    & \includegraphics[width=\widthplot,height=\heightplot]{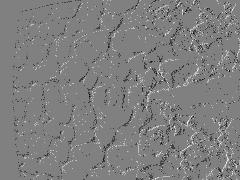}
    & \includegraphics[width=\widthplot,height=\heightplot]{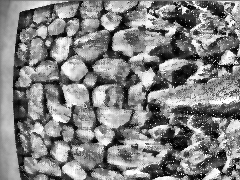}
    & \includegraphics[width=\widthplot,height=\heightplot]{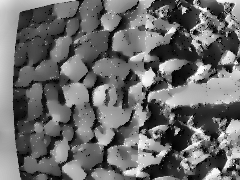}
    & \includegraphics[width=\widthplot,height=\heightplot]{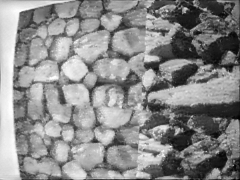}
    & \includegraphics[width=\widthplot,height=\heightplot]{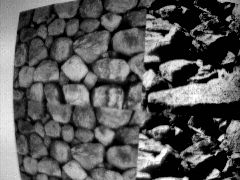}\\[\vspacecols]
    \includegraphics[width=\widthplot,height=\heightplot]{images/comp_event_camera_dataset/preview_shapes.png}
    & \includegraphics[width=\widthplot,height=\heightplot]{images/comp_event_camera_dataset/shapes_6dof/136_events.png}
    & \includegraphics[width=\widthplot,height=\heightplot]{images/comp_event_camera_dataset/shapes_6dof/processed_local/136_HF.png}
    & \includegraphics[width=\widthplot,height=\heightplot]{images/comp_event_camera_dataset/shapes_6dof/processed_local/136_MR.png}
    & \includegraphics[width=\widthplot,height=\heightplot]{images/comp_event_camera_dataset/shapes_6dof/processed_local/136_Ours_RT.png}
    & \includegraphics[width=\widthplot,height=\heightplot]{images/comp_event_camera_dataset/shapes_6dof/processed_local/136_groundtruth.png}\\[\vspacecols]
    \includegraphics[width=\widthplot,height=\heightplot]{images/comp_event_camera_dataset/preview_office.png}
    & \includegraphics[width=\widthplot,height=\heightplot]{images/comp_event_camera_dataset/office_zigzag/104_events.png}
    & \includegraphics[width=\widthplot,height=\heightplot]{images/comp_event_camera_dataset/office_zigzag/processed_local/104_HF.png}
    & \includegraphics[width=\widthplot,height=\heightplot]{images/comp_event_camera_dataset/office_zigzag/processed_local/104_MR.png}
    & \includegraphics[width=\widthplot,height=\heightplot]{images/comp_event_camera_dataset/office_zigzag/processed_local/104_Ours_RT.png}
    & \includegraphics[width=\widthplot,height=\heightplot]{images/comp_event_camera_dataset/office_zigzag/processed_local/104_groundtruth.png}\\[\vspacecols]
    \includegraphics[width=\widthplot,height=\heightplot]{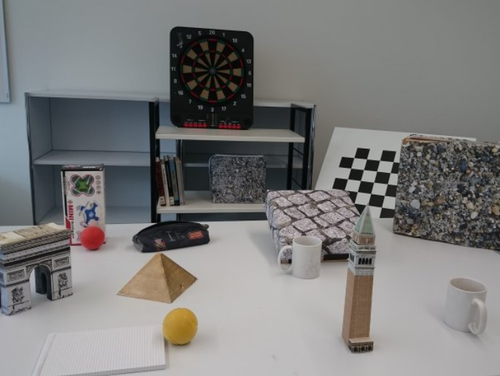}
    & \includegraphics[width=\widthplot,height=\heightplot]{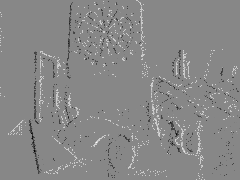}
    & \includegraphics[width=\widthplot,height=\heightplot]{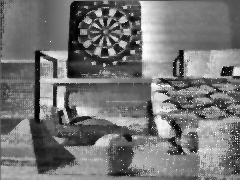}
    & \includegraphics[width=\widthplot,height=\heightplot]{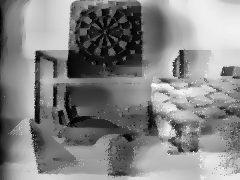}
    & \includegraphics[width=\widthplot,height=\heightplot]{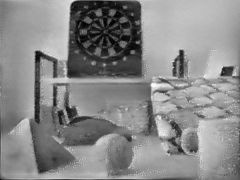}
    & \includegraphics[width=\widthplot,height=\heightplot]{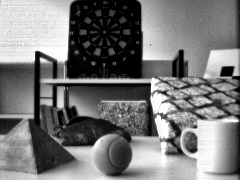}\\[\vspacecols]
    \includegraphics[width=\widthplot,height=\heightplot]{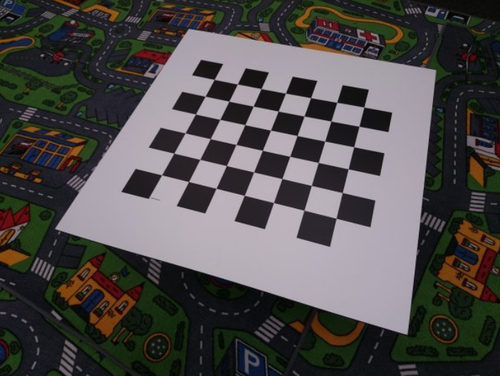}
    & \includegraphics[width=\widthplot,height=\heightplot]{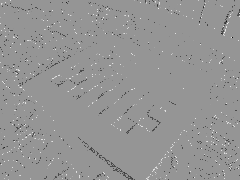}
    & \includegraphics[width=\widthplot,height=\heightplot]{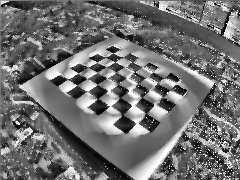}
    & \includegraphics[width=\widthplot,height=\heightplot]{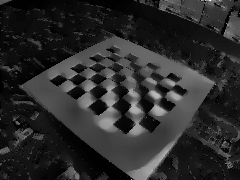}
    & \includegraphics[width=\widthplot,height=\heightplot]{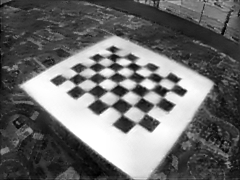}
    & \includegraphics[width=\widthplot,height=\heightplot]{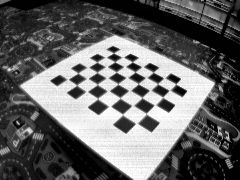}\\[\vspacecols]
    (a) Scene Preview & (b) Events & (c) HF & (d) MR & (e) Ours & (f) Ground truth\\
    \end{tabular}
\caption{Qualitative comparison of our reconstruction method with two recent competing approaches, MR \cite{Munda18ijcv} and HF \cite{Scheerlinck18accv}, on sequences from \cite{Mueggler17ijrr}, which contain ground truth frames from a DAVIS240C sensor. Our method successfully reconstructs fine details (textures in the second and third row) compared to other methods, while avoiding ghosting effects (particulary visible in the shapes sequences on the fourth row).
}
\label{fig:comp_event_camera_dataset_supp}
\end{figure*}

\setlength{\tabcolsep}{0.3ex} %
\global\long\def\heightplot{3.4cm} %
\global\long\def\widthplot{3.4cm} %
\global\long\def\vspacecols{0.3ex} %
\begin{figure*}
	\centering
    \begin{tabular}{ccccc}
    \includegraphics[width=\widthplot,height=\heightplot]{images/comp_Bardow_dataset/events/1_jumping.png}
    & \includegraphics[width=\widthplot,height=\heightplot]{images/comp_Bardow_dataset/comp_MR/jumping/1_SOFIE.png}
    & \includegraphics[width=\widthplot,height=\heightplot]{images/comp_Bardow_dataset/comp_MR/jumping/1_CF_filtered.png}
    & \includegraphics[width=\widthplot,height=\heightplot]{images/comp_Bardow_dataset/comp_MR/jumping/1_MR.png}
    & \includegraphics[width=\widthplot,height=\heightplot]{images/comp_Bardow_dataset/comp_MR/jumping/1_ours_LRT_HDR.png}\\[\vspacecols]
    \includegraphics[width=\widthplot,height=\heightplot]{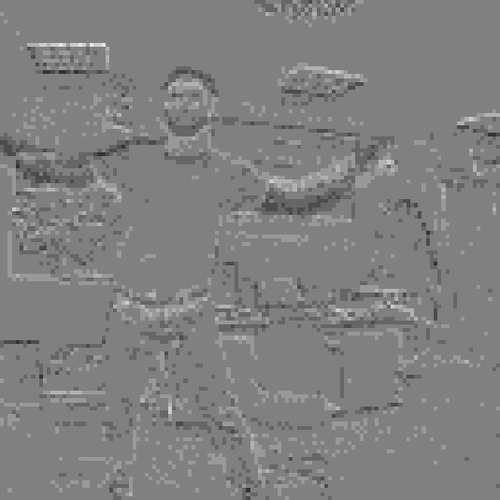}
    & \includegraphics[width=\widthplot,height=\heightplot]{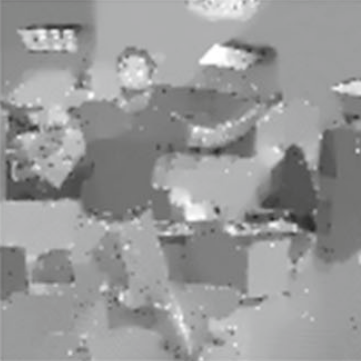}
    & \includegraphics[width=\widthplot,height=\heightplot]{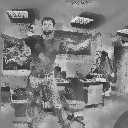}
    & \includegraphics[width=\widthplot,height=\heightplot]{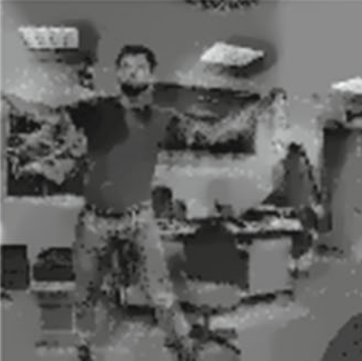}
    & \includegraphics[width=\widthplot,height=\heightplot]{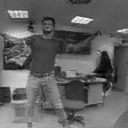}\\[\vspacecols]
    \includegraphics[width=\widthplot,height=\heightplot]{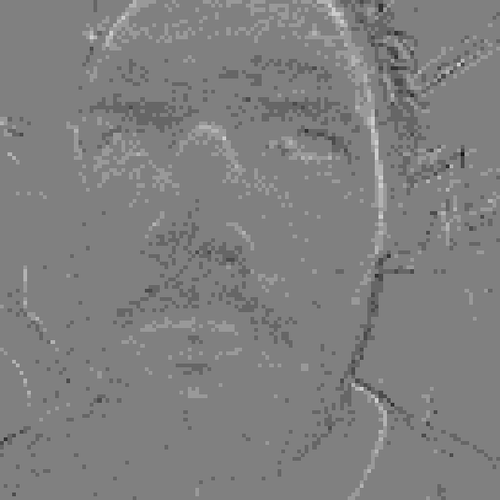}
    & \includegraphics[width=\widthplot,height=\heightplot]{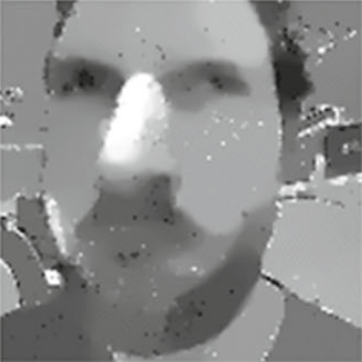}
    & \includegraphics[width=\widthplot,height=\heightplot]{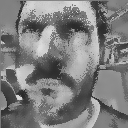}
    & \includegraphics[width=\widthplot,height=\heightplot]{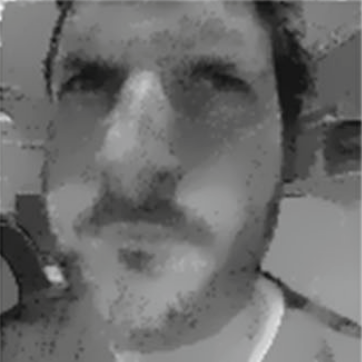}
    & \includegraphics[width=\widthplot,height=\heightplot]{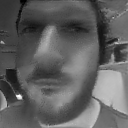}\\[\vspacecols]
    \includegraphics[width=\widthplot,height=\heightplot]{images/comp_Bardow_dataset/events/2_face.png}
    & \includegraphics[width=\widthplot,height=\heightplot]{images/comp_Bardow_dataset/comp_MR/face/2_SOFIE.png}
    & \includegraphics[width=\widthplot,height=\heightplot]{images/comp_Bardow_dataset/comp_MR/face/2_CF_filtered.png}
    & \includegraphics[width=\widthplot,height=\heightplot]{images/comp_Bardow_dataset/comp_MR/face/2_MR.png}
    & \includegraphics[width=\widthplot,height=\heightplot]{images/comp_Bardow_dataset/comp_MR/face/2_ours_LRT_HDR.png}\\[\vspacecols]
    \includegraphics[width=\widthplot,height=\heightplot]{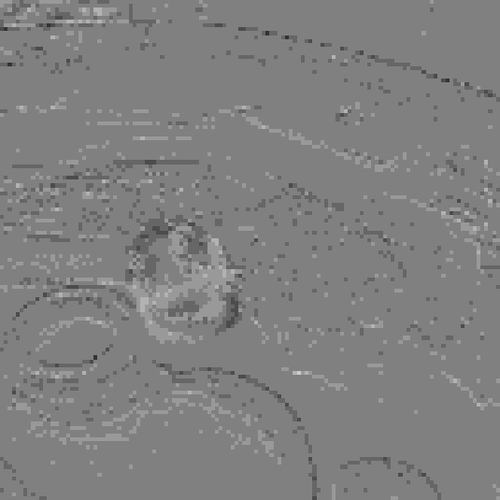}
    & \includegraphics[width=\widthplot,height=\heightplot]{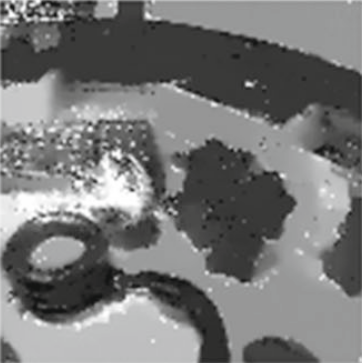}
    & \includegraphics[width=\widthplot,height=\heightplot]{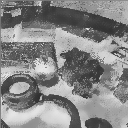}
    & \includegraphics[width=\widthplot,height=\heightplot]{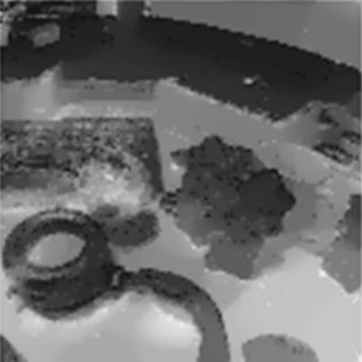}
    & \includegraphics[width=\widthplot,height=\heightplot]{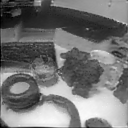}\\[\vspacecols]
    (a) Events & (b) SOFIE \cite{Bardow16cvpr} & (c) HF \cite{Scheerlinck18accv} & (d) MR \cite{Munda18ijcv} & (e) Ours\\
    \end{tabular}
\caption{Qualitative comparison of our reconstruction method with various competing approaches. We used the datasets from \cite{Bardow16cvpr}. The dataset does not contain ground truth images, thus only a qualitative comparison is possible. For SOFIE and MR, we used images provided by the authors, for which the parameters were tuned for each dataset. For HF, we ran the code provided by the authors, manually tuned the parameters on these datasets to achieve the best visual quality, and additionally applied a bilateral filter to clean the high frequency noise present in the original reconstructions.
}
\label{fig:comp_bardow_supp}
\end{figure*}

\global\long\def\heightplot{3.5cm} %
\global\long\def\widthplot{4.658cm} %
\begin{figure*}
	\centering
    \begin{tabular}{ccc}
    \includegraphics[width=\widthplot,height=\heightplot]{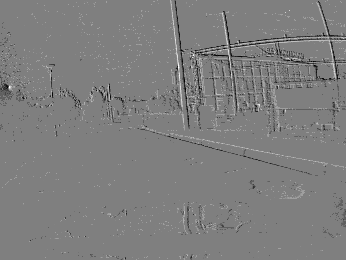}
    & \includegraphics[width=\widthplot,height=\heightplot]{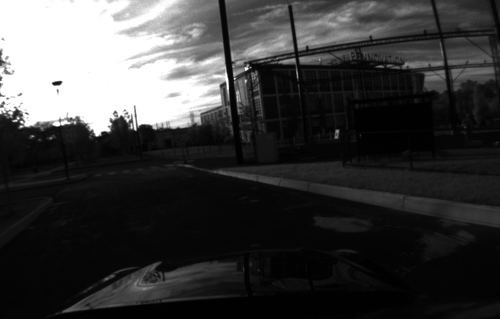}
    & \includegraphics[width=\widthplot,height=\heightplot]{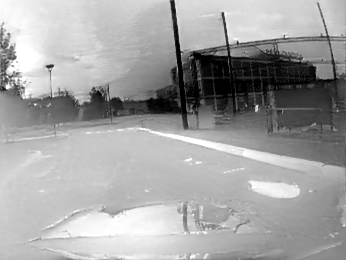}\\[\vspacecols]
    \includegraphics[width=\widthplot,height=\heightplot]{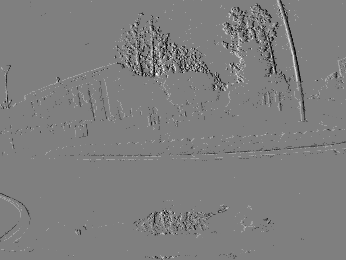}
    & \includegraphics[width=\widthplot,height=\heightplot]{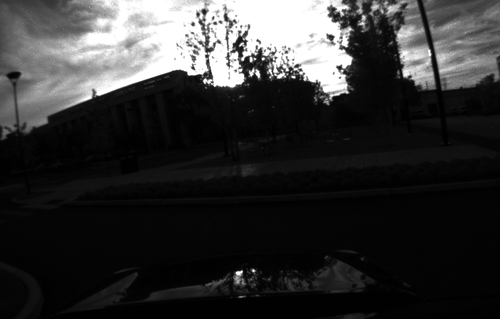}
    & \includegraphics[width=\widthplot,height=\heightplot]{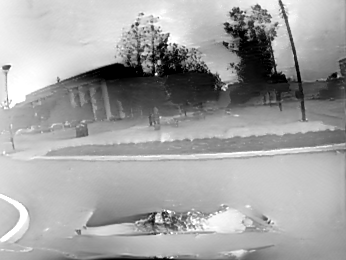}\\[\vspacecols]
    \includegraphics[width=\widthplot,height=\heightplot]{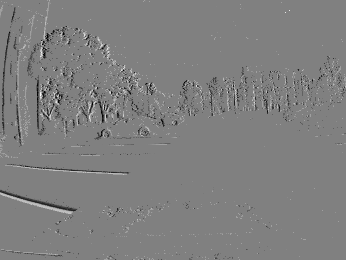}
    & \includegraphics[width=\widthplot,height=\heightplot]{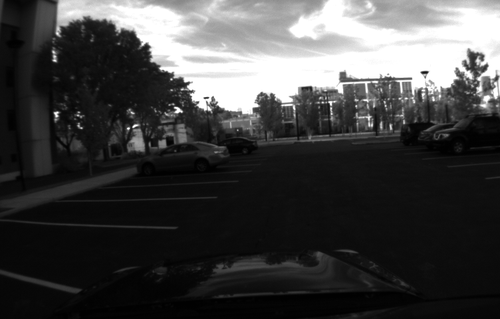}
    & \includegraphics[width=\widthplot,height=\heightplot]{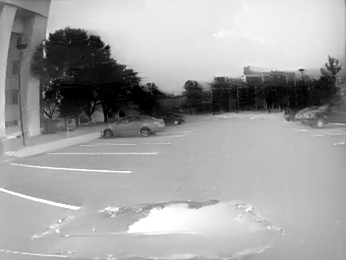}\\[\vspacecols]
    \includegraphics[width=\widthplot,height=\heightplot]{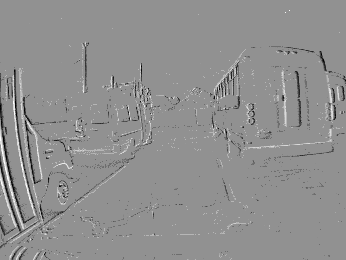}
    & \includegraphics[width=\widthplot,height=\heightplot]{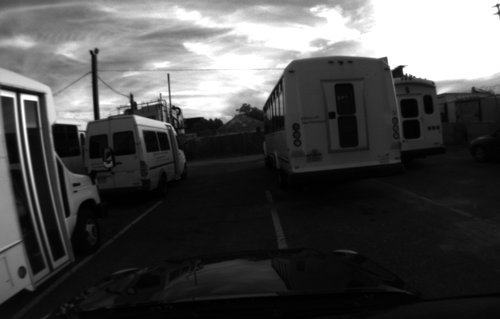}
    & \includegraphics[width=\widthplot,height=\heightplot]{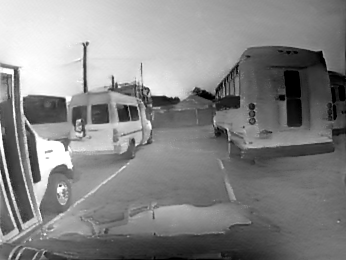}\\[\vspacecols]
    \includegraphics[width=\widthplot,height=\heightplot]{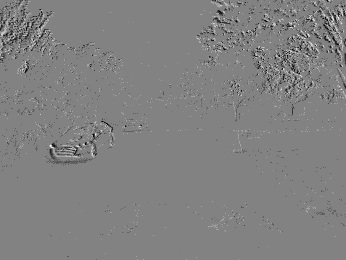}
    & \includegraphics[width=\widthplot,height=\heightplot]{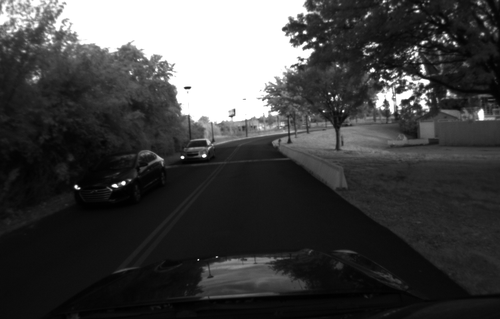}
    & \includegraphics[width=\widthplot,height=\heightplot]{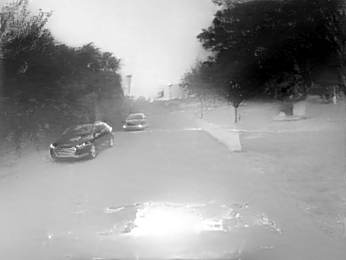}\\[\vspacecols]
    (a) Events & (b) VI sensor frame & (c) Our reconstruction\\
    \end{tabular}
\caption{Example HDR reconstructions on the MVSEC automotive dataset \cite{Zhu18ral}. The standard frames were recorded with a high-quality VI sensor with auto-exposure activated. Because the camera is facing directly the sun, the standard frames (b) are either under- or over-exposed since the limited dynamic range of the standard sensor cannot cope with the high dynamic range of the scene. By contrast, the events (a) capture the whole dynamic range of the scene, which our method successfully reconstructs to high dynamic range images (c), allow to discover details that were not visible in the standard frames.
}
\label{fig:qualitative_hdr_supp}
\end{figure*}

\global\long\def\heightplot{3.5cm} %
\global\long\def\widthplot{4.658cm} %
\begin{figure*}
	\centering
    \begin{tabular}{ccc}
    \includegraphics[width=\widthplot,height=\heightplot]{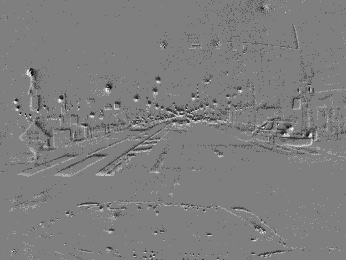}
    & \includegraphics[width=\widthplot,height=\heightplot]{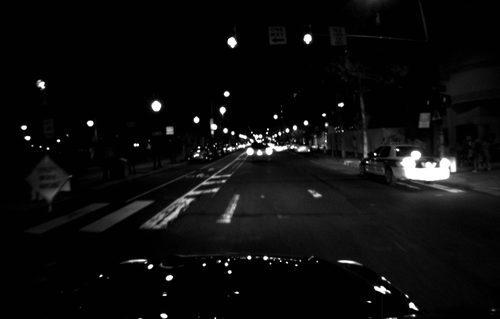}
    & \includegraphics[width=\widthplot,height=\heightplot]{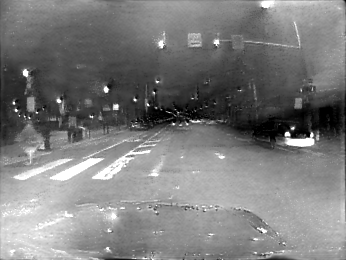}\\[\vspacecols]
    \includegraphics[width=\widthplot,height=\heightplot]{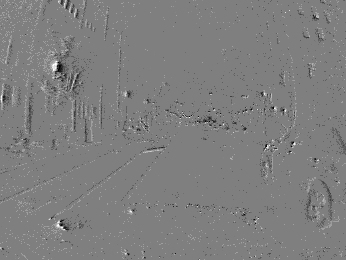}
    & \includegraphics[width=\widthplot,height=\heightplot]{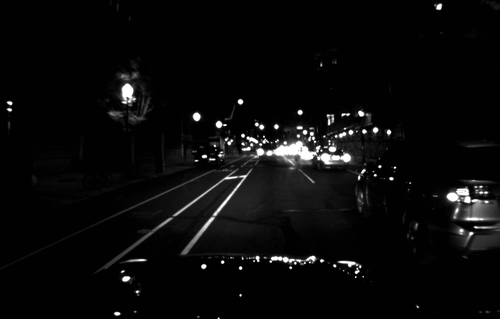}
    & \includegraphics[width=\widthplot,height=\heightplot]{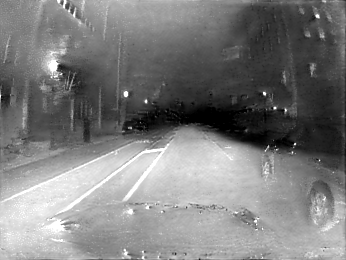}\\[\vspacecols]
    \includegraphics[width=\widthplot,height=\heightplot]{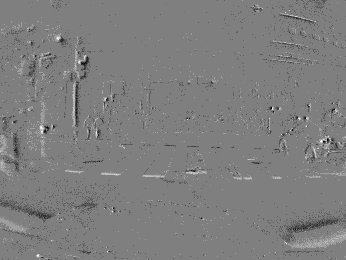}
    & \includegraphics[width=\widthplot,height=\heightplot]{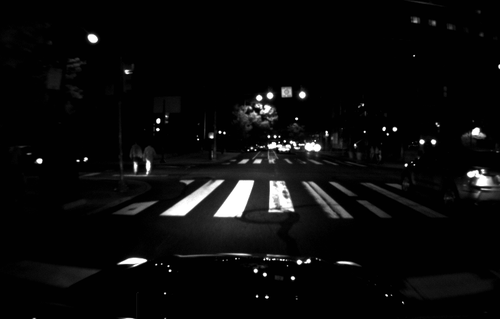}
    & \includegraphics[width=\widthplot,height=\heightplot]{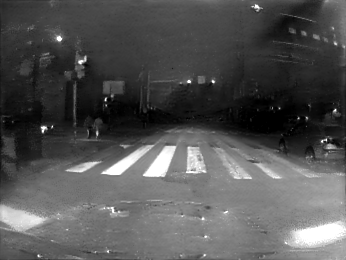}\\[\vspacecols]
    \includegraphics[width=\widthplot,height=\heightplot]{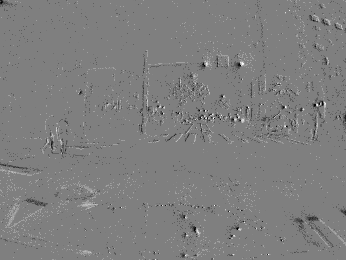}
    & \includegraphics[width=\widthplot,height=\heightplot]{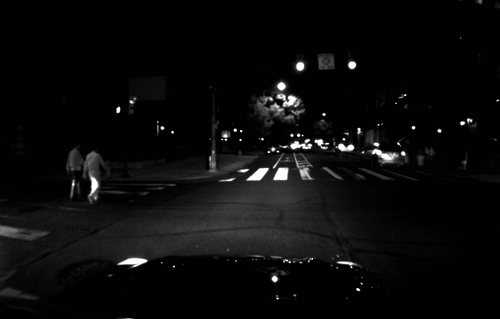}
    & \includegraphics[width=\widthplot,height=\heightplot]{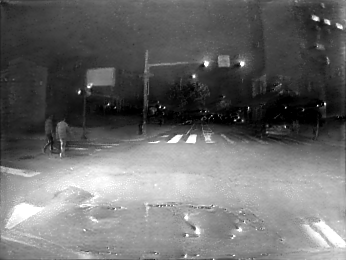}\\[\vspacecols]
    \includegraphics[width=\widthplot,height=\heightplot]{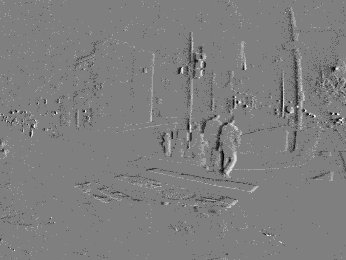}
    & \includegraphics[width=\widthplot,height=\heightplot]{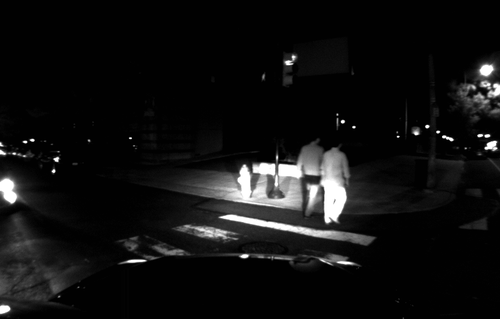}
    & \includegraphics[width=\widthplot,height=\heightplot]{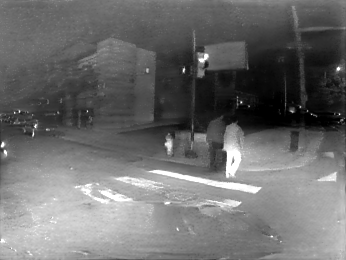}\\[\vspacecols]
    (a) Events & (b) VI sensor frame & (c) Our reconstruction\\
    \end{tabular}
\caption{Example HDR reconstructions on the MVSEC automotive dataset \cite{Zhu18ral} at night. The standard frames were recorded with a high-quality VI sensor with auto-exposure activated. Because of low light during the night, the standard frames (b) are severely degraded. By contrast, the events (a) still can capture the whole dynamic range of the scene, which our method successfully recovers (c), allowing to discover details that were not visible in the standard frames.
}
\label{fig:qualitative_night_supp}
\end{figure*}

\global\long\def\heightplot{2.03cm} %
\global\long\def\widthplot{2.7cm} %
\global\long\def\vspacecols{0.05cm} %
\setlength{\tabcolsep}{0.05cm} %
\global\long\def\vspacecols{0.12cm} %

\begin{figure}
	\centering
    \begin{tabular}{ccc}

    \rotatebox{90}{\kern0.26cm \footnotesize Indoors \cite{Scheerlinck19cvprw}}
    \includegraphics[width=\widthplot,height=\heightplot]{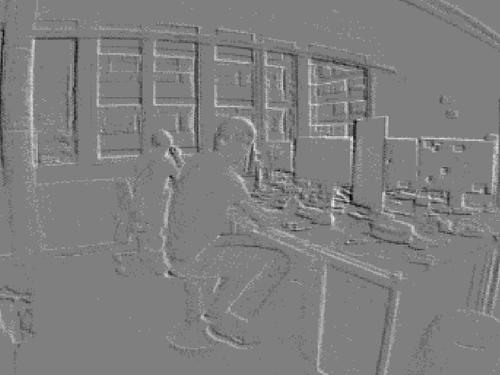}
    & \includegraphics[width=\widthplot,height=\heightplot]{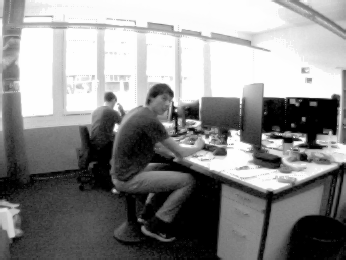}
    & \includegraphics[width=\widthplot,height=\heightplot]{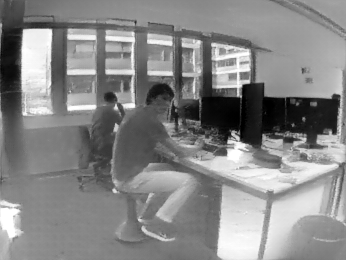}\\[\vspacecols]

    \rotatebox{90}{\kern0.285cm \footnotesize Outdoors \cite{Scheerlinck18accv}}
    \includegraphics[width=\widthplot,height=\heightplot]{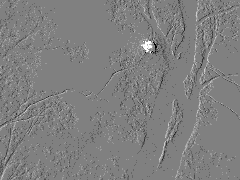}
    & \includegraphics[width=\widthplot,height=\heightplot]{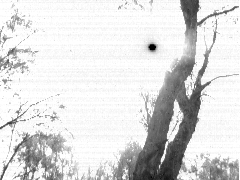}
    & \includegraphics[width=\widthplot,height=\heightplot]{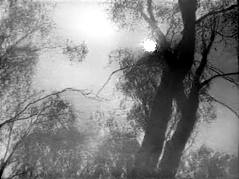}\\[\vspacecols]

    \rotatebox{90}{\kern0.05cm \footnotesize Night Drive \cite{Zhu18ral}}
    \includegraphics[width=\widthplot,height=\heightplot]{images/qualitative/MVSEC/night/1098_events.png}
    & \includegraphics[width=\widthplot,height=\heightplot]{images/qualitative/MVSEC/night/1098_VISENSOR.png}
    & \includegraphics[width=\widthplot,height=\heightplot]{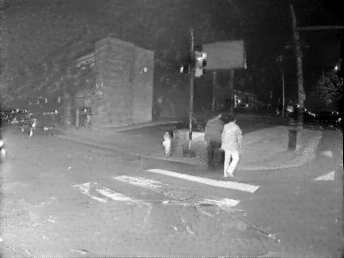}\\[\vspacecols]

    \rotatebox{90}{Night Drive \cite{Scheerlinck18accv}}
    \includegraphics[width=\widthplot,height=\heightplot]{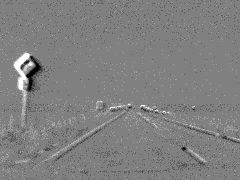}
    & \includegraphics[width=\widthplot,height=\heightplot]{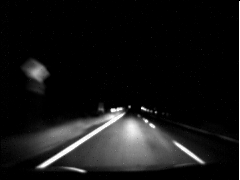}
    & \includegraphics[width=\widthplot,height=\heightplot]{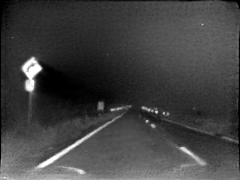}\\[\vspacecols]
    (a) \footnotesize Events & (b) \footnotesize Frame & (c) \footnotesize Reconstruction\\
    \end{tabular}
\caption{Video reconstruction under challenging lighting.
First row: indoor sequence \cite{Scheerlinck19cvprw}.
Second row: outdoor sequence from \cite{Scheerlinck18accv}.
Third row: night driving sequence from \cite{Zhu18ral}.
Fourth row: night driving sequence from \cite{Scheerlinck18accv}.
The frames from the conventional camera (b) suffer from under- or over-exposure, while the events (a) capture the whole dynamic range of the scene, which our method successfully recovers (c).
}
\label{fig:hdr_reconstructions_public_datasets}
\end{figure}

\section{Object Classification}

Below we detail the exact modalities of our reconstruction method for each of the dataset which we used for our evaluation of object classification (Section 5.1 in the paper), as well as the specific architectures used and training modalities.

\mypara{N-MNIST.}
To reconstruct images with our networks, we used an event window of $\NumEvents=1@000$ events.
We passed every event sequence into our network, resulting in a video, from which we keep the final image as input for the classification network.
To match the images from the original MNIST dataset, we additionally binarize the reconstructed image (whose values lie in $[0,1]$) with a threshold of $0.5$.
The train and test images were normalized so that the mean value of each image is $0.1307$ and the variance $0.3081$.
We used the official train and test split provided in the M-NNIST dataset.
As there is no standard state of the art architecture for MNIST, we used a simple CNN architecture as our classification network, composed of the following blocks:

\begin{itemize}
  \item \small 2D convolution (stride: 5, output channels: 32) + ReLU
  \item \small 2D convolution (stride: 5, output channels: 64) + ReLU
  \item \small 2D max pooling (size: 2) + Dropout
  \item \small Fully connected layer (output size: 128 neurons) + ReLU
  \item \small Fully connected layer (output size: 10 neurons)
\end{itemize}

We used the cross entropy loss, and trained the network for 15 epochs using the ADAM optimizer, with a learning rate of 0.001.

\mypara{N-CARS.}
We used windows of events with a fixed temporal size of $\SI{20}{\ms}$, and used the last reconstructed image from the video as input to the classification network.
We used the official train and test split provided by the N-CARS dataset.
We used a ResNet18 \cite{He16cvpr} architecture (with an additional fully connected final layer with $2$ output neurons), initialized with weights pretrained on ImageNet \cite{Russakovsky15ijcv}, and fine-tuned the network using the reconstructed images from the training set for 20 epochs, using SGD with a learning rate of $0.001$ (decayed by factor of $0.1$ every $7$ epochs), and momentum of $0.1$.

\mypara{N-Caltech101.}
For image reconstruction, we used windows of $\NumEvents=10@000@$ events and used the last reconstructed image as input to the classification network.
Since there is no official train and test split for the N-Caltech101 dataset, we split the dataset randomly into two third training sequences ($5@863$ sequences) and one third testing sequences ($2@396$ sequences), following the methodology used by HATS \cite{Sironi18cvpr}.
The train and test images were converted to 3-channel grayscale images (\ie the three channels are the same), and normalized so that the mean value of each image is $0.485$ and the variance $0.229$.
We also performed data augmentation at train time (random horizontal flips, and random crop of size $224$).
At test time, we resized all the images to $256 \timess 256$ and cropped the image around the center with a size of $224$.
We used a ResNet18 architecture (with an additional fully-connected final layer with $101$ output neurons), initialized with weights pretrained on ImageNet, and fine-tuned the network using the reconstructed images from the training set for 25 epochs using SGD with an initial learning rate of $0.001$ (decayed by a factor of $0.1$ every 7 epochs) and momentum of $0.1$.
Fig.~\ref{fig:preview_ncaltech101_supp} shows additional reconstruction examples from the N-Caltech101 dataset.

\setlength{\tabcolsep}{0.3ex} %
\global\long\def\heightplot{2.8cm} %
\global\long\def\vspacecols{0.3ex} %
\begin{figure*}
	\centering
    \begin{tabular}{ccc}
    \includegraphics[height=\heightplot]{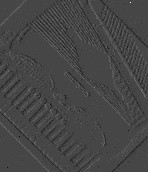}
    & \includegraphics[height=\heightplot]{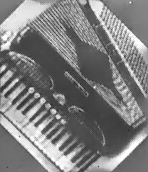}
    & \includegraphics[height=\heightplot]{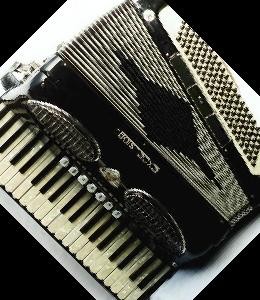}\\[\vspacecols]
    \includegraphics[height=\heightplot]{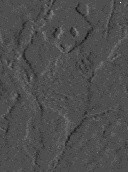}
    & \includegraphics[height=\heightplot]{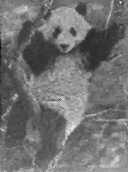}
    & \includegraphics[height=\heightplot]{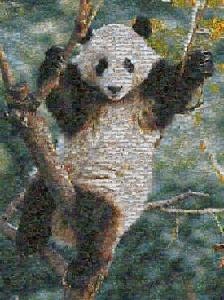}\\[\vspacecols]
    \includegraphics[height=\heightplot]{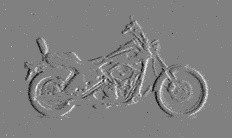}
    & \includegraphics[height=\heightplot]{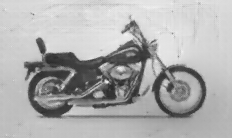}
    & \includegraphics[height=\heightplot]{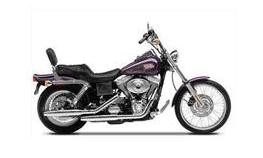}\\[\vspacecols]
    \includegraphics[height=\heightplot]{images/N-Caltech101/wild_cat/events_0001.jpg}
    & \includegraphics[height=\heightplot]{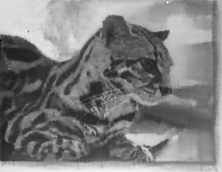}
    & \includegraphics[height=\heightplot]{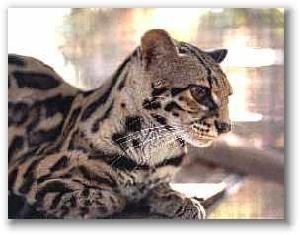}\\[\vspacecols]
    \includegraphics[height=\heightplot]{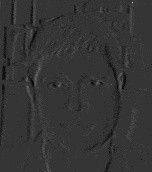}
    & \includegraphics[height=\heightplot]{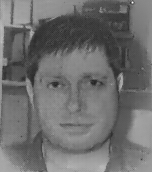}
    & \includegraphics[height=\heightplot]{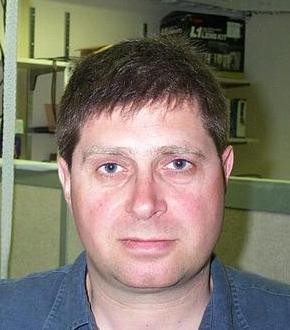}\\[\vspacecols]
    \includegraphics[height=\heightplot]{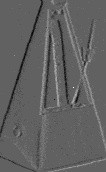}
    & \includegraphics[height=\heightplot]{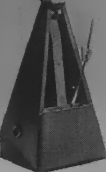}
    & \includegraphics[height=\heightplot]{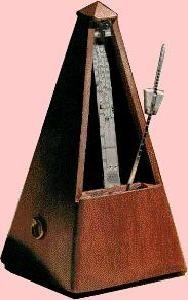}\\[\vspacecols]
    \includegraphics[height=\heightplot]{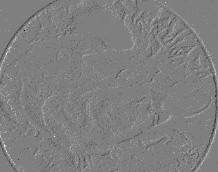}
    & \includegraphics[height=\heightplot]{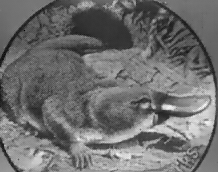}
    & \includegraphics[height=\heightplot]{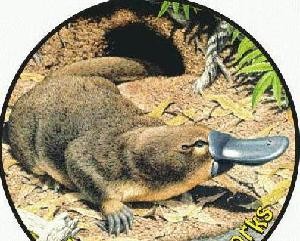}\\[\vspacecols]
    (a) Events & (b) Our Reconstruction & (c) Original Image\\
    \end{tabular}
\caption{(a) Previews of some event sequences from the N-Caltech101 dataset \cite{Orchard15fns} which features event sequences converted from the Caltech101 dataset. (b) our reconstructions (from events only) preserve many of the details and statistics of the original images (c). Note that these datasets feature planar motion (since Caltech101 images were projected on white wall to record the events), which coincides with the type of motions present in the simulated data, which explains in part the outstanding visual quality of the reconstructions.
}
\label{fig:preview_ncaltech101_supp}
\end{figure*}

\section{Visual-Inertial Odometry}

Figs.~\ref{fig:boxplots_supp_overall},~\ref{fig:boxplots_supp_translation} and~\ref{fig:boxplots_supp_6dof} provide additional results on the visual-inertial odometry experiments presented in the main paper.
Specifically, they provide, for each sequence used in our evaluation, the evolution of the mean translation and rotation error as a function of the travelled distance for our approach, UltimateSLAM (E+I), and UltimateSLAM (E+F+I).

\setlength{\tabcolsep}{0.1ex} %
\global\long\def\heightplot{2.9cm} %
\global\long\def\vspacecols{0.0ex} %
\begin{figure*}
	\centering
    \begin{tabular}{cc}
    \includegraphics[height=\heightplot,trim={0 0 8cm 0},clip,]{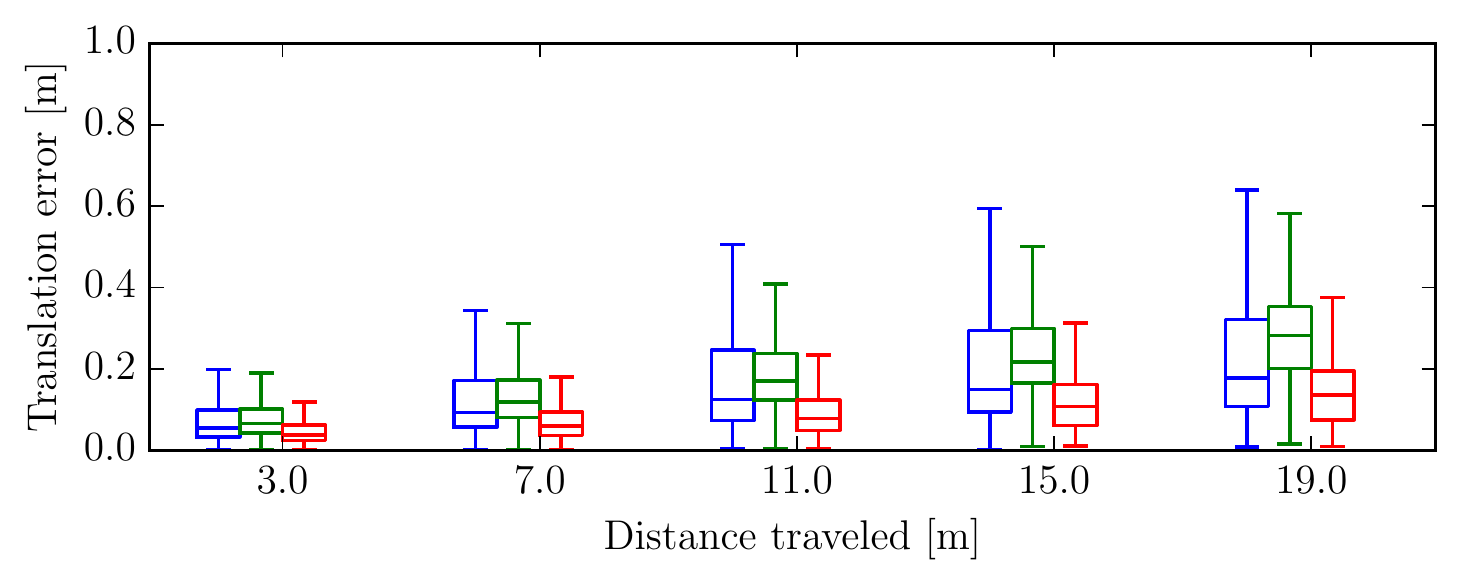}
    & \includegraphics[height=\heightplot]{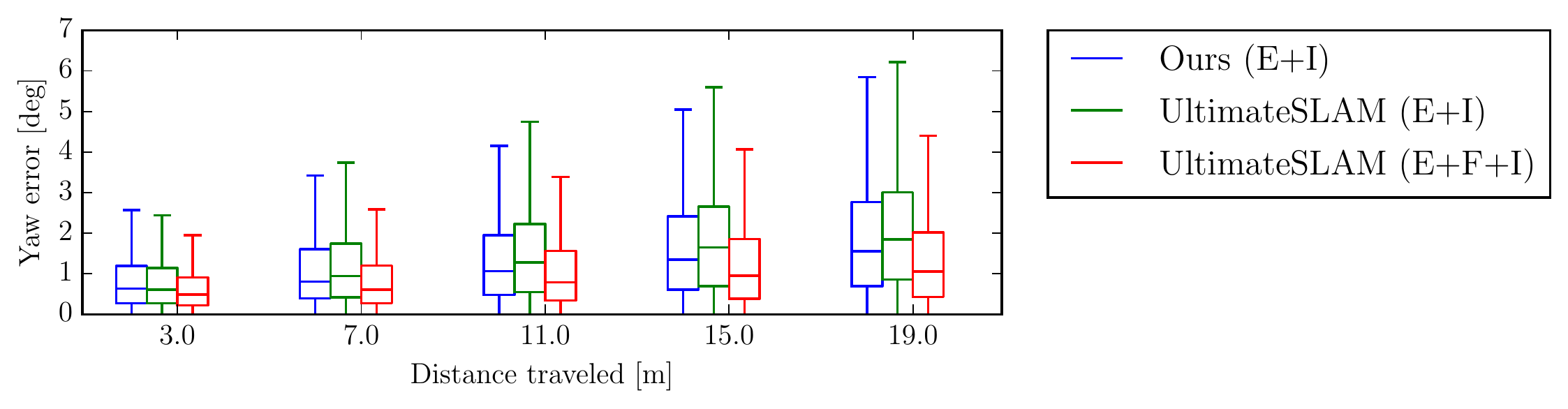}\\[0mm]
    \end{tabular}
\caption{Evolution of the overall mean translation error (in meters) and mean rotation error (in degrees), averaged across all the datasets used in our evaluation.}
\label{fig:boxplots_supp_overall}
\end{figure*}

\setlength{\tabcolsep}{0.1ex} %
\global\long\def\heightplot{2.9cm} %
\global\long\def\vspacecols{0.0ex} %
\begin{figure*}
	\centering
    \begin{tabular}{cc}
    \includegraphics[height=\heightplot,trim={0 0 8cm 0},clip,]{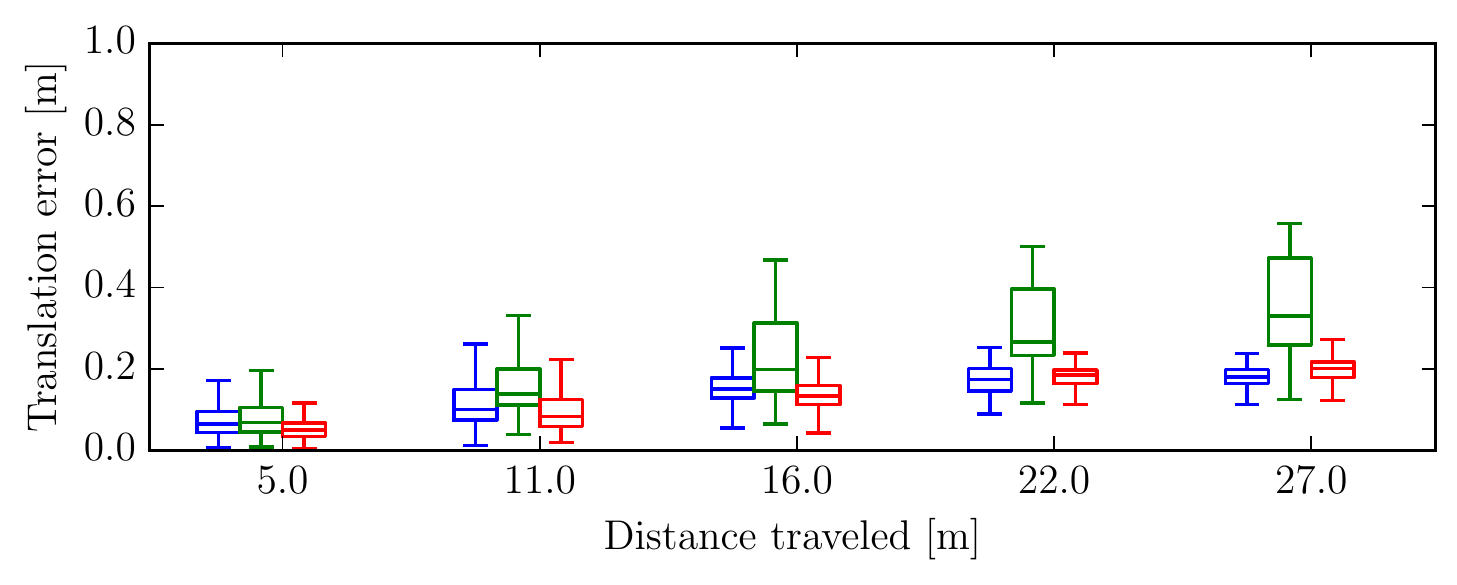}
    & \includegraphics[height=\heightplot]{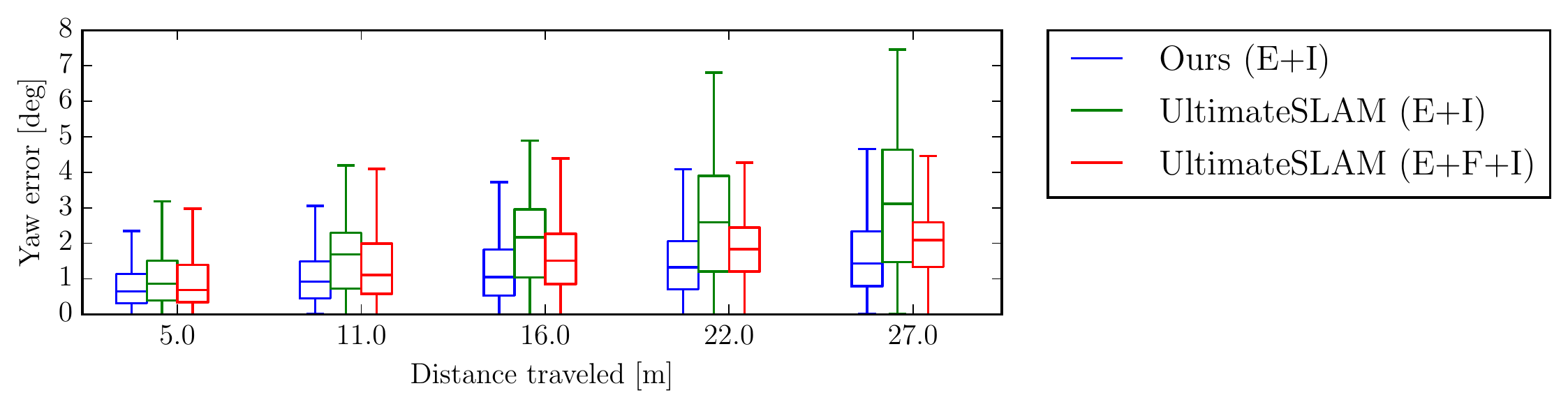}\\[0mm]
    \includegraphics[height=\heightplot,trim={0 0 8cm 0},clip,]{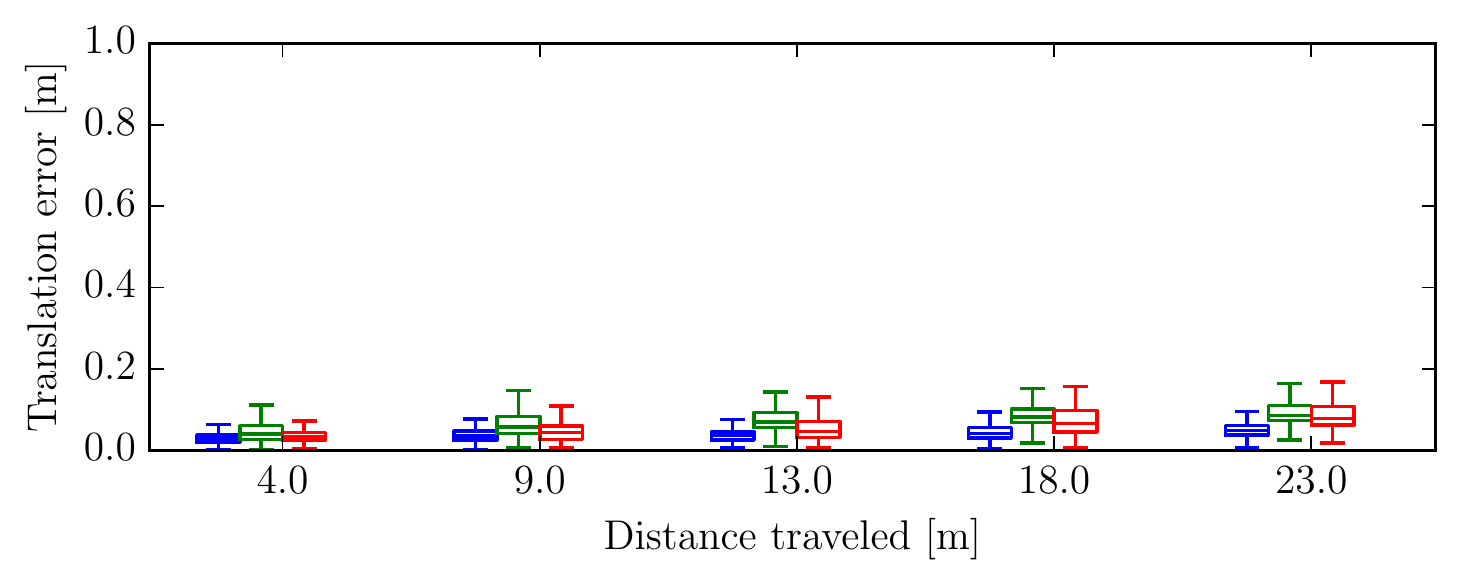}
    & \includegraphics[height=\heightplot]{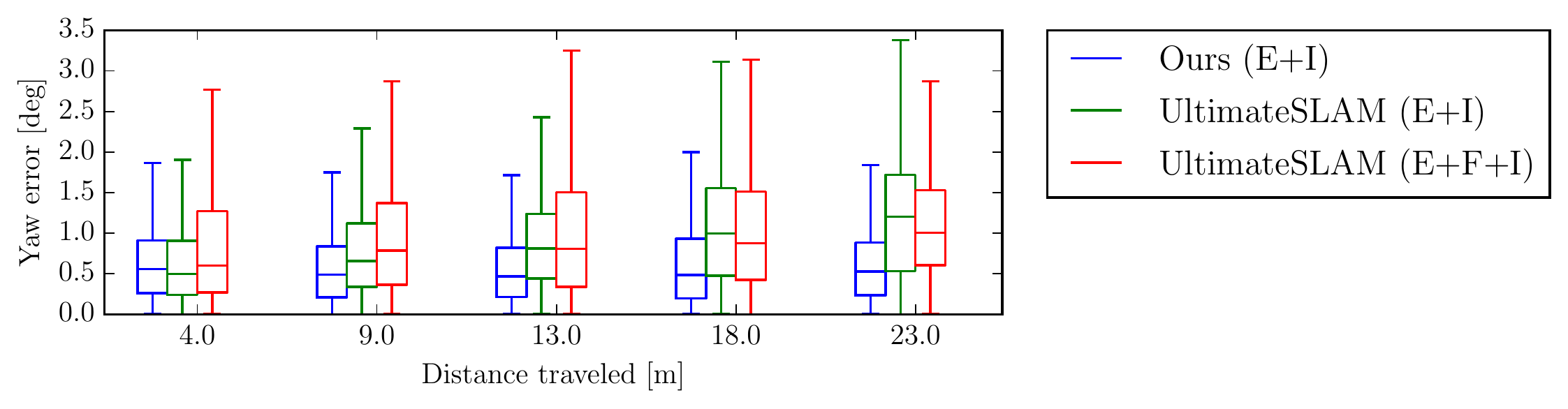}\\[0mm]
    \includegraphics[height=\heightplot,trim={0 0 8cm 0},clip,]{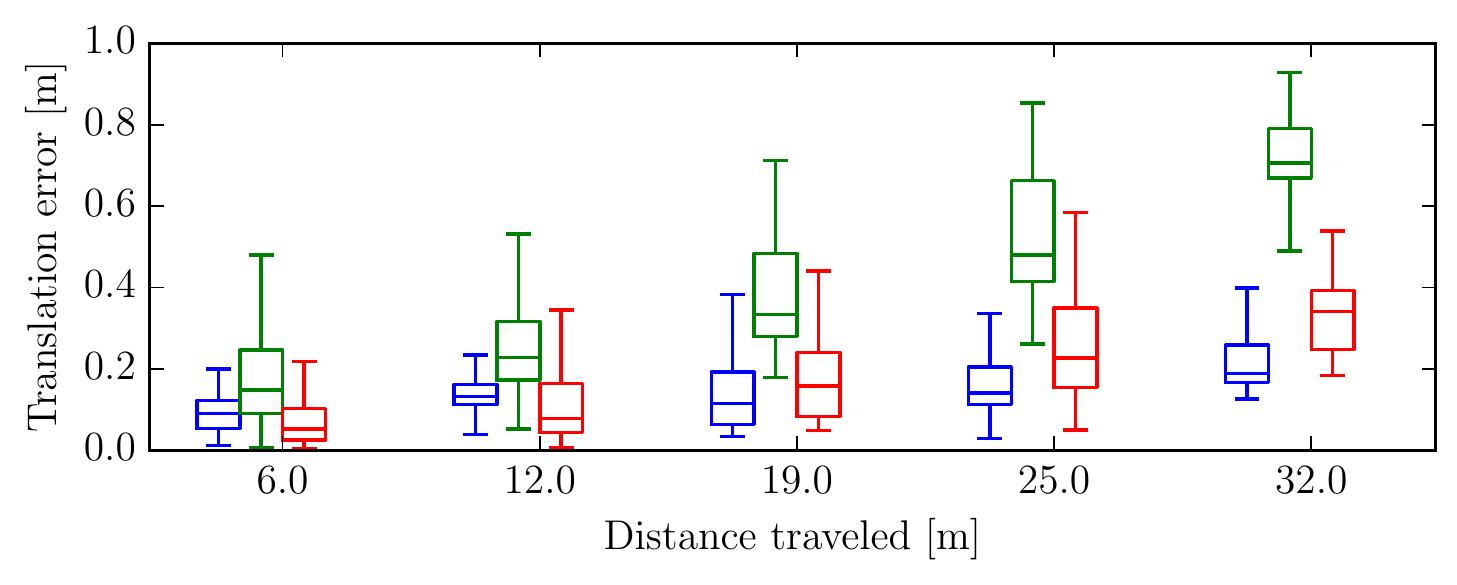}
    & \includegraphics[height=\heightplot]{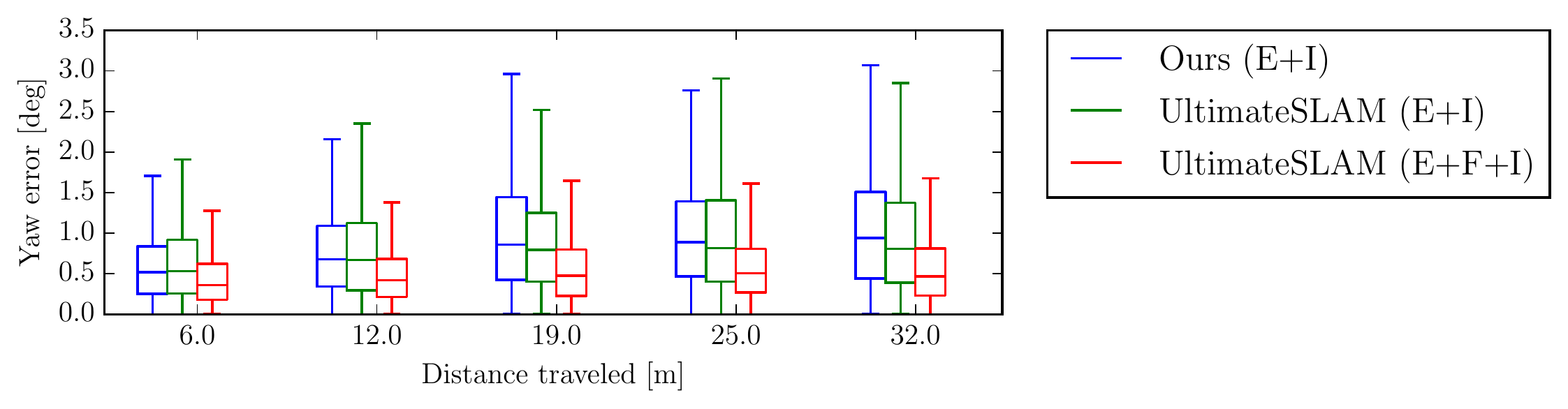}\\[0mm]
    \includegraphics[height=\heightplot,trim={0 0 8cm 0},clip,]{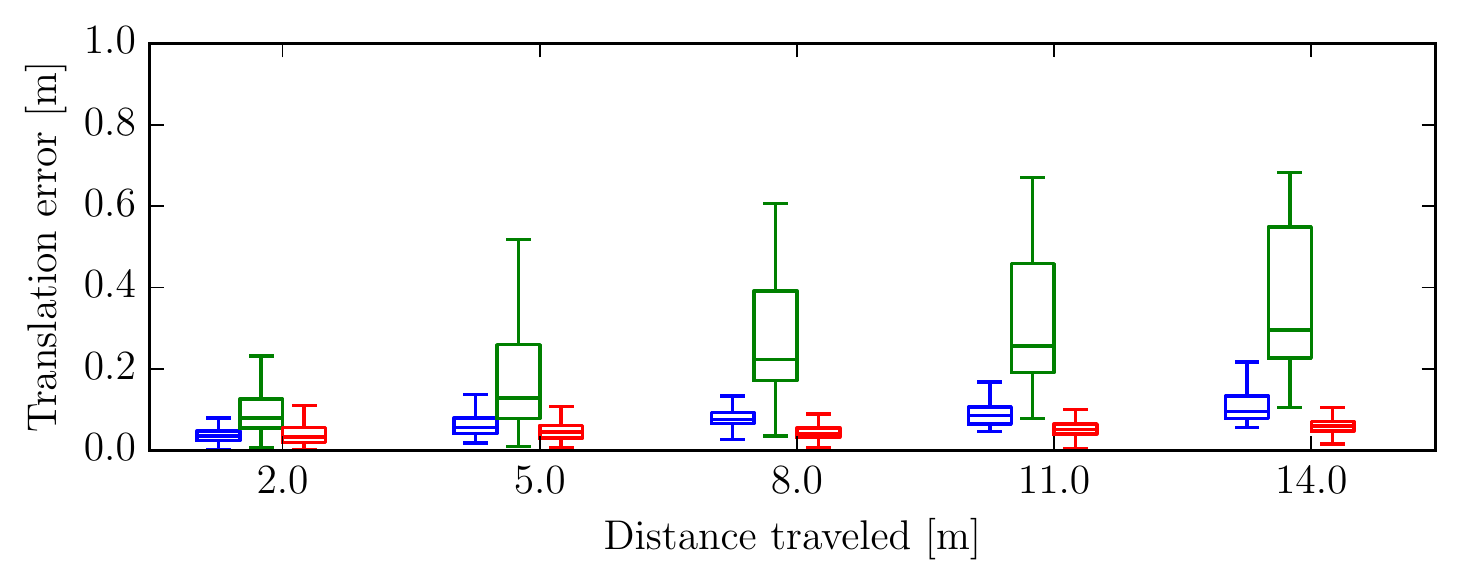}
    & \includegraphics[height=\heightplot]{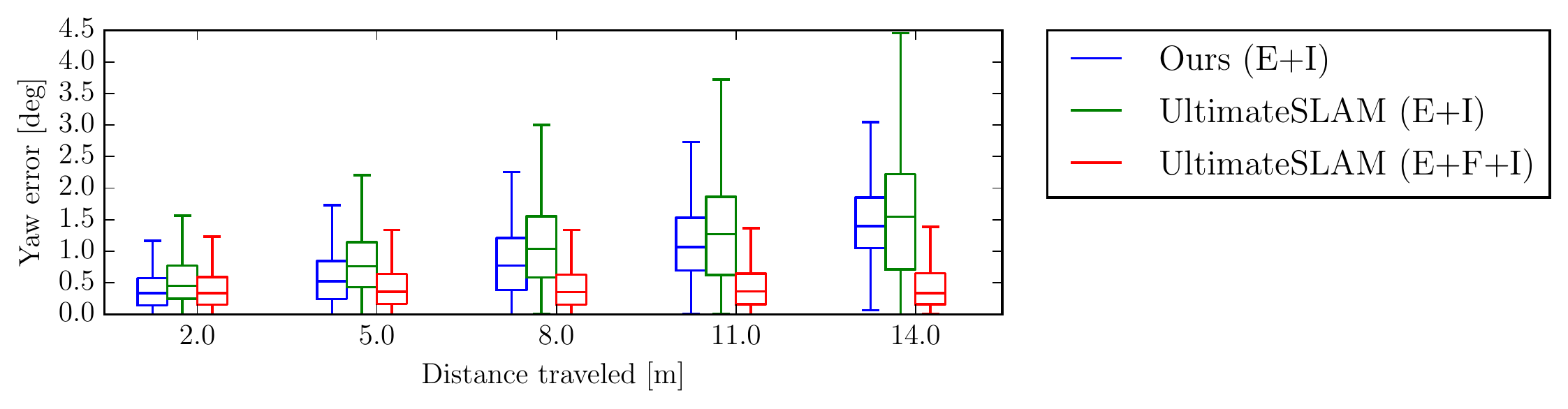}\\[0mm]
    \end{tabular}
\caption{Evolution of the mean translation error (in meters) and mean rotation error (in degrees), as a function of the travelled distance. Sequences from top to bottom: 'shapes\_translation', 'poster\_translation', 'boxes\_translation', 'dynamic\_translation'.}
\label{fig:boxplots_supp_translation}
\end{figure*}

\setlength{\tabcolsep}{0.1ex} %
\global\long\def\heightplot{2.9cm} %
\global\long\def\vspacecols{0.0ex} %
\begin{figure*}
	\centering
    \begin{tabular}{cc}
    \includegraphics[height=\heightplot,trim={0 0 8cm 0},clip,]{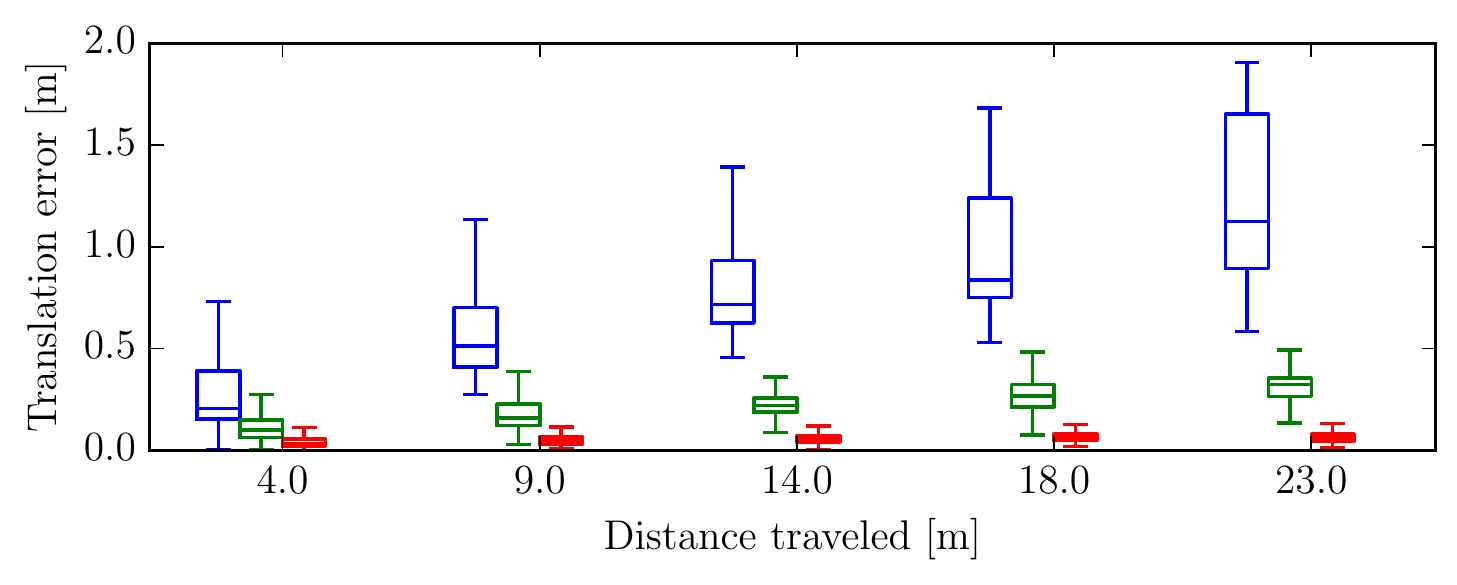}
    & \includegraphics[height=\heightplot]{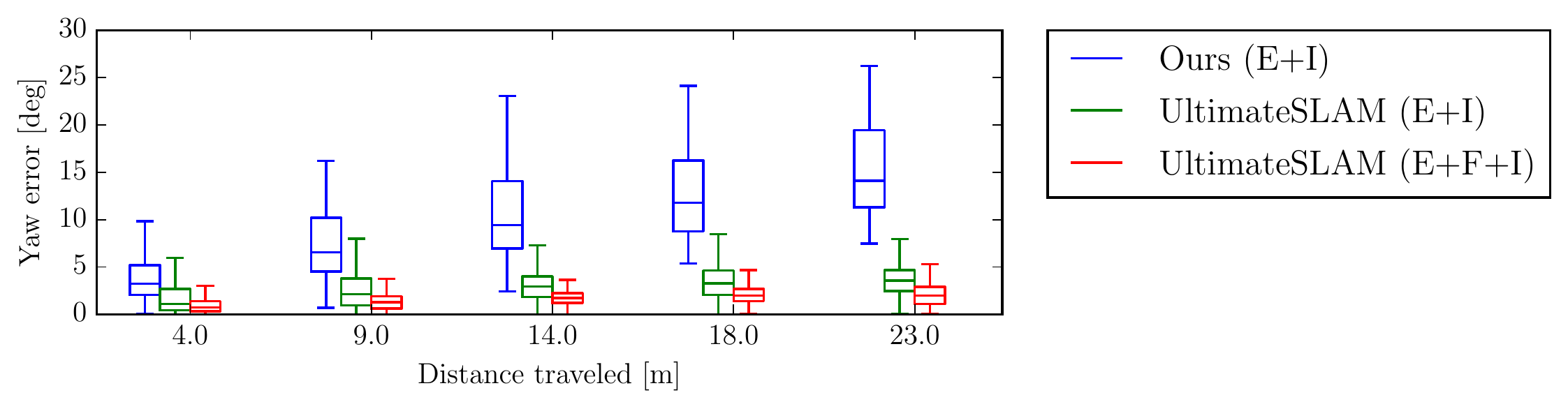}\\[0mm]
    \includegraphics[height=\heightplot,trim={0 0 8cm 0},clip,]{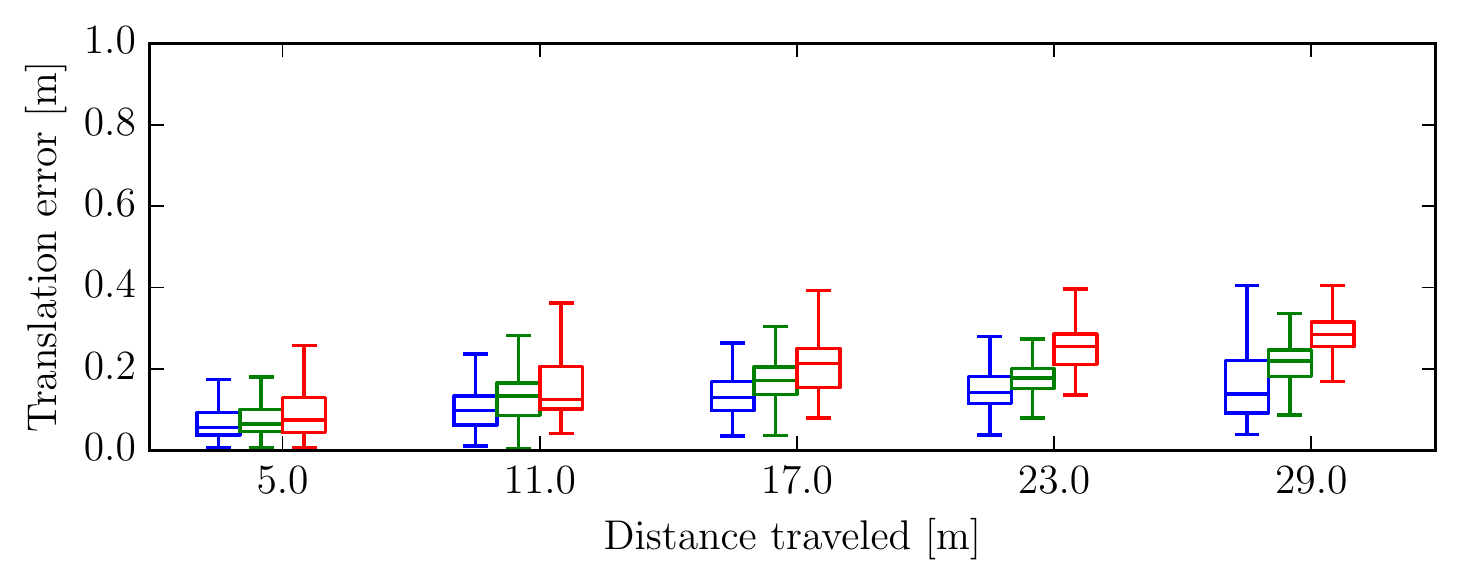}
    & \includegraphics[height=\heightplot]{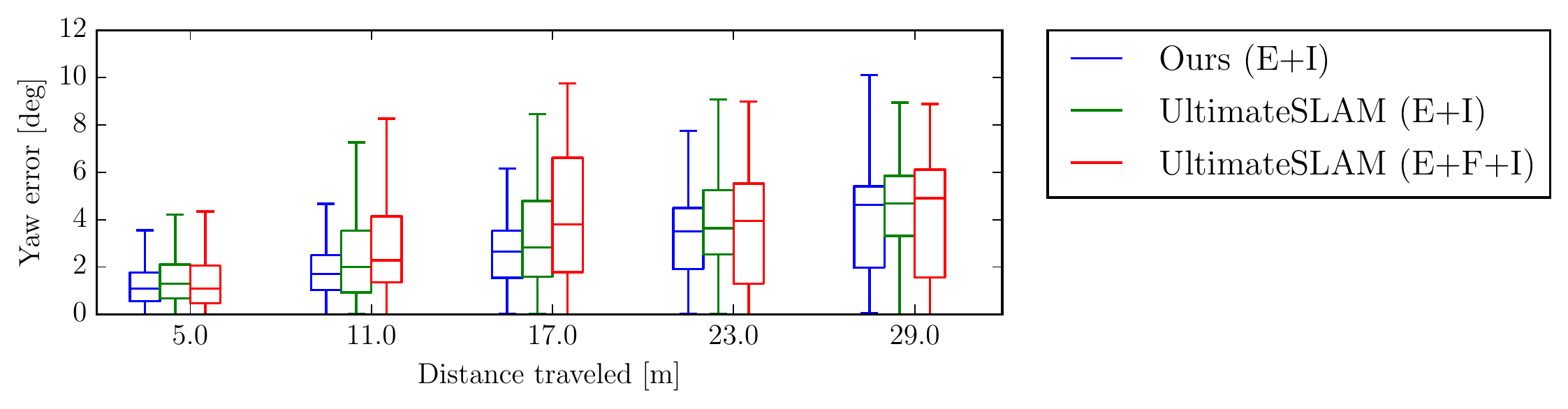}\\[0mm]
    \includegraphics[height=\heightplot,trim={0 0 8cm 0},clip,]{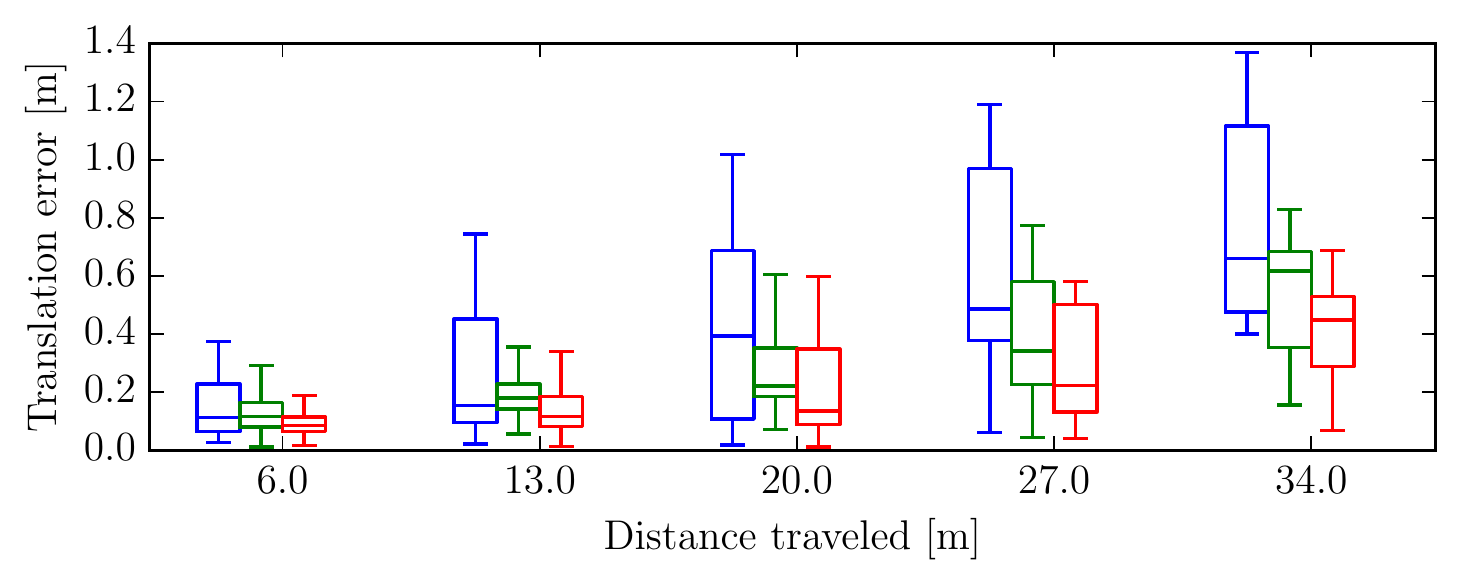}
    & \includegraphics[height=\heightplot]{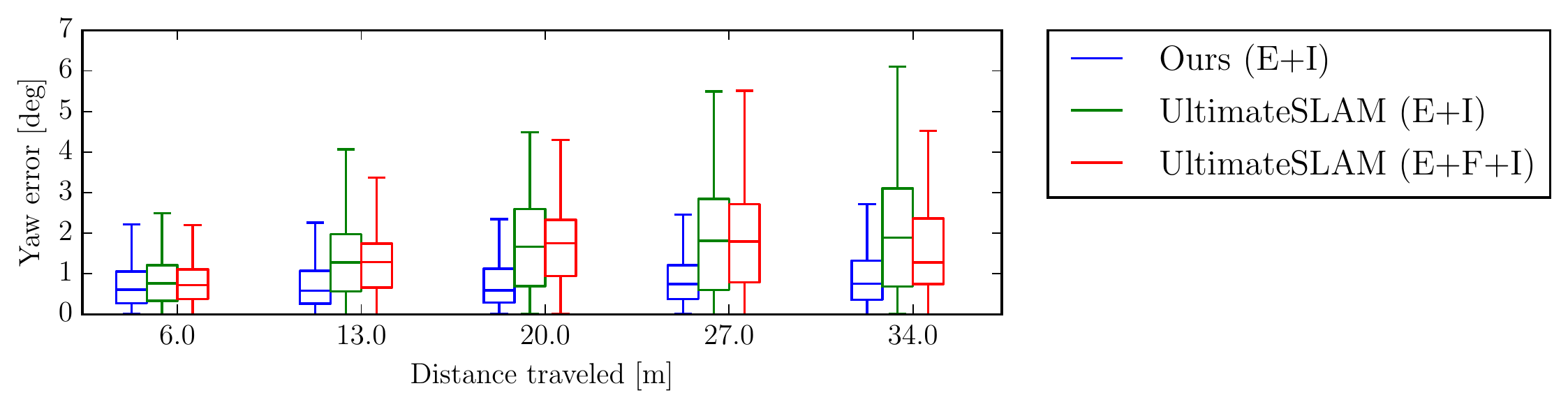}\\
    \includegraphics[height=\heightplot,trim={0 0 8cm 0},clip,]{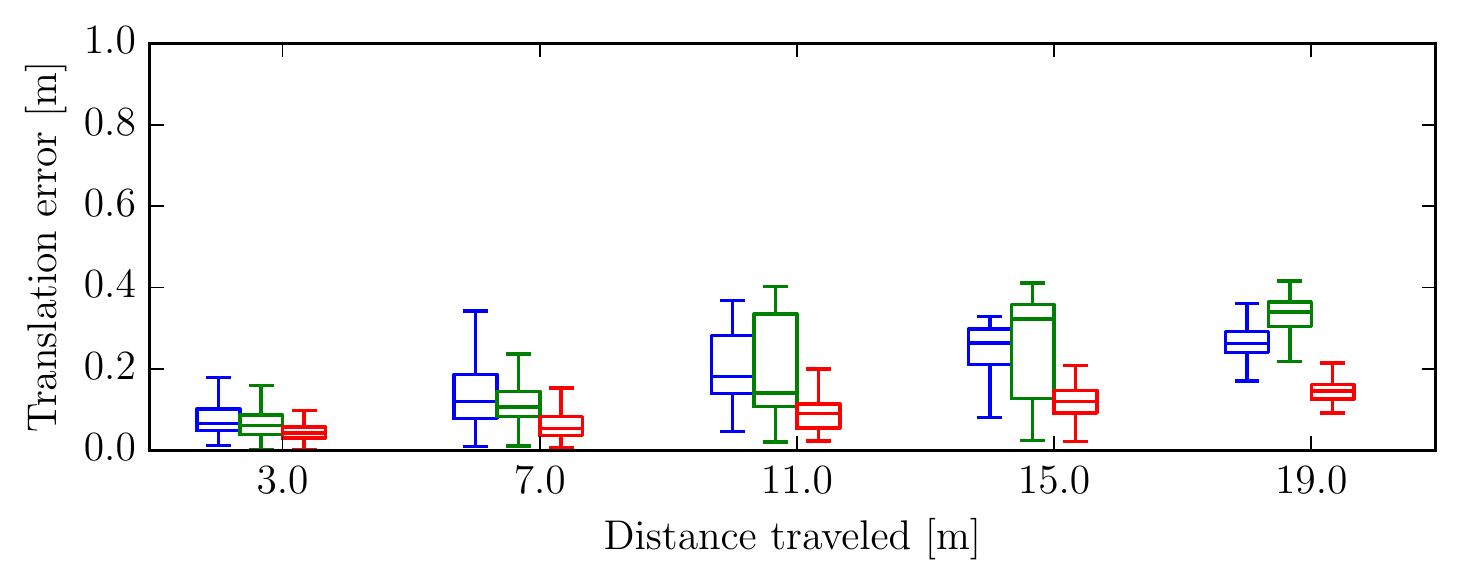}
    & \includegraphics[height=\heightplot]{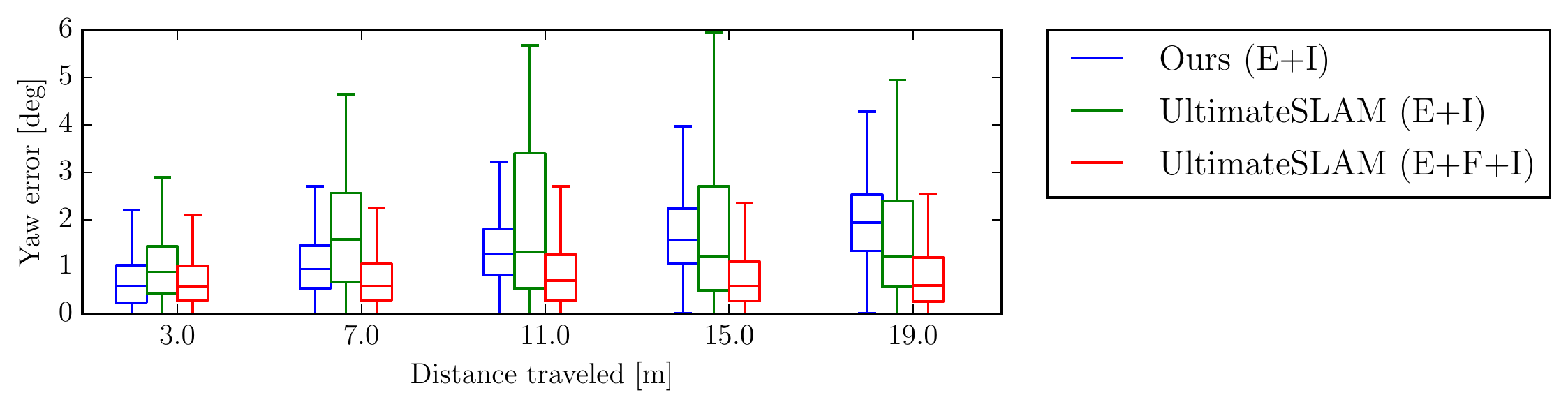}\\[0mm]
    \includegraphics[height=\heightplot,trim={0 0 8cm 0},clip,]{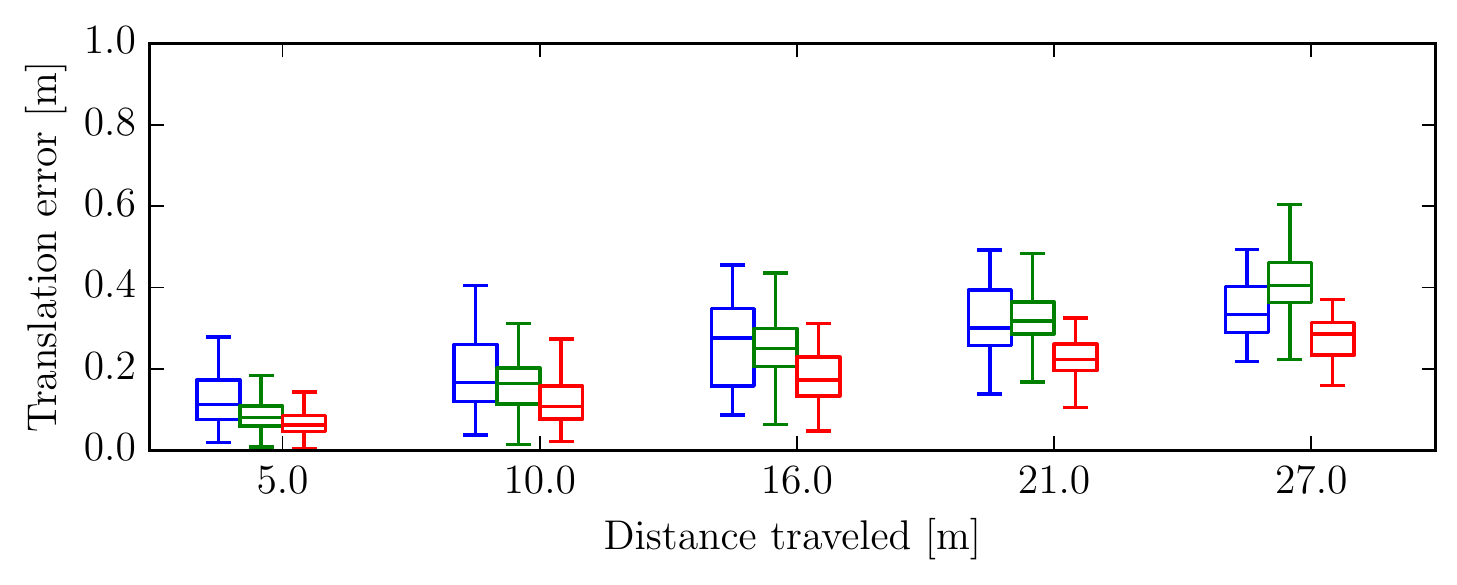}
    & \includegraphics[height=\heightplot]{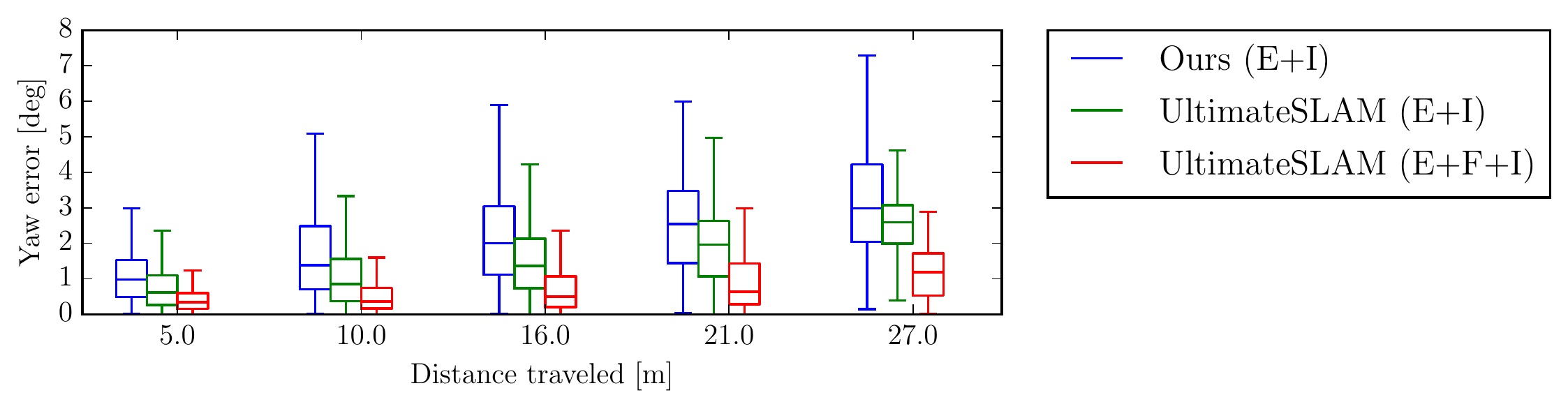}\\[0mm]

    \end{tabular}
\caption{Evolution of the mean translation error (in meters) and mean rotation error (in degrees), as a function of the travelled distance. Sequences from top to bottom: 'shapes\_6dof', 'poster\_6dof', 'boxes\_6dof', 'dynamic\_6dof', 'hdr\_boxes'.}
\label{fig:boxplots_supp_6dof}
\end{figure*}

\end{document}